%% file: arxiv-main.tex
\documentclass[12pt]{article}

\usepackage{times}
\usepackage[utf8]{inputenc}
\usepackage{graphicx}
\usepackage{amsmath,amssymb,amsfonts}
\usepackage{hyperref}
\usepackage[T1]{fontenc}
\usepackage{multirow}
\usepackage{algorithm}
\usepackage{algpseudocode}
\usepackage{subfiles}
\usepackage{tabularx}
\usepackage{booktabs}
\usepackage{arydshln}
\usepackage{xurl}

\usepackage{adjustbox}

\usepackage{ifthen}
\usepackage{caption}
\usepackage{bbm}
\usepackage{array}
\usepackage[table]{xcolor}
\usepackage{colortbl}
\usepackage{longtable}
\usepackage{ragged2e}
\usepackage{rotating}
\usepackage{placeins}

\definecolor{forcecol}{RGB}{245,245,255}
\definecolor{velcol}{RGB}{245,255,245}
\definecolor{cor}{RGB}{30, 100, 200}
\definecolor{cor}{RGB}{0,0,0}
\definecolor{cor2}{RGB}{30, 100, 200}
\definecolor{cor2}{RGB}{0,0,0}
\definecolor{cor3}{RGB}{30, 100, 200}
\definecolor{revisionblue}{RGB}{0, 70, 200} 
\definecolor{frontierModels}{RGB}{170, 0, 170} 

\newcommand{\bestdrl}[1]{\underline{#1}}
\newcolumntype{L}[1]{>{\RaggedRight\arraybackslash}p{#1}}
\newenvironment{sciabstract}{%
\begin{quote} \bf}
{\end{quote}}

\newcommand{\beginsupplement}{%
    \setcounter{table}{0}
    \renewcommand{\thetable}{S\arabic{table}}%
    \setcounter{figure}{0}
    \renewcommand{\thefigure}{S\arabic{figure}}%
}

\title{\textcolor{cor}{Models Matter:} Physical Interaction Effects Reveal the Necessary Model Structure for Reliable Contact-Rich Manipulation Learning}

\author{
Florian Voigt$^{1,2}$ and Sami Haddadin$^{2,\ast}$\\
\\
\normalsize{$^1$Chair of Robotics and Systems Intelligence, MIRMI,}\\
\normalsize{Technical University of Munich,}\\
\normalsize{Georg-Brauchle-Ring 60-62, Munich, 80992, Germany}\\
\normalsize{$^2$Mohamed Bin Zayed University of Artificial Intelligence,}\\
\normalsize{Masdar City, Abu Dhabi, United Arab Emirates}\\
\normalsize{$^\ast$Corresponding author:
sami.haddadin@mbzuai.ac.ae}
}

\date{}

\begin{document}

\baselineskip24pt

\captionsetup{
    font={small, stretch=0.9},
    labelfont=bf,
    aboveskip=4pt,
    belowskip=0pt,
    width=\textwidth
}

\maketitle

\begin{sciabstract}
A central question in embodied intelligence is what physical structure must be represented for a learning system to acquire reliable real-world skills. Scaling data and model capacity can improve learning when task-relevant structure is accessible in the observation stream, but contact-rich interaction provides a distinct regime: crucial physical effects are only indirectly or not reliably observable through standard robot interfaces. Here we use contact-rich manipulation, specifically tight-clearance
insertion with tolerances ranging from below 1\,mm down to the
micrometre level, as an experimental test of whether scale within a
structurally incomplete representation can substitute for model
knowledge.

Through a systematic two-year study across 22 contact-rich insertion tasks spanning electrical connectors, mechanical sockets, keys, and plugs with sub-millimetre clearances and physical regimes including pressure, deformation, interference fit, clamping, and friction, we evaluate a continuum of practically actionable learning architectures with increasing model knowledge --- from Proximal Policy Optimisation and Soft Actor-Critic in joint space with gravity compensation only, through architectures with increasing physical structure in Cartesian task space, to a fully model-informed structured approach. Mean success rate increases consistently across this spectrum, from 5.7\% at the lowest evaluated level to 37.4\% (PPO) and 40.2\% (SAC) at the highest model-informed deep\,RL level, with a per-task upper bound of 46.2\% when selecting the better algorithm per task, while the fully structured approach achieves 100\% success across all 22 tasks at a mean execution speed of 1.02\,Ex\,s$^{-1}$, learned directly on real hardware in under one hour per task without simulation.

The resulting boundary is representational rather than algorithm-specific: reliable performance requires embodiment, interaction, and manipulation-process structure to be present in the learning representation, whether supplied explicitly or acquired by a mechanism capable of identifying it.

We further validate the approach across four real-world industrial assembly workflows spanning multiple manipulation classes in collaboration with leading manufacturing companies. All leading Frontier models for robot manipulation have so far been demonstrated predominantly on pick-and-place, household and deformable-object tasks at centimetre-scale tolerances, without reported reliable sub-millimetre insertion under physical real-world interaction effects such as clamping, interference fit or friction. To our knowledge, no prior work, including the publicly reported results of frontier models or industrial humanoids, has demonstrated autonomous contact-rich assembly at this level of task diversity, real-world tolerance and manufacturing relevance, especially without additional sensor arrays or task-specific robot-side tooling, i.e. customised end-effector or gripper mechanisms --- relying predominantly on the embodied intelligence and capabilities of the robotic system. As a projected dual-arm configuration scaled linearly from single-arm measurements, the system achieves 150--200\% of human daily throughput and up to 400\% of human productivity per unit labour cost under continuous operation, validated over at least 50 full workflows per assembly, with each subtask learned in under 30 minutes.

These results position contact-rich manipulation as a concrete empirical case of a broader principle for embodied intelligence: reliable physical learning depends not only on the scale of the learner, but on whether its representation contains the structure demanded by the interaction.
\end{sciabstract}


\section*{Introduction}
\subfile{./sections/introduction}

\section*{Results}
\subfile{./sections/results}

\section*{Discussion}
\subfile{./sections/discussion}

\section*{Materials and Methods}
\subfile{./sections/methods}

\newpage
\renewcommand\refname{References}
\bibliography{arxiv-bibliography}
\bibliographystyle{ieeetr}

\newpage
\section*{Acknowledgments}
\textcolor{cor2}{The authors thank the participating industrial partners for their collaboration and provision of assembly components, and KME\,--\,Kompetenzzentrum Mittelstand GmbH for their continuous and thorough support, in particular Georg~Liedl. We further thank Samuel~Schneider for his valuable support in setting up the experiments, Rolando~Franjga for assistance with the industrial validation automation, and Reinhard~Grassmann for discussion of the results.}

\section*{Funding}
\textcolor{cor}{The authors gratefully acknowledge the funding of this work by the Alfried Krupp von Bohlen und Halbach Foundation, the Deutsche Forschungsgemeinschaft through the Gottfried Wilhelm Leibniz Programme (award to S. Haddadin; grant no. HA7372/3-1) and the KME – Kompetenzzentrum Mittelstand GmbH.}

\section*{Author contributions}
F.~Voigt and S.~Haddadin designed the study. 
F.~Voigt implemented and conducted the experiments. 
S.~Haddadin conceived the scientific questions and hypotheses, developed the foundations of the models-matter approach, and provided continuous scientific direction.
F.~Voigt and S.~Haddadin interpreted the results.
F.~Voigt and S.~Haddadin wrote the manuscript.
All authors read and approved the final manuscript.

\section*{Competing interests}
S.~Haddadin has a potential conflict of interest as the
\textcolor{cor}{inventor of the Franka Emika robot technology.}

\section*{AI-Assisted Editing and Visualization}
OpenAI ChatGPT was used during manuscript preparation as an editorial assistant. Its use was limited to English-language refinement, formulation, stylistic editing, restructuring, and streamlining of author-developed text, as well as assistance in the iterative graphical realization of selected author-created schematic figures. All scientific concepts, technical content, figure concepts and information content, data, analyses, interpretations, and conclusions were developed by the authors. AI-assisted graphical outputs were based on author-defined scientific content and were subsequently reviewed and revised by the authors. No AI-based tool was used to generate scientific results or determine the scientific conclusions of this work. The authors take full responsibility for the final manuscript.

\section*{Data and materials availability}
Analysis scripts, core data, and representative videos supporting the findings of this work are privately available at \url{https://figshare.com/s/67e85ec252232281e33d}~\cite{voigt2026data}. 
Interactive visualisations and video demonstrations are available at \url{https://roboticactionelmo.github.io/models_matter/}~\cite{mm_website}. Industrial assembly data are provided with the consent of the respective industry partners. Raw learning data and full trajectory recordings, exceeding 10\,TB in total, are available from the corresponding author upon reasonable request.

\newpage
\beginsupplement

\newpage
\subfile{sections/supplementary}

\end{document}

%% file: sections/introduction.tex
A central question in embodied intelligence is what structure a learning system must represent to acquire reliable physical capability. Much of modern machine learning seeks to replace hand-specified structure with scale --- larger models, more data, and richer pretraining. This strategy is powerful when the information required for the task is accessible in the observation stream. Physical interaction creates a distinct regime: task-critical quantities such as deformation, friction, clamping, interference fit, and force transmission are only indirectly or not reliably observable through standard robot interfaces. Contact-rich manipulation therefore provides an experimentally tractable setting in which to ask a broader representational question: can scale compensate for missing physical structure, or must that structure be represented in the learning system?

We address this question through contact-rich manipulation --- the reliable execution of tasks involving tight tolerances, constrained force exchange, and significant physical interaction effects. Despite substantial progress over the past decade, learning-based approaches have not translated into reliable practice for tactilely challenging tasks of this kind. We show that this gap reflects a strong dependence of learning on physical interaction structure, that is, the embodiment of the robot and the physical effects governing contact, and provide systematic empirical evidence across 22 tasks and five physical interaction-effect regimes.

Recent advances in robot learning have been driven primarily by deep reinforcement learning (DRL), using algorithms such as Proximal Policy Optimisation (PPO)~\cite{schulman2017proximal}, Soft Actor-Critic (SAC)~\cite{haarnoja2018soft}, and Deep Deterministic Policy Gradient (DDPG)~\cite{lillicrap2015continuous}, as well as imitation learning approaches such as Diffusion Policy~\cite{chi2023diffusion}. These approaches have demonstrated strong performance on unconstrained tasks such as pick-and-place and coarse object relocation, where performance depends primarily on spatial reasoning and visual perception. Their effectiveness on tactilely challenging, contact-rich tasks, however, remains an active and incompletely characterised research challenge~\cite{kroemer2021review}. The
emerging trajectory toward large behaviour models and
vision-language-action models~\cite{barreiros2026lbm, black2025pi0, brohan2023rt2} inherits the same structural limitation: these approaches assume that scale, diverse pretraining data, and richer visual
representations can substitute for explicit physical structure, an assumption that has been the subject of active scientific
debate~\cite{amato2025debate}. 
This makes contact-rich manipulation more than a domain benchmark: it provides a physical test of a general assumption in contemporary machine learning --- whether missing task structure can be compensated by scale while the underlying representation remains unchanged.

\begin{sidewaysfigure}[t]
    \centering
    \includegraphics[width=1.00\textwidth]{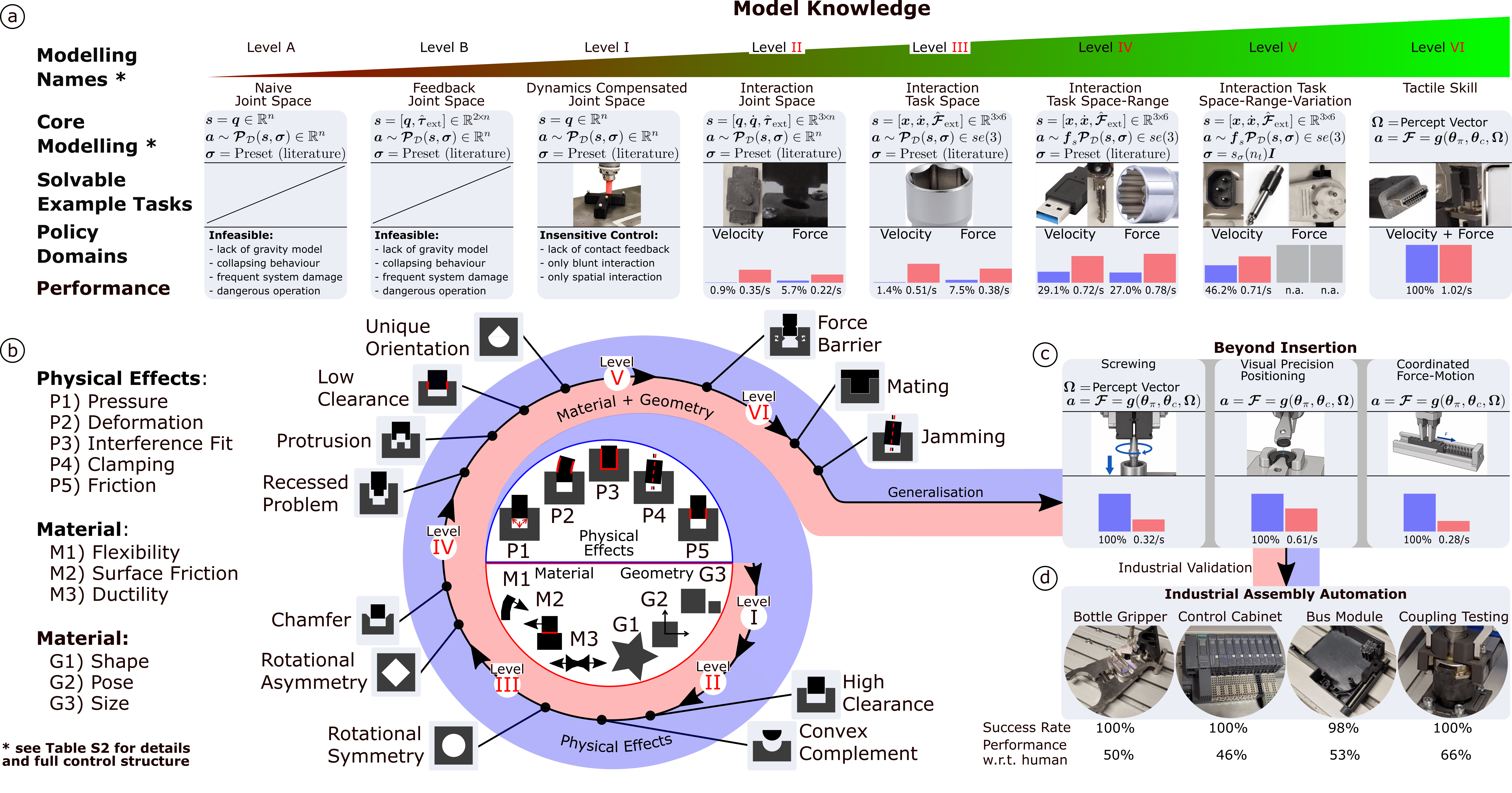}
    \captionsetup{margin={0.0cm, -0cm}}
    \caption{%
    \linespread{0.85}\selectfont
\textbf{(a)}~\textbf{Model knowledge directly increases
performance.} Multiple levels of model knowledge are defined
in increasing order, grounded in principled mechanical system and manipulation
models not previously studied in
this systematic form, progressing from naive joint-space
deep\,RL with no structural assumptions to the fully
model-informed Tactile Skill \textcolor{cor}{model}; five are
experimentally evaluated (\textsc{II}--\textsc{VI}).
Level~\textsc{A} operates without external joint torque
feedback, while Level~\textsc{B} includes it in the form of
generalised external joint torques $\boldsymbol{\tau}_\mathrm{ext}$.
All levels from Level~\textsc{I} onward include model-based
gravity and Coriolis compensation --- a property shared by all
real-world manipulation learning approaches using
torque-controlled systems, as purely end-to-end learning is
infeasible in practice. Intermediate levels incorporate
interaction joint space (\textsc{II}), interaction task space
(\textsc{III}), action-range scaling (\textsc{IV}), and
adaptive variance reduction (\textsc{V}), evaluated in
velocity-based impedance control and feedforward force
domains. $\boldsymbol{\Theta}_\pi$ and
$\boldsymbol{\Theta}_c$ denote the policy and control
parameters of the Tactile Skill (\textsc{VI}), respectively.
Full level definitions are given in Supplementary
Table~\ref{tab:model_tree}. Blue bars indicate success rate;
red bars mean task completion speed (s$^{-1}$).
\textbf{(b)}~\textbf{Physical interaction taxonomy.}
Contact-rich tasks are governed by three fundamental layers
--- geometry, material properties, and physical interaction
effects. Characteristics are arranged along an axis of
increasing task difficulty; Roman numerals indicate the
minimum model knowledge level required to address each.
\textbf{(c)}~\textbf{Generalisation.} Extension of the
Tactile Skill \textcolor{cor}{model} from insertion to
screwing, visually guided precise positioning, and coordinated
force--motion --- among the most prevalent skill classes in
tactile manufacturing.
\textbf{(d)}~\textbf{Industrial validation.} Robustness and dual-arm
productivity relative to a human worker across four full real-world
industry partner validations spanning gripper assembly, bus module
assembly, control cabinet assembly, and hydraulic coupling testing,
indicating that such a learning framework can translate to industrial
practice without additional sensor arrays or task-specific robot-side
tooling (aside from 3D-printed workpiece holders), at industrially
relevant productivity: setup in 15 minutes to 2 days depending on
workflow complexity, subtask learning in 5--30 minutes, execution in
1--6\,s per individual subtask operation (longer for multi-turn
screwing requiring successive regrasps), across a broad range of
manipulation skills and tasks.
    }
    \label{fig:main}
\end{sidewaysfigure}
\FloatBarrier

\textcolor{cor}{Recent industrial claims of production-ready performance on
connector insertion~\cite{sanctuary2026wireplugging} illustrate
this debate, yet involve a custom end-effector designed for
the specific task and a target socket that moves passively and
freely on a conveyor --- conditions under which physical
interaction effects such as clamping or interference are
minimal. This is furthermore demonstrated on only a single
task in a highly optimised environment, limiting its relevance
as a counterexample to the findings presented here.} These findings provide systematic empirical
evidence indicating that,
within standard unstructured learning \textcolor{cor}{models}, relevant physical
interaction structure cannot be reliably inferred within practical sample budgets.

Furthermore, they suggest that reliable and physically safe learning becomes unlikely to be achieved within practical experimental budgets under such unstructured \textcolor{cor}{models}. This interpretation is supported by the repeated destruction of real components observed under unstructured learning
and failure to achieve reliable performance even after
extended training.

The observed limitations of standard learning frameworks on contact-rich tasks cannot be explained solely by data quantity or model capacity. Direct visual observation
cannot \textcolor{cor}{observe nor} resolve the relevant \textcolor{cor}{underlying} physics: tasks requiring precise force
exchange, including insertion, screwing, pressing, and
deformation-sensitive assembly, are governed by physical interaction
effects including pressure, deformation, interference fit, clamping, and
friction, none of which leave a reliable visual
signature~\cite{kroemer2021review}. Simulation cannot substitute for
real physical interaction: even state-of-the-art contact simulation
frameworks explicitly rely on rigid-body approximations that cannot
capture deformation, material elasticity, or surface friction variability,
and the resulting sim-to-real gap persists despite substantial progress in
simulation fidelity~\cite{narang2022factory, chebotar2019closing,
andrychowicz2020learning, todorov2012mujoco}. \textcolor{cor2}{Most critically, deep\,RL approaches typically restrict both observations and actions to standard manipulator interfaces --- joint positions, joint velocities, and, where available, estimated external joint torques. To our knowledge, no other platform offers characteristics comparable to those of the Franka Emika robot, the de facto reference platform in robot manipulation research and the most widely available torque-controlled manipulator in academic laboratories.} These are high-level signals derived from the robot's internal
models --- among them the inertia matrix $\hat{\boldsymbol{M}}(\boldsymbol{q})$,
required for external torque estimation and Coriolis
compensation, and the gravity vector $\hat{\boldsymbol{g}}(\boldsymbol{q})$,
required for gravity compensation --- already abstracting over
the robot's raw physical behaviour (see Supplementary
Table~\ref{tab:model_tree} for a detailed account). \textcolor{cor2}{A learning system operating through such interfaces receives
model-processed state summaries, but the robot's embodiment
--- its physical structure, dynamics, impedance
characteristics, and force transmission properties governing
contact interaction --- remains structurally invisible to it.
This does by no means imply that embodiment models are unnecessary;
on the contrary, contemporary system models must
exist first to produce these interfaces.}

Prior work has explored structured skill formalisms combining impedance
control with meta-parameter learning~\cite{johannsmeier2019framework},
sim-to-real transfer with compliance tuning~\cite{tang2023industreal},
hybrid imitation and reinforcement learning for insertion tasks with
clearances up to 1\,mm~\cite{wang2022adaptive}, and model-based variable
impedance learning~\cite{anand2023modelbased}. Despite these advances,
reliable autonomous execution across diverse real-world geometries and
physical regimes remains an open problem, with existing approaches
typically evaluated on just a small number of tasks under controlled
conditions and frequently relying on simulation or human demonstrations
as prerequisites. The closest prior work in terms of scale and scope,
IndustReal~\cite{tang2023industreal}, achieves 83\% success on
geometrically simple 3D-printed pegs and holes with clearances of 0.5
to 0.6\,mm, but explicitly fails on electronic connector insertion,
requires thousands of GPU-hours of simulation, and does not address complex physical interaction regimes present in real-world components.
\textcolor{cor}{A further and underappreciated consequence of applying end-to-end
deep RL to real hardware under significant physical interaction
effects is component destruction: without the correct physical
structure encoded in the policy, the robot applies forces without
the necessary precision and directionality, resulting in systematic
component damage across failed training attempts --- an outcome
that reflects not sample inefficiency but a fundamental mismatch
between the force demands of the task and the physical capabilities
of an unstructured policy. Basic physical structure such as
impedance control~\cite{hogan1985impedance}, inspired by the
compliance mechanisms underlying safe human--environment
interaction, provides a principled foundation for this; learning
approaches building on impedance control have demonstrated improved
safety and performance in contact-rich
tasks~\cite{anand2023modelbased,buchli2011learning}.
Task-level structure interfacing to learning through explicit
parametrisation has been proposed
by~\cite{johannsmeier2019framework}.}

These works support the view that physical structure is beneficial and in some cases necessary to encode explicitly, albeit they are conducted on limited task sets and do not provide systematic evidence across physical interaction regimes. \textcolor{cor}{Here, physical structure refers to embodiment, interaction and process structure, including the task-specific properties that parameterise the manipulation process.} Sim-to-real approaches for contact
tasks~\cite{narang2022factory,chebotar2019closing,andrychowicz2020learning,tang2023industreal} reduce real-world data requirements by training in simulation, but the residual sim-to-real gap for deformation, friction variability, and material compliance remains a documented and unsolved problem~\cite{todorov2012mujoco}. Imitation learning methods~\cite{chi2023diffusion,zhao2023act}
have demonstrated strong performance on visually rich and bimanual
manipulation tasks, yet to date have not reported reliable performance on sub-millimetre-clearance insertion of real industrial components under significant physical interaction effects. Large behaviour models and vision-language-action
systems~\cite{barreiros2026lbm,black2025pi0,brohan2023rt2} represent the frontier of scale-based approaches; their contact-task performance is evaluated in~\cite{barreiros2026lbm}, which explicitly notes limitations on tasks requiring precise force
exchange. We refer to these large vision-language-action and behaviour models pretrained at scale as frontier models for robot manipulation. Publicly reported results of such models from 2025--2026 are consistent with this assessment: success rates above 99\,\% have been reported only for narrow, repeated tasks at centimetre-scale tolerance, such as laundry and napkin folding~\cite{sunday2026act2,dyna2025dyna1};
generalisation to unseen homes reaches 56\,\% on household
tasks~\cite{figure2026helix25}; and the most detailed public per-task evaluation reports 89.6\,\% averaged over insertion tasks without stating clearances~\cite{deepmind2026gr2}. None of these reports addresses sub-millimetre insertion under significant real-world physical interaction effects, which is absolutely crucial for physical AI to enter the real world. To our knowledge, the present work provides the first systematic large-scale empirical study of this representational question in real physical interaction, across 22 tasks, five physical interaction regimes, and a spectrum of learning approaches from increasingly model-informed deep\,RL to a fully structured model-informed framework. The benchmark is used as a controlled experimental probe of how representation determines learnability under real physical interaction, rather than as an application-specific comparison of insertion methods.

We define \emph{model knowledge} as the extent to which known physical structure and task-specific properties are encoded in the full learning representation available to the learning algorithm, and we distinguish three layers. \emph{Embodiment knowledge} describes the physical structure and
dynamics of the robotic system (kinematics, inertia, Jacobian, and gravity, Coriolis and friction compensation). \emph{Interaction-control knowledge} describes the relationship between robot motion and exchanged effort (joint- or
task-space representation, effort and flow variables, impedance control, physically meaningful action ranges). \emph{Process knowledge} describes the temporal and physical organisation of the manipulation task itself (approach, contact, compliant alignment, insertion, phase transitions, feedforward force),
together with the task-specific quantities that parameterise it (Supplementary Information table~\ref{tab:model-knowledge-layers}). These layers are properties of the learning representation and are distinct from the search/learning mechanism (PPO, SAC, DDPG, or the Tactile Skill optimiser),
which identifies the task-dependent parameters that remain once the structure is fixed. Our central claim is therefore representational rather than algorithm-specific: reliable contact-rich manipulation learning requires the embodiment, interaction, and process layers to be represented --- whether supplied explicitly or acquired by a mechanism capable of identifying them --- while changing or scaling the search mechanism alone cannot supply structure that is absent from the representation. We show that the degree of model knowledge encoded in a learning framework, specifically the extent to which it encodes robot dynamics, impedance structure, and physical interaction effects, is a primary determinant of learning performance across all tested levels and a necessary condition for achieving reliable execution in contact-rich
manipulation. We demonstrate this through a two-year systematic study across 22 contact-rich insertion tasks spanning real electrical connectors, mechanical sockets, keys, and plugs with clearances down to sub-0.1 mm, evaluating a spectrum of learning approaches from unstructured deep RL to a fully structured Tactile Skill model (Fig.~\ref{fig:main}). We further characterise how five physical interaction effects (pressure, deformation, interference fit, clamping, and friction; P1–P5) determine task difficulty across all learning levels, and validate the Tactile Skill model across four industrial real-world assembly workflows spanning multiple manipulation classes in collaboration with leading manufacturing companies. Although the evaluated task set spans diverse geometries and interaction regimes, it does not exhaust the full range of robotic embodiments, materials, or assembly conditions encountered in industrial practice.

We show that increasing model knowledge produces a consistent
positive trend in learning performance on average across all
22 tasks, algorithms, and control domains. The Interaction
Joint Space model (Level~\textsc{II}) achieves a mean success
rate of 5.7\% in the force domain and 0.9\% in the velocity
domain, rising to 7.5\% and 1.4\% for the Interaction Task
Space model (Level~\textsc{III}), and to 27.0\% and 29.1\%
for the Interaction Task Space -- Range model
(Level~\textsc{IV}). Model-informed deep\,RL at the highest
examined level, the Interaction Task Space -- Range --
Variation model (Level~\textsc{V}, velocity domain), achieves
a mean success rate of 46.2\% \textcolor{cor}{at a mean
execution speed of 0.71\,Ex\,s$^{-1}$}, failing to reach
reliable performance across the majority of tasks.
When deep\,RL does succeed,
execution speeds across levels range from
0.81--1.08\,Ex\,s$^{-1}$ (adjusted mean over successful
trials), comparable to the Tactile Skill, indicating that
the fundamental limitation is categorical failure rather than
slow execution. A structured model encoding
robot dynamics, impedance control, and feedforward force
achieves 100\% success across all tasks at a
mean execution speed of 1.02\,Ex\,s$^{-1}$
(0.74--1.26\,Ex\,s$^{-1}$ across tasks), learned directly
on real hardware in under one hour per task without
simulation, under the assumption of a known goal pose ---
which in practice can be obtained from a single visual
teach-in or standard vision system.

Applied to industrial assembly, the model achieves 150 to 200\% of human daily throughput projected for a dual-arm system, and up to 400\% of human productivity per unit labour cost, indicating that explicitly encoding physical interaction structure is a necessary condition for reliable and economically viable contact-rich manipulation rather than an optional optimisation.

The main contributions of this work are:

\begin{itemize}

    \item A systematic large-scale empirical test of a representational hypothesis in embodied learning, using 22 contact-rich tasks on 22 individual Franka robotic arms to determine how increasing embodiment, interaction, and process knowledge changes learnability across algorithms and control domains, revealing a categorical performance boundary between increasingly model-informed deep\,RL and the structured Tactile Skill model.

    \item A physical effect taxonomy for contact-rich insertion tasks characterising how pressure, deformation, interference fit, clamping, and friction determine task difficulty and govern policy behaviour across all examined levels of model knowledge, supported by regression analysis linking normalised effect magnitudes to observed performance metrics.

    \item A mechanistic analysis of learned policy landscapes based on the 33 parameters of the structured Tactile Skill model, revealing narrow, multimodal success basins within the parameter space, whose structure is likely shaped in part by the discrete phase transitions between skill execution phases.

    \item Real-world validation of the structured learning model
    across four full industrial assembly workflows spanning four manipulation classes: insertion, screwing, \textcolor{cor}{visual} precision positioning, and \textcolor{cor}{coordinated} force–motion, involving rigid, flexible, and
    spring-loaded components across workflows of up to 23 subtasks, confirming the \textcolor{cor}{model's} generality and achieving 150 to 200\% of human daily throughput operating as a dual-arm system and up to 400\% of human productivity per unit labour cost under the evaluated deployment conditions.

\end{itemize}

A natural question is whether the observed boundary reflects only the current frontier of deep\,RL or a more general representational requirement. Recent large behaviour models and vision-language-action systems~\cite{barreiros2026lbm,black2025pi0,brohan2023rt2} have demonstrated non-trivial generalisation on visually rich manipulation tasks, and we do not dispute their progress on the tasks they address. The empirical scope of our claim is deliberately specific, however, pivotal to the field of physical AI: contact-rich tasks governed by sub-millimetre clearances and physical interaction effects including deformation, interference fit, and friction. The representational implication is broader: scaling samples or policy capacity within the same structurally incomplete representation does not supply task-relevant physical structure that is absent from the information available to the learner. This does not imply that such structure cannot itself be learned; recurrent policies, system-identification methods, self-modelling, demonstrations, or simulation may provide mechanisms for acquiring it. Our claim is that reliable physical learning requires the relevant embodiment, interaction, and process structure to enter the representation somewhere. The evidence provided here tests this requirement systematically across 22 tasks and five physical interaction regimes.

Interactive visualisations, videos and supplementary materials are available at the project website: \url{https://roboticactionelmo.github.io/models_matter/}.

%% file: sections/results.tex
\begin{figure}[!t]
    \centering
    \includegraphics[width=1.0\textwidth]{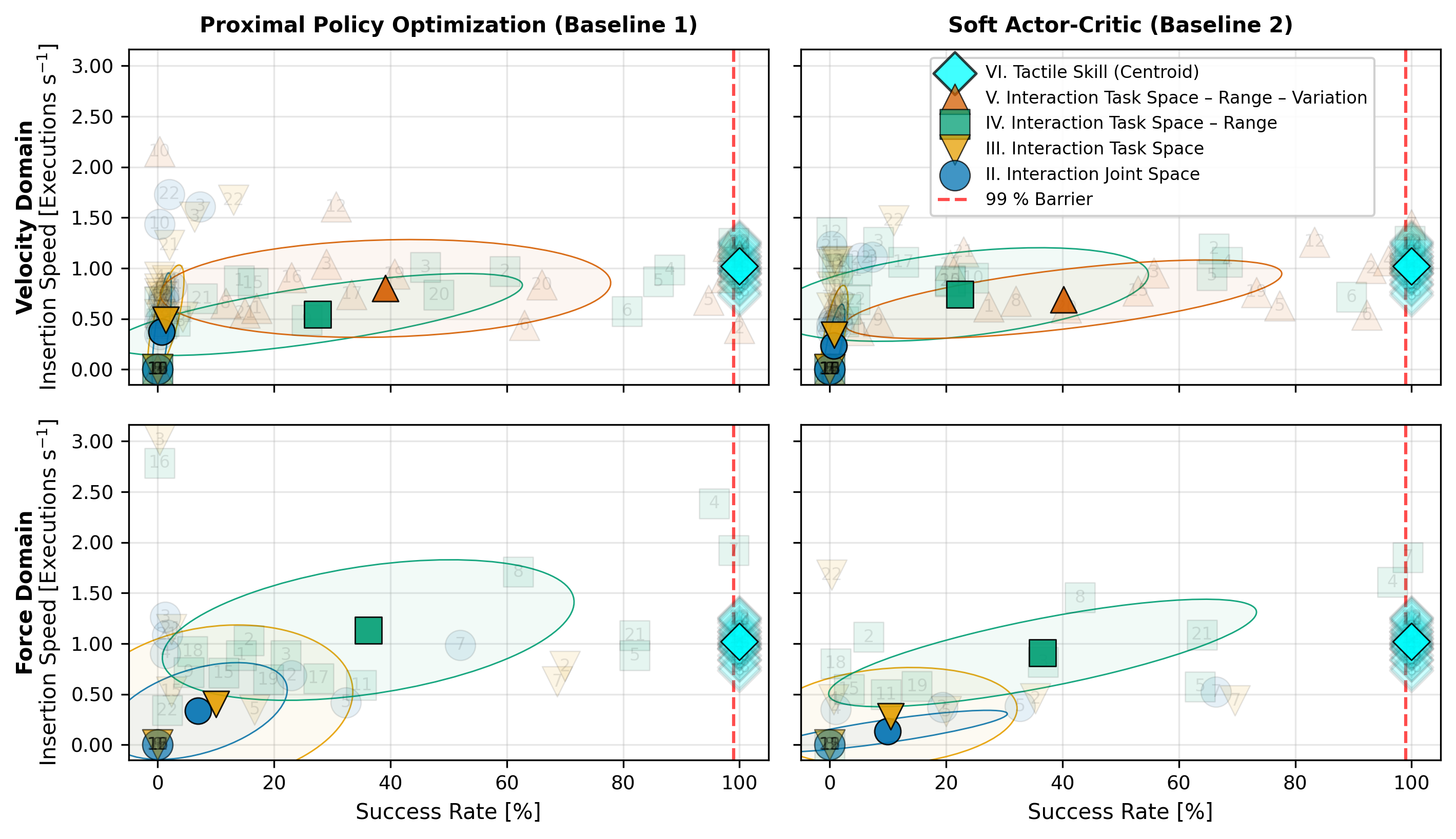}
\caption{%
\textbf{Success rate versus insertion speed across all 22
tasks, model knowledge levels, and algorithms.}
Each translucent point represents the mean success rate
(final 100 evaluation trials) and mean insertion speed
(executions\,s$^{-1}$ among successful trials) for a single
task--algorithm--level combination. Marker shape encodes
model knowledge level (\textsc{II}--\textsc{VI}); a dashed
vertical line marks the 99\% success barrier. Centroids per
level--algorithm combination are shown with $\pm1\sigma$
intervals, including entries at $(0, 0)$ for entirely
unsuccessful configurations. The distribution is strongly
bimodal: deep\,RL configurations cluster predominantly below
20\% success at lower model knowledge levels, with higher
levels achieving improved but still limited performance, while
all Tactile Skill results cluster above the 99\% barrier at
0.5--1.3\,executions\,s$^{-1}$ across all tasks. Deep\,RL
exceeds 98\% success only for geometrically simple tasks with
low physical effect magnitudes, namely the 3D-printed
ellipsoid (task~7) and one socket variant (task~2).
}
    \label{fig:performance_main}
\end{figure}

\begin{table}[htbp]
  \centering
  \caption{%
  \textbf{Task-wise success rates across model knowledge levels and control domains.}
  Values report mean success rates (\%) over the final 100 trials of each 500-trial experiment.
  For each task and level, the best-performing deep\,RL algorithm (PPO or SAC) based on mean success rate is reported.
  Columns are grouped by control domain \textcolor{cor}{(force and velocity)}, with the Tactile Skill \textcolor{cor}{model} shown separately.
  Blue values indicate success rates $\geq$90\%, and green values indicate $\geq$99\% success.
  Underlined values denote the highest deep\,RL success rate per task.
  }
  \label{tab:success_best}
  \small
  \setlength{\tabcolsep}{4pt}
  \begin{tabular}{lrrrrrrrr}
    \toprule
         & \multicolumn{3}{c}{Force Domain} & \multicolumn{4}{c}{Velocity Domain} & VI. Tactile \\
        \cmidrule(lr){2-4} \cmidrule(lr){5-8} \cmidrule(lr){9-9}
        Object [\%] & II & III & IV & II & III & IV & V & Skill \\ 
        \midrule

1. 12point-socket-24  
& 0.0 & 0.0 & 14.3  
& 0.0 & 0.0 & 14.0 & \bestdrl{27.3}  
& \textcolor{green!60!black}{100.0} \\

2. 12point-socket-36  
& 23.0 & 70.0 & 15.7  
& 0.0 & 0.0 & 66.0 & \bestdrl{\textcolor{green!60!black}{100.0}} 
& \textcolor{green!60!black}{100.0} \\

3. 6point-socket-24  
& 1.3 & 0.3 & 36.7  
& 7.3 & 6.3 & 46.0 & \bestdrl{55.7}  
& \textcolor{green!60!black}{100.0} \\

4. 6point-socket-32  
& 1.3 & 0.7 & \textcolor{blue}{96.7}  
& 0.0 & 0.3 & 88.0 & \textcolor{blue}{\bestdrl{98.0}}  
& \textcolor{green!60!black}{100.0} \\

5. 6point-socket-41  
& 32.7 & 20.0 & 82.0  
& 0.0 & 0.3 & 86.0 & \textcolor{blue}{\bestdrl{94.7}}  
& \textcolor{green!60!black}{100.0} \\

6. abus-e30  
& 0.0 & 0.0 & 0.0  
& 0.0 & 0.3 & 89.7 & \textcolor{blue}{\bestdrl{92.3}}  
& \textcolor{green!60!black}{100.0} \\

7. ellipsoid  
& 66.3 & 69.7 & \textcolor{green!60!black}{99.3}  
& 1.0 & 0.0 & \textcolor{green!60!black}{99.7}  
& \textcolor{green!60!black}{\bestdrl{100.0}}  
& \textcolor{green!60!black}{100.0} \\

8. europlug  
& 0.0 & 0.0 & \bestdrl{62.0}  
& 0.0 & 0.0 & 0.0 & 32.0  
& \textcolor{green!60!black}{100.0} \\

9. gb-2099  
& 0.0 & 0.0 & 5.3  
& 0.0 & 0.0 & \bestdrl{25.7} & 8.3  
& \textcolor{green!60!black}{100.0} \\

10. hdmi  
& 0.0 & 0.0 & 0.0  
& 0.3 & 0.7 & \bestdrl{24.7} & 0.7  
& \textcolor{green!60!black}{100.0} \\

11. hexkey-12  
& 0.0 & 0.0 & \bestdrl{35.0}  
& 1.3 & 4.3 & 0.0 & 5.0  
& \textcolor{green!60!black}{100.0} \\

12. hexkey-17  
& 0.0 & 0.0 & 0.0  
& 0.3 & 1.0 & 0.3 & \bestdrl{83.3}  
& \textcolor{green!60!black}{100.0} \\

13. hexkey-19  
& 0.0 & 0.0 & 0.0  
& 0.0 & 0.0 & \bestdrl{0.7} & 0.0  
& \textcolor{green!60!black}{100.0} \\

14. hexkey-6  
& 0.0 & 0.0 & 0.0  
& \bestdrl{2.3} & 1.7 & 0.0 & 0.0  
& \textcolor{green!60!black}{100.0} \\

15. iec-c13  
& 0.0 & 2.3 & 11.3  
& 0.0 & 0.3 & 20.7 & \textcolor{blue}{\bestdrl{97.7}}  
& \textcolor{green!60!black}{100.0} \\

16. padlock-burgwächter  
& 0.0 & 0.0 & 0.3  
& 0.0 & 0.0 & 2.3 & \bestdrl{23.0}  
& \textcolor{green!60!black}{100.0} \\

17. schuko  
& 0.0 & 0.0 & 27.7  
& 0.0 & 0.0 & 12.7 & \bestdrl{33.3}  
& \textcolor{green!60!black}{100.0} \\

18. schuko 2  
& 0.0 & 0.0 & \bestdrl{6.0}  
& 0.0 & 0.0 & 3.0 & 1.7  
& \textcolor{green!60!black}{100.0} \\

19. trrs-6.35mm aluminum rod  
& 0.0 & 0.0 & 19.0  
& 0.3 & 0.3 & 0.0 & \bestdrl{73.3}  
& \textcolor{green!60!black}{100.0} \\

20. type-l  
& 0.0 & 0.0 & 0.0  
& 0.0 & 0.0 & 48.3 & \bestdrl{66.0}  
& \textcolor{green!60!black}{100.0} \\

21. usb-a  
& 1.7 & 2.3 & \bestdrl{82.0}  
& 1.0 & 2.0 & 7.7 & 23.0  
& \textcolor{green!60!black}{100.0} \\

22. usb-c  
& 0.0 & 0.3 & 1.7  
& 5.3 & \bestdrl{13.0} & 4.3 & 1.0  
& \textcolor{green!60!black}{100.0} \\

\midrule

\textbf{Mean}
& \textbf{5.7}
& \textbf{7.5}
& \textbf{27.0}
& \textbf{0.9}
& \textbf{1.4}
& \textbf{29.1}
& \textbf{46.2}
& \textbf{100.0} \\

\bottomrule
\end{tabular}
\end{table}

Across all 22 insertion tasks, both PPO and SAC, and both the
velocity domain and force domain, increasing model knowledge
produces a consistent positive trend on average in learning performance
(Fig.~\ref{fig:performance_main}). Both deep\,RL algorithms achieve a mean success rate of 37.4\%
(PPO) and 40.2\% (SAC) at the highest evaluated model knowledge
level in the velocity domain (Level\,\textsc{V}, Interaction Task Space -- Range -- Variation)\textcolor{cor}{, at a mean execution speed of 0.76\,Ex\,s$^{-1}$ (PPO) and
0.69\,Ex\,s$^{-1}$ (SAC)}, with the best-performing algorithm
per task achieving on average 46.2\% \textcolor{cor}{ at
0.71\,Ex\,s$^{-1}$} across all 22 tasks.
In the force domain at Level\,\textsc{IV} (Interaction Task Space -- Range) --- the highest safely
evaluable force level, as discussed below --- PPO achieves 26.3\%
and SAC 20.0\%\textcolor{cor}{, at a mean execution speed of
0.83\,Ex\,s$^{-1}$ and 0.50\,Ex\,s$^{-1}$ respectively}, with
a per-task best of 27.0\% \textcolor{cor}{ at
0.78\,Ex\,s$^{-1}$}.
\textcolor{cor2}{The Tactile Skill model solves all 22 tasks,
with only 4 of 211 completed learning runs failing to converge, all
on the two hardest objects (9 of the 220 runs were incomplete and
therefore removed from the dataset; tasks-solved criterion, see
Methods),} at a mean execution speed of
\textcolor{cor}{1.02\,Ex\,s$^{-1}$}. The performance boundary
between the deep\,RL regime and the Tactile Skill regime is
therefore categorical rather than gradual.

Figure~\ref{fig:performance_main} shows success rate against insertion
speed for all tasks and levels across PPO and SAC in both control
domains. The distribution is strongly bimodal. The large majority of
deep\,RL configurations cluster at success rates below 20\%, with
insertion speeds near zero, reflecting policies that fail to achieve
reliable insertion within the 500-trial budget. All Tactile Skill
results cluster tightly behind the 99\% barrier, achieving both high
success rates and insertion speeds of 0.5--1.3\,executions\,s$^{-1}$
across all tasks. The intermediate range between 30\% and 95\% success
is \textcolor{cor}{sparsely populated}, indicating that partial learning is rare and
unstable: policies \textcolor{cor}{predominately} fail categorically or succeed reliably.

Within the deep\,RL results, mean success rate increases with
model knowledge level across both algorithms and both control
domains, though individual tasks exhibit level-specific
reversals in 11~cases with drops exceeding 5\%. These
reversals may be attributable to the limited number of
independent runs per configuration --- three per task and
model level combination, reflecting the practical constraints
of real-hardware experimentation --- and to the
exploitation-promoting effect of adaptive variance reduction
at Level\,\textsc{V}, which can cause the policy to converge
to local optima on tasks where broader exploration at
Level\,\textsc{IV} was more productive. While three
independent runs per configuration is below what would be
needed to fully characterise rare success events, the
variance structure across the 22-task benchmark is
sufficiently consistent to support the aggregate trends
reported here. Real-world physical variability --- including
temperature fluctuations during operation and microscale
wear on contact surfaces --- may further contribute to
run-to-run variance in ways that are not fully controllable
in a real-hardware study of this scale. Nevertheless, the
aggregate findings across 308 task–algorithm–level/domain configurations (22 tasks
$\times$ 2 algorithms $\times$ 7 levels) are consistent in
direction, supporting the robustness of the reported trends.
Level\,\textsc{II} (Interaction Joint Space) achieves near-zero success
on the large majority of tasks in both domains.
Level\,\textsc{III} (Interaction Task Space) shows marginal
improvement.
Levels\,\textsc{IV} and \textsc{V}, incorporating action scaling and
adaptive variance reduction respectively, produce the highest deep\,RL
success rates but remain far below reliable insertion on the majority
of tasks.
At the aggregate level, the velocity (V) and force (F) domains perform
comparably at Level\,\textsc{IV} (column means of 26.2\%
\textcolor{cor}{at 0.52\,Ex\,s$^{-1}$} velocity versus 26.3\%
\textcolor{cor}{at 0.83\,Ex\,s$^{-1}$} force for PPO), with
the velocity domain pulling ahead only at Level\,\textsc{V}
where no force domain equivalent was safely evaluable. At the
individual task level, however, the domains diverge
substantially: tasks such as \textcolor{cor}{\emph{6pt-socket-32}}
(F-IV: 96.7\%\textcolor{cor}{, 1.61\,Ex\,s$^{-1}$}),
\textcolor{cor}{\emph{USB-A}}
(F-IV: 82.0\%\textcolor{cor}{, 1.08\,Ex\,s$^{-1}$}), and
\textcolor{cor}{\emph{europlug}}
(F-IV: 62.0\%\textcolor{cor}{, 1.71\,Ex\,s$^{-1}$}) achieve
markedly higher success in the force domain, while tasks such
as \textcolor{cor}{\emph{abus-e30}}
(V-IV: 89.7\%\textcolor{cor}{, 0.72\,Ex\,s$^{-1}$}),
\textcolor{cor}{\emph{type-l}} (V-IV: 48.3\%\textcolor{cor}{, 0.74\,Ex\,s$^{-1}$}),
and \textcolor{cor}{\emph{hdmi}}
(V-IV: 24.7\%\textcolor{cor}{, 0.90\,Ex\,s$^{-1}$}) show the
reverse pattern. This task-level divergence reflects that the
appropriate control modality is determined by the specific
physical interaction effects present: tasks requiring large
precisely controlled axial forces benefit from direct wrench
control, while tasks requiring velocity-regulated approach
trajectories benefit from impedance-based velocity control.

The per-task results (Tab.~\ref{tab:success_best}) reveal that
deep\,RL performance is highly task-dependent. At
Level\,\textsc{V} velocity and Level\,\textsc{IV} force, several tasks
approach or exceed the 90\% success threshold in the best-performing
algorithm: \textcolor{cor}{\emph{12pt-socket-36}} (100.0\%), \textcolor{cor}{\emph{iec-c13}} (97.7\%), \textcolor{cor}{\emph{6pt-socket-32}}
(98.0\%), and \textcolor{cor}{\emph{6pt-socket-41}} (94.7\%) in the velocity domain, and
\textcolor{cor}{\emph{6pt-socket-32}} (96.7\%) and \textcolor{cor}{\emph{Ellipsoid}} (99.3\%) in the force domain.
Task~7 (\emph{Ellipsoid}) is the only task where both PPO and SAC
consistently approach the 99\% barrier across both control domains
and both evaluated levels, reaching 99.0--99.7\% at
Level\,\textsc{IV} and 100\% at Level\,\textsc{V}, consistent with
its geometrically simple, rotationally symmetric insertion profile and
uniformly low physical effect magnitudes. Tasks with high physical effect complexity, including the
\textcolor{cor}{\emph{USB-C}} connector (22), one of the
\textcolor{cor}{\emph{Schuko}} power outlets (18), and small hex keys
(13, 14), remain at low success rates across all deep\,RL
configurations, with a maximum of 13\% for these tasks. The Tactile Skill achieves 100\% success across all 22 tasks
without exception, with each task learned directly on real
hardware in under one hour. Task numbers correspond to objects shown in
Fig.~\ref{fig:collective_objects} in Methods.

\subsection*{Limits of Deep Reinforcement Learning: Clearance and
Geometry}
\phantomsection
\label{sec:limits_of_DRL}

\begin{figure}[!t]
    \centering
    \includegraphics[width=1.0\textwidth]{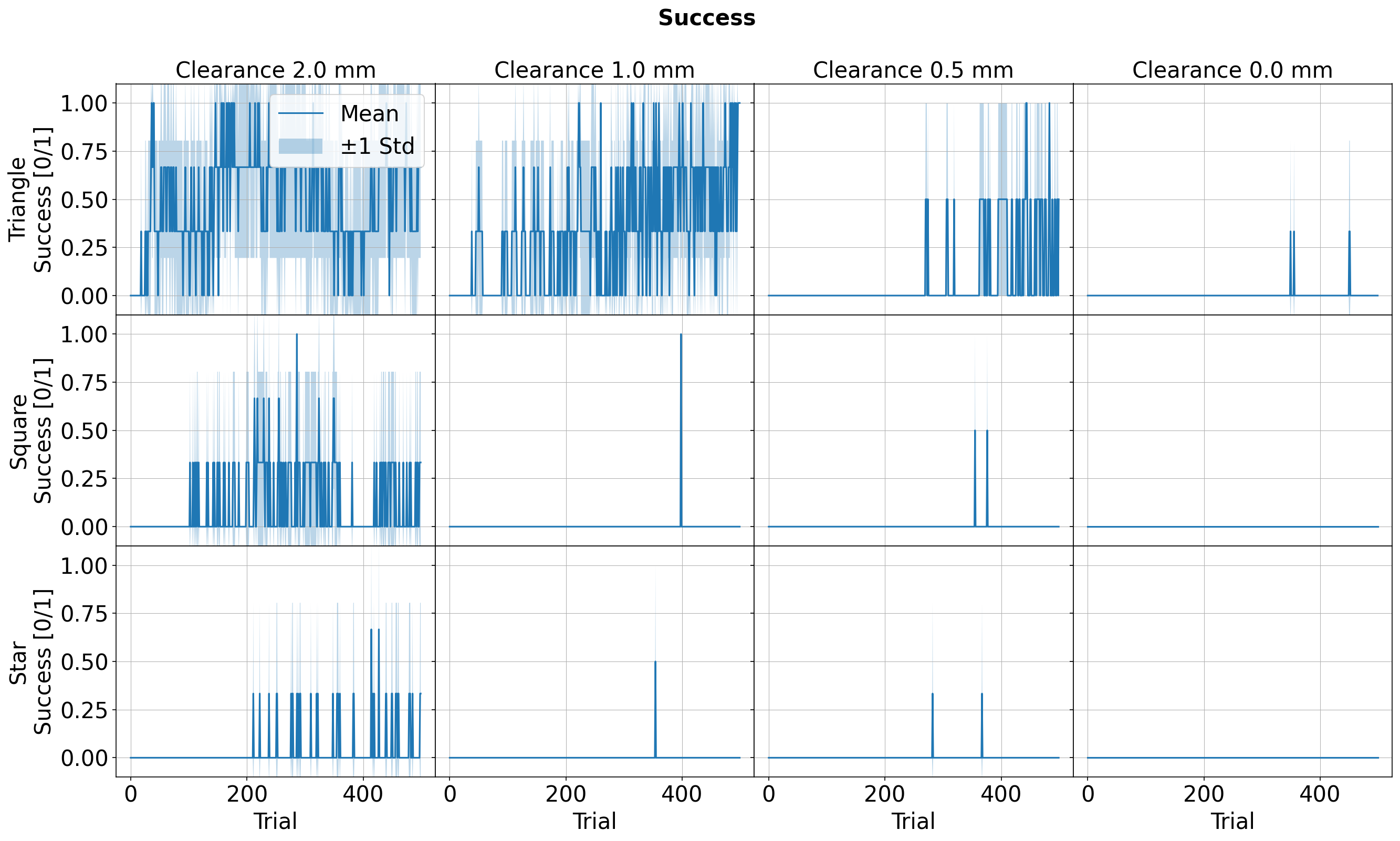}
\caption{%
\textbf{Per-trial success rate for three peg-and-hole cross-sections
(triangular, square, star) at four nominal clearance levels
(2.0, 1.0, 0.5, 0.0\,mm).}
Each panel shows mean success rate $\pm$ standard deviation across three independent runs of SAC in the Interaction Joint Space velocity domain over 500 trials.
At 2.0\,mm clearance, the triangular cross-section exhibits intermittent success without convergence, while the square and star geometries show only isolated successes.
As clearance decreases, performance consistently degrades, with a reduction in successful trials across all geometries.
For the triangular case, a downward trend is visible; for square and star geometries, success drops to zero beyond 2.0\,mm except for rare outliers.
These results indicate that the performance of generic
end-to-end deep\,RL systematically degrades on contact-rich
insertion once clearance falls below a critical threshold ---
for this experiment approximately 1.0\,mm --- even for
simple, purpose-manufactured geometries with limited material
effects, with Level~\textsc{II} SAC serving as a
representative instantiation of this class of approaches.
}
    \label{fig:drl_cg_success}
\end{figure}

\textcolor{cor}{To isolate the contribution of geometric difficulty from physical
interaction effects, a dedicated experiment was conducted using
purpose-manufactured peg-and-hole parts with three controlled
cross-sectional shapes --- triangular, square, and star --- at four
nominal clearance levels: 0.0, 0.5, 1.0, and 2.0\,mm
(Fig.~\ref{fig:drl_cg_success}). Parts were fabricated using a
Bambu Lab X1C~\cite{bambulab_x1c} printer with PLA filament, which has a
dimensional accuracy of $\pm$0.1\,mm; nominal clearances below
this tolerance should therefore be interpreted as near-zero
clearance conditions rather than precise geometric specifications.
Unlike real industrial components, these parts were designed to
exhibit minimal material effects, allowing the limits of deep RL
to be characterised as a function of geometry and clearance alone.
Experiments used the Interaction Joint Space model
(\textsc{II}), the lowest physically feasible level of model-informed
end-to-end deep RL.}

At 2.0\,mm nominal clearance, the triangular cross-section achieves
intermittent success with rates fluctuating between 0\% and 80\% across
trials without convergence, consistent with a policy that has located a
local optimum in the reward landscape but continues exploration that
repeatedly moves away from it. The square and star cross-sections show
only anecdotal successes at 2.0\,mm, hinting at general learnability
but suggesting that convergence to even a roughly robust policy would
require substantially more trials. As clearance decreases to 1.0\,mm, the square and star
cross-sections collapse immediately to near-zero success,
while the triangle shows a strongly reduced learning rate and,
despite apparent convergence to a local optimum, fails to
achieve reliable insertion. At 0.5\,mm and 0.0\,mm, the triangle shows further reduced
success rates, consistent with the trend, while the square
and star exhibit effectively zero success throughout.
Distance-to-goal trajectories
(Supplementary Fig.~\ref{fig:drl_cg_distance}) confirm no
systematic reduction in distance to goal for the square and
star at any clearance below 2.0\,mm, and only a marginal reduction for the
triangle up to 1.0\,mm, beyond which all three shapes show
trajectories consistent with exploratory noise --- no evidence of progressive skill acquisition was observed within the 500-trial campaign.

For the evaluated Level~\textsc{II} SAC formulation,
performance deteriorates sharply as nominal clearance is
reduced below approximately 1\,mm, even for controlled
purpose-manufactured geometries with limited material effects.
These findings indicate that generic end-to-end deep\,RL
fails to achieve reliable contact-rich insertion even under
the most favourable geometric conditions --- simple shapes,
controlled clearances, and limited material effects --- once
nominal clearance falls below a critical threshold, here observed at
approximately 1.0\,mm. 

The non-convex geometry of the star
cross-section, which requires precise rotational alignment of multiple
corners before insertion can succeed, and the differing numbers of
corners across shapes, produce consistently lower success rates for the
star than for the triangular and square cross-sections across all
clearance levels. The failure is therefore not attributable to the complexity of
industrial objects but reflects a fundamental limitation of
end-to-end deep\,RL under tight geometric constraints ---
\textcolor{cor}{underscoring that model knowledge is necessary for physically
meaningful manipulation learning even for reduced physical interaction
effects and when failure is driven by clearance alone.}

\subsection*{Physical Safety During Learning}

\begin{figure}[!t]
    \centering
    \includegraphics[width=1.0\textwidth]{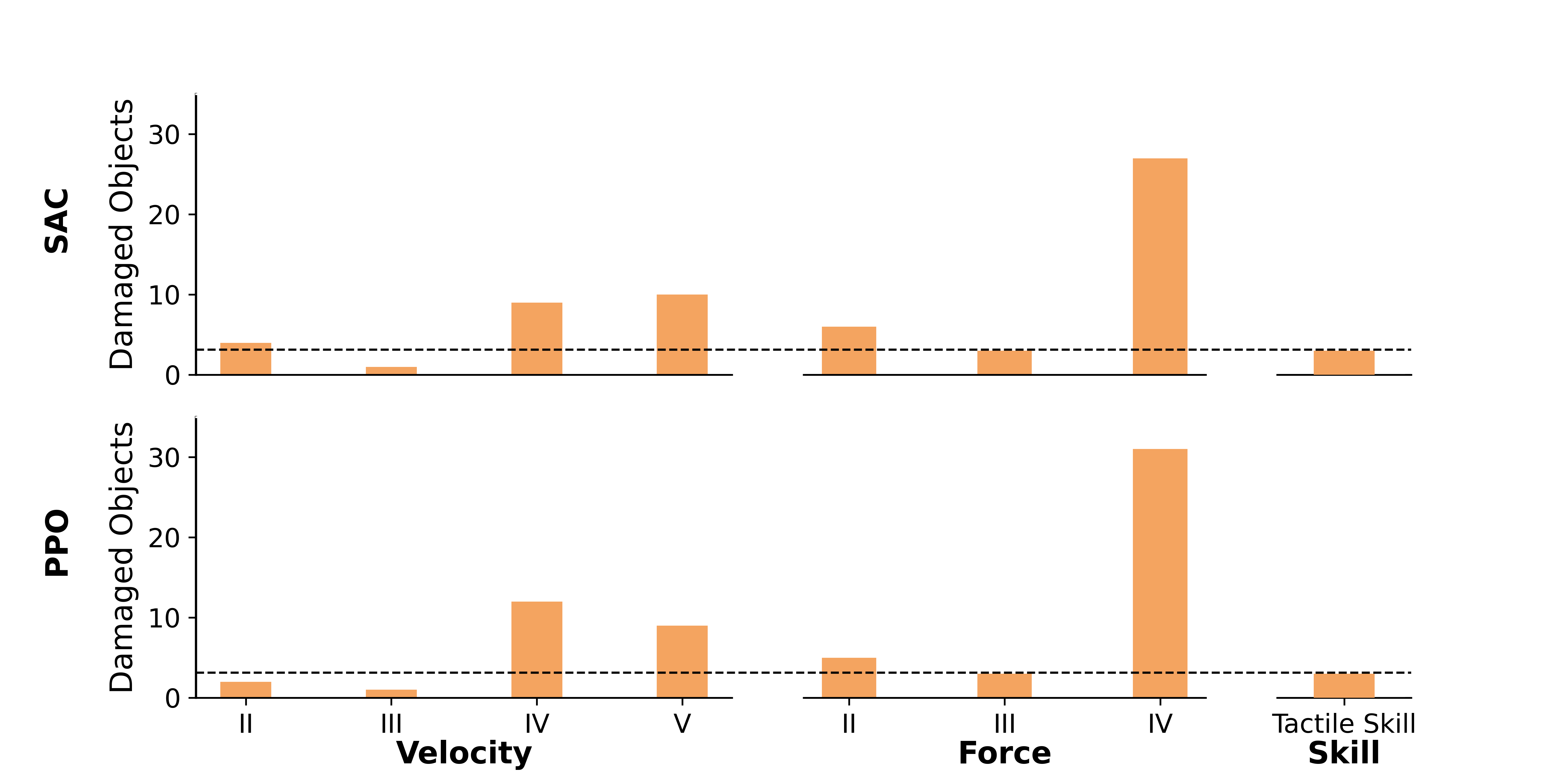}
\caption{%
  \textbf{Component damage per model knowledge level and 
  control domain across the full 22-task benchmark.}
  Bars show the number of objects damaged or destroyed 
  during the learning campaign for SAC (top) and PPO 
  (bottom) across the velocity and force control domains, 
  and for the Tactile Skill.
  Destruction rates increase sharply with action scaling: 
  the force domain at Level~\textsc{IV} produces the 
  highest damage counts of 26--31 objects across SAC and 
  PPO respectively, as scaled force commands generate 
  high-energy contact events during failed insertions.
  Force-domain experiments at Level~\textsc{V} were not 
  conducted across the full benchmark, as this level exceeded the experimental safety threshold.
  The Tactile Skill produces no damage beyond normal 
  operational wear; the dashed line indicates this 
  baseline wear level for comparison.
  These results confirm that explicit physical structure 
  improves both task performance and safety during 
  learning.%
}
    \label{fig:destroyed_statistics}
\end{figure}

A direct and underappreciated consequence of applying unstructured
deep\,RL to real \textcolor{cor}{physical systems} under significant physical interaction effects
is component destruction (Fig.~\ref{fig:destroyed_statistics}).
Destruction rates increase substantially with action scaling: the force
domain at Level\,\textsc{IV} produces the highest destruction counts,
reaching 26--31 damaged objects across SAC and PPO respectively, as
scaled force commands generate high-energy contact events during failed
insertions. Force domain experiments at Level\,\textsc{V} were not
conducted across the full benchmark as this level
produced destruction rates exceeding the safety threshold of the
experimental protocol; the highest safely evaluable force domain level
is therefore Level\,\textsc{IV}. The Tactile Skill \textcolor{cor}{model}, by
contrast, produces destruction rates at or below the background level
observed from normal wear, reflecting the impedance-based control
structure that implicitly limits undesired contact forces regardless of
the learned parameters. \textcolor{cor}{This demonstrates that increasing model knowledge alone does
not necessarily improve safety, and that principled model
design can simultaneously improve learning success and
physical safety while reducing hardware wear during the
learning process.}

\subsection*{Physical Interaction Effects Determine Task Difficulty}
\begin{figure}[!t]
    \centering
    \includegraphics[width=1.0\textwidth]{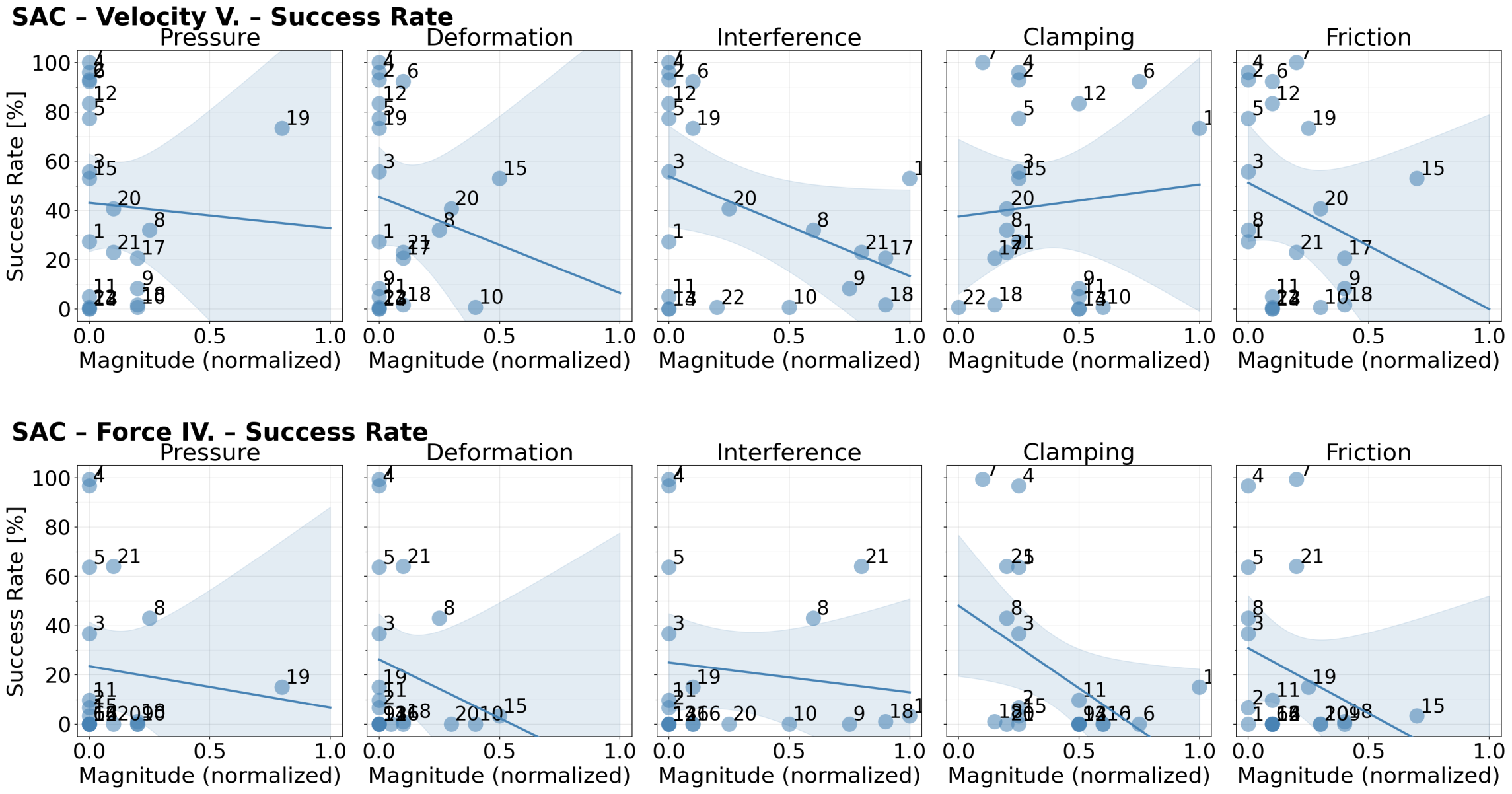}
\caption{%
  \textbf{Regression of insertion success rate against 
  normalised physical effect magnitude.}
  Individual task success rates are plotted against 
  normalised magnitude for each physical interaction 
  effect (P1~pressure, P2~deformation, P3~interference 
  fit, P4~clamping, P5~friction), with linear regression 
  fits. Task indices correspond to 
  Table~\ref{tab:success_best}.
  \textbf{(top)}~Level~\textsc{V} velocity domain (SAC; 
  Interaction Task Space--Range--Variation), the highest 
  model knowledge level evaluated in the velocity domain.
  Deformation~(P2), interference fit~(P3), and 
  friction~(P5) exhibit consistently negative trends; 
  pressure~(P1) is approximately flat; clamping~(P4) 
  shows a weak positive trend.
  \textbf{(bottom)}~Level~\textsc{IV} force domain (SAC; 
  Interaction Task Space--Range), the highest model 
  knowledge level safely evaluable in the force domain — 
  Level~\textsc{V} force-domain action scaling exceeded 
  the experimental safety threshold.
  In contrast to the velocity domain, clamping~(P4) 
  exhibits the strongest negative trend, while 
  deformation~(P2) and friction~(P5) remain strongly 
  negative; interference fit~(P3) shows a weak 
  high-variance trend; pressure~(P1) remains 
  approximately flat.%
}
    \label{fig:sac_regression_success}
\end{figure}

The five investigated physical interaction
effects, namely pressure (P1), deformation (P2), interference
fit (P3), clamping (P4), and friction (P5), produce systematic
relationships with deep\,RL success rate across the benchmark,
as shown by regression analysis against normalised effect
magnitudes at the highest evaluated model knowledge levels
(Fig.~\ref{fig:sac_regression_success}). \textcolor{cor2}{Further effects such as adhesion and magnetic forces exist but are not independently examined here, as they are of diminished magnitude for the tasks studied. Closely related physical phenomena --- including vibration, impact, thermal fluctuation, and lubrication --- are not effects in themselves but generate or modulate contact effects, and were either negligible or not independently observable within the experimental scope. Finally, wear occupies a different category altogether: a slow, cumulative process that gradually alters geometry and surface properties, thereby modifying effect magnitudes over time rather than constituting an effect in its own right; sound assembly design seeks to minimise it rather than accommodate it.} Results for PPO and for lower model
knowledge levels are consistent in direction and are provided in the
Supplementary Information Data. The pattern differs qualitatively between the velocity
and force domains, revealing that the control interface shapes not only
whether learning succeeds but which physical properties determine
failure.

In the velocity domain at Level\,\textsc{V}
(Fig.~\ref{fig:sac_regression_success}), deformation (P2), interference fit
(P3), and friction (P5) exhibit consistently negative trends against
success rate. Deformation~(P2) produces the steepest negative slope
for SAC across both evaluated levels, reflecting the difficulty of
navigating dynamically changing contact geometry without explicit
material compliance knowledge. For PPO at Level\,\textsc{V},
deformation shows a positive trend, suggesting that the
exploitation-promoting effect of adaptive variance reduction
specifically helps PPO converge to strategies that navigate elastic
deformation, a pattern not observed in SAC whose entropy-based
exploration continues to be penalised by deforming contact geometry.
Friction similarly produces strong task-level sensitivity, reflecting
the dependence of successful insertion on precise approach angle and
surface contact conditions that are not directly observable in the
state representation available to deep\,RL. Interference fit shows a
moderate negative trend, consistent with a force barrier that must be
overcome but which does not prevent success once the correct force
magnitude is applied. Pressure (P1) is flat at Level\,\textsc{V} ---
weakly negative at Level\,\textsc{IV} and uninformative at the highest
evaluated level, suggesting that adaptive variance reduction is
sufficient to overcome pressure-dominated tasks even when deformation
and friction remain limiting.

Clamping (P4) shows a weak positive trend for SAC at
Level\,\textsc{V} in the velocity domain, the only effect--level
combination for SAC across the benchmark in which the regression
slope reverses sign. This
anomaly does not reflect a general property of clamping but a confound
specific to the benchmark composition: tasks with high clamping
magnitudes are predominantly socket wrenches and spring-loaded
connectors whose passive contact guidance reduces the effective
dimensionality of the policy search once partial insertion is achieved,
since the spring contacts constrain the remaining motion and make the
final phase of insertion more robust to parameter variation. This
effect is not reproduced for PPO at the same level, suggesting it
depends on the specific interaction between SAC's entropy-regularised
exploration and the passive stabilisation provided by clamping
contacts. The anomaly illustrates the combinatorial nature of physical
effect interactions: single-effect magnitude is not a sufficient
predictor of task difficulty when geometric and physical properties
co-vary with the effect profile.

In the force domain at Level\,\textsc{IV}
(Fig.~\ref{fig:sac_regression_success}), the dominant predictor of failure
shifts from deformation to clamping. Clamping shows the steepest
negative slope across all force domain success panels, steeper than
any effect in the velocity domain. This reflects the heightened
sensitivity of direct wrench control to spring-loaded contact forces:
in the velocity domain, the impedance-based control structure provides
implicit compliance that partially absorbs clamping loads, whereas
direct wrench commands must actively counteract them without this
self-stabilising mechanism. Friction retains a strongly negative trend,
consistent with the velocity domain. Deformation is moderately
negative. Interference fit shows a shallow trend with high residual
variance, suggesting that the force barrier it presents is navigated
inconsistently in wrench control. Pressure is flat in the force domain
as well.

The residual variance in both panels is high throughout, reflecting
the combinatorial nature of co-occurring effects. Tasks combining high
magnitudes across multiple effects, such as the \textcolor{cor}{\emph{Schuko}} outlet
(P2, P3, P4, P5) or \textcolor{cor}{\emph{USB-C}} connector (P2, P3, P5), consistently
produce the lowest success rates across all deep\,RL configurations,
while tasks with predominantly low effect profiles achieve the highest.
Task difficulty cannot be predicted from any single physical effect in
isolation.

Insertion speed given success shows a consistent negative relationship
with physical effect magnitude across both domains and both model
knowledge levels, confirming that policies achieving success on
physically demanding tasks do so more slowly. The force domain shows
steeper speed penalties than the velocity domain, reflecting the
exploratory contact behaviour required by direct wrench control.
At Level\,\textsc{V} velocity, PPO shows a reversal in which higher
effect magnitudes are associated with faster execution, consistent with
exploitation convergence under adaptive variance reduction locking the
policy into efficient trajectories; this reversal is absent for SAC,
whose entropy-based exploration maintains broader trajectory
distributions even after convergence. Insertion speed panel figures for all
level and algorithm combinations are provided in the Supplementary Information.

\subsection*{Policy Landscape Structure and Parameter Importance}

\begin{figure}[!t]
    \centering
    \includegraphics[  trim=0cm 0cm 0cm 0cm,
  clip,width=1.0\textwidth]{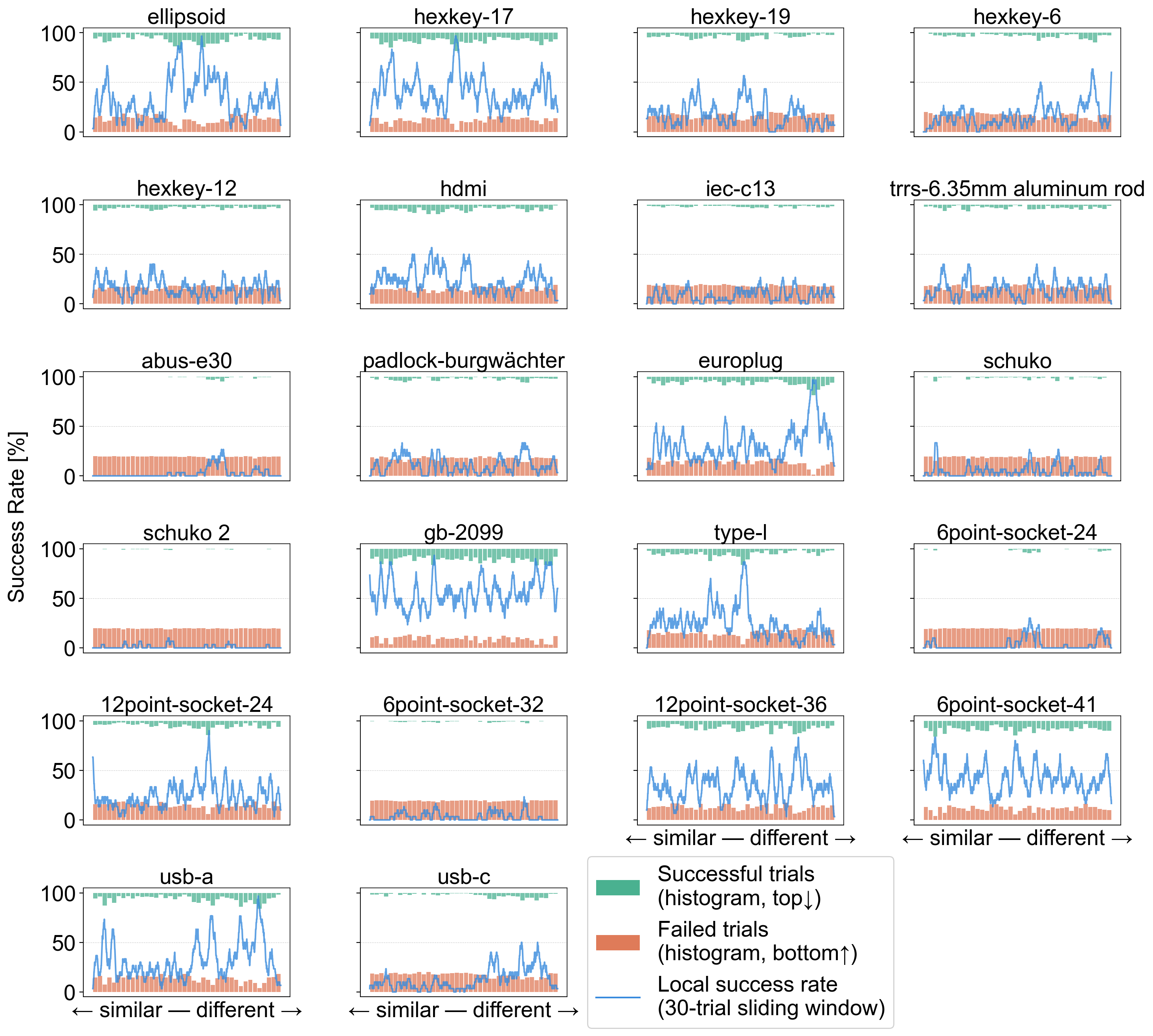}
\caption{%
  \textbf{Policy landscape structure 
  of the 33-dimensional Tactile Skill 
  parameter space $\boldsymbol{\theta}$, 
  visualised via Hilbert curve 
  indexing.}
  Each panel represents one of the 
  22 benchmark tasks. The $x$-axis 
  encodes the Hilbert curve index 
  of each trial's parameter vector 
  $\boldsymbol{\theta}$, preserving 
  local neighbourhood structure in 
  the high-dimensional space; trials 
  with similar parameter vectors 
  appear at similar $x$-positions.
  Green histograms (top) 
  show the density of successful 
  trials; orange histograms (bottom) show the density of failed 
  trials.
  The blue line indicates a 
  30-trial sliding-window local 
  success rate.
  Successful trials form isolated, 
  narrow clusters separated by 
  failure-dominated regions rather 
  than broad continuous success 
  zones, consistent with the 
  multimodal landscape structure 
  described in the Results. This is robust to projection choice---resolution, space-filling
curve, and dimension ordering (see Supplementary Information,
Fig.~\ref{fig:hilbert_sensitivity}).}
    \label{fig:hilbert}
\end{figure}

\begin{figure}[!t]
    \centering
    \includegraphics[width=1.0\textwidth]{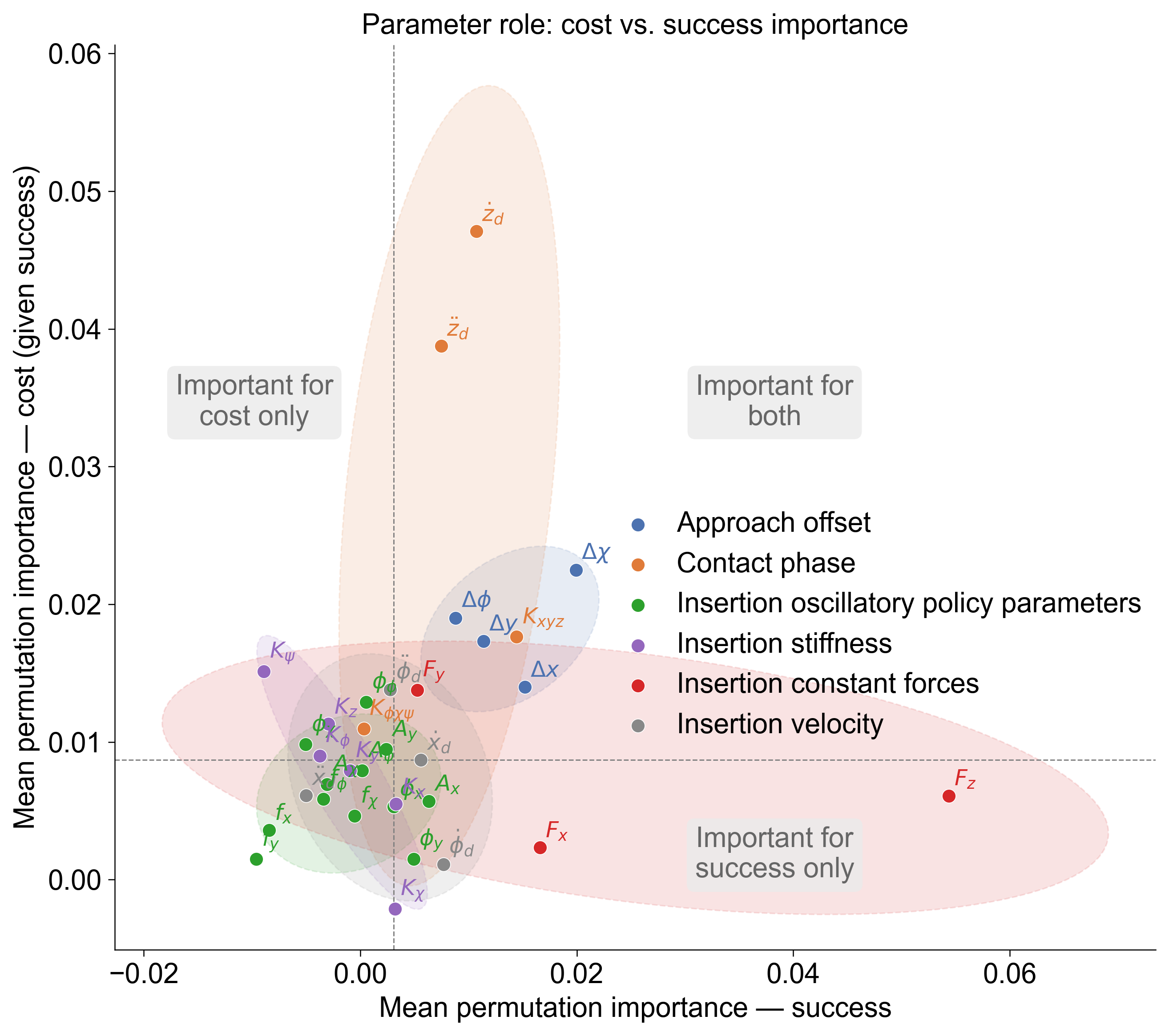}
    \caption{%
  \textbf{Global permutation-based feature importance for insertion
  success and cost across 18 tasks.}
  Each point represents one of the 33 Tactile Skill parameters,
  plotted by its mean permutation importance for success ($x$-axis)
  against cost given success ($y$-axis), averaged across tasks and
  independent runs.
  Feedforward force magnitude~$F_z$ dominates success prediction,
  reflecting its role in overcoming force barriers, friction, and
  clamping loads, while contact velocity~$\dot{z}_d$ dominates cost
  prediction.
  The partial decoupling of $F_z$ and $\dot{z}_d$ indicates that
  success and execution efficiency are governed by distinct aspects
  of the parameter vector.%
}
    \label{fig:scatter_cost_success}
\end{figure}

A full description of the Tactile Skill \textcolor{cor}{model} parameter structure and the
permutation importance methodology is provided in the Materials and Methods.
Analysis of the learned parameter landscapes reveals that success
regions in the 33-dimensional Tactile Skill parameter space are narrow
and multimodal across 22 tasks (Fig.~\ref{fig:hilbert}). Sorted by
their Hilbert curve index, which preserves local neighbourhood
structure in high-dimensional space, successful trials cluster in
isolated regions separated by large failure-dominated areas rather than
forming broad continuous zones. \textcolor{cor}{This may be in part due to the discrete phase transitions
between the individual phases of the skill model. } The percentages shown in
Fig.~\ref{fig:hilbert} indicate the fraction of trials that were
successful during the learning campaign, not the success rate of the
final selected policy, which achieves 100\% following
parameter selection. Cluster width and density vary systematically with
physical effect magnitude: tasks with lower effect profiles, such as
\textcolor{cor}{\emph{gb-2099}} (55\% trial success rate during learning), exhibit more
numerous and broader success clusters, while tasks with high effect
magnitudes, such as \textcolor{cor}{\emph{abus-e30}} (3\% trial success rate), show only
sparse and isolated success events.

Permutation-based feature importance analysis identifies 
which parameters govern insertion success and execution 
efficiency across the 18 tasks with sufficient successful 
trials for reliable estimation; the remaining four tasks 
are omitted due to insufficient success rates 
(Fig.~\ref{fig:scatter_cost_success}, Supplementary Information 
Fig.~\ref{fig:ranking_success}, \ref{fig:ranking_cost}). Globally, the feedforward force
$F_z$ is the single strongest predictor of insertion success,
reflecting its role in overcoming force barriers, friction, and
clamping loads. The lateral approach offset $\Delta\chi$ and constant
force $F_x$ are the next most important success parameters. For
execution speed given success, contact velocity $\dot{z}_d$ dominates,
followed by approach offsets. These two objectives are partially
decoupled: $F_z$ is important for success but not for execution speed,
while $\dot{z}_d$ is important for execution speed but not for success,
indicating that the correct contact velocity is the primary determinant
of cycle time once the insertion force structure is sufficient.

At the task level, the relative importance of parameters shifts with
the physical effect profile. For \textcolor{cor}{\emph{USB-C}} (high deformation, interference
fit, friction), lateral oscillatory amplitude $A_x$ and constant force $F_z$ dominate
success, reflecting the need to navigate elastic deformation and
overcome a force barrier through active perturbation. For the \textcolor{cor}{\emph{Schuko}}
outlet (high clamping, deformation, interference fit), the approach
offset $\Delta\chi$ and lateral forces $F_x$, $F_z$ are most
important, consistent with spring-loaded clamping contacts that require
a specific approach geometry before axial force can be effective. For
hexagonal socket wrenches (high friction, geometric alignment),
approach offsets $\Delta\chi$, $\Delta x$ and rotational stiffness
$K_\psi$ dominate, reflecting the need for precise rotational alignment
prior to contact. \textcolor{cor}{The parameter importance structure is thus physically interpretable
and task-specific rather than generic across the benchmark, although shaped
in part by the sequential, stop-motion execution of the skill model's phases.}

The parameters that drive failure are not simply the inverse of those
that drive success (Supplementary Information Fig.~\ref{fig:scatter_failure}). $F_z$ remains the
strongest failure predictor --- insufficient feedforward force is the
primary failure mode --- but the remaining failure-driving parameters
differ from the success rankings, indicating that failure arises from
distinct parameter combinations rather than uniformly low values of
success-critical parameters.

\textcolor{cor2}{The SVM classifier used for the importance analysis achieves
F1 scores of 0.77--1.00 for success vs.\ failure
classification and 0.32--0.93 for low-cost vs.\ high-cost
classification across all 22 tasks
(Supplementary Fig.~\ref{fig:svm_f1}), indicating that the
parameter space contains learnable structure that the SVM
approximately captures for success prediction, with
substantially more variability for cost prediction. The
permutation importance analysis itself is restricted to the
18 tasks with sufficient successful trials for reliable
per-parameter estimation.}

\subsection*{Energy Consumption Comparison}
\textcolor{cor2}{Beyond task performance, deep\,RL models and the structured
Tactile Skill model differ substantially in resource
consumption, a direct consequence of their differing degrees
of model knowledge.} Robot energy dominates: the arm and real-time
controller together draw 42\,kJ per 100 trials, compared to
38.5\,kJ for the local PC, 6.1\,kJ for the training PC, and just
0.5\,kJ for inference
(Supplementary Information Table~\ref{tab:energy}). Because
computational cost is small relative to robot operation, energy
consumption per trial is nearly identical across all model knowledge
levels: 87.1\,kJ per 100 trials for deep\,RL versus 80.5\,kJ per
100 trials for the Tactile Skill, the difference being solely the
training and inference cost of deep\,RL. The difference in total
consumption therefore arises principally from trial count:
each deep\,RL experiment was executed for 500 trials, yielding
a total consumption of approximately 435.5\,kJ, whereas the
Tactile Skill requires only 160 trials, consuming
approximately 128.8\,kJ --- a factor of $3.4\times$ lower,
achieved by a system that simultaneously reaches 100\% success
on all tasks \textcolor{cor}{at a mean execution speed of
1.02\,Ex\,s$^{-1}$}. \textcolor{cor2}{If the deep\,RL models were executed until
convergence, one would likely observe a direct inverse
correlation between the degree of model knowledge and both
learning time and the associated energy consumption.}

\subsection*{Translation to Industrial Assembly and Validation}

\begin{figure}[!b]
    \centering
    \includegraphics[width=1.0\textwidth]{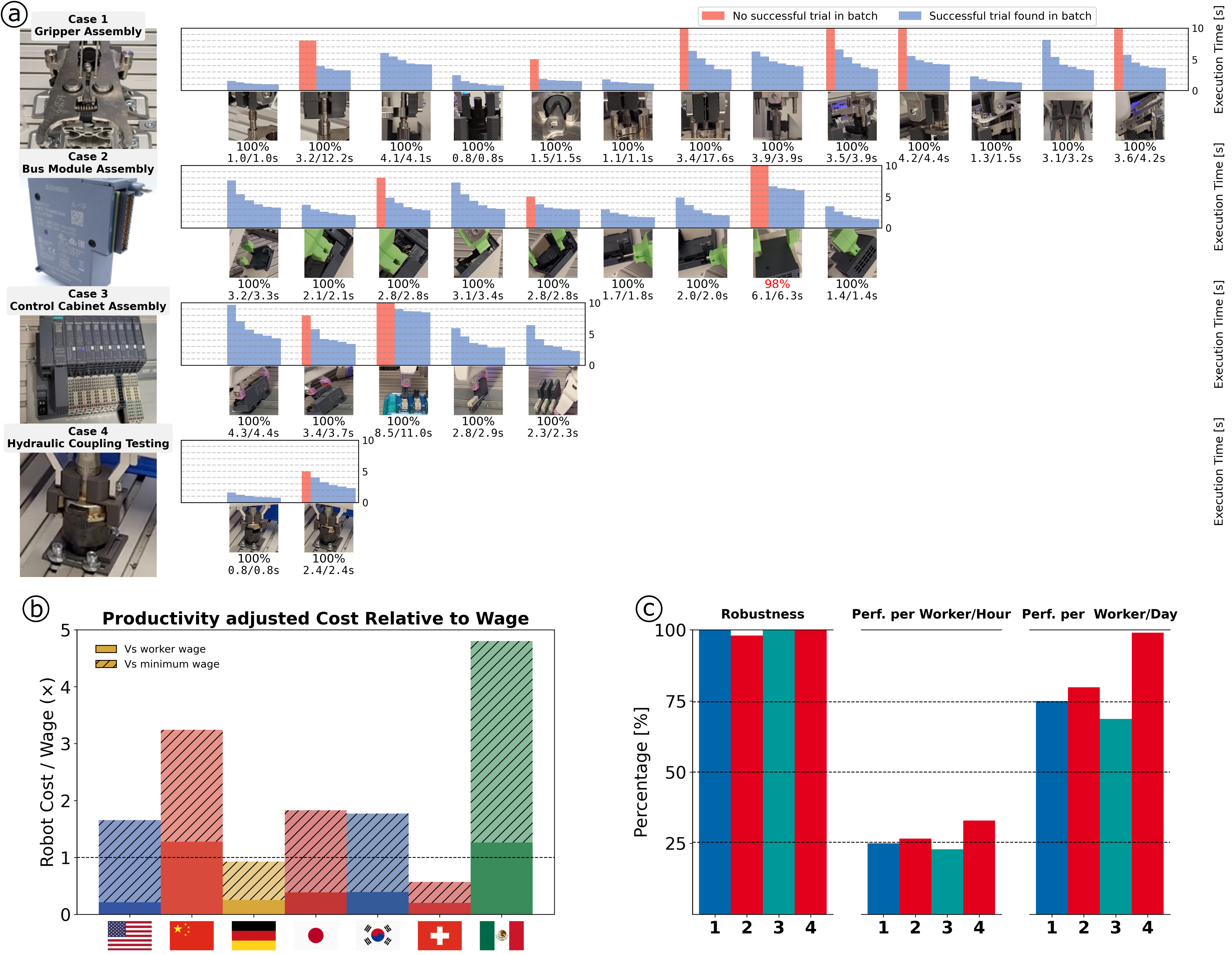}
\caption{%
\textbf{Industrial assembly validation across four manufacturing partners.}
\textbf{(a) Learning curves for contact-rich subtask types across four industrial partner assemblies}. Each bar group shows mean execution time per trial; red indicates batches (10 trials) with no successful trial, and blue indicates batches with at least one success. Annotations report the final success rate [\%] and the mean
    execution time together with the total execution time adjusted
    for retries and multiple executions [s] per subtask. All subtask types are learned within 30\,min. The \textcolor{cor}{model} achieves 100\% success over 50 consecutive full assembly 
runs across \textcolor{cor2}{three of the four assemblies, with the exception of case 2, 
where a single subtask failed once (98\%)}.
\textbf{(b) Productivity-adjusted robot cost per unit output relative to labour wages.} Bars show the ratio of robot operational cost to assembly worker wage (solid) and minimum wage (hatched); a ratio below 1.0 (dashed line) indicates lower cost than human labour. The system reaches cost parity with assembly worker wages across all evaluated high-wage economies, and with minimum wage in Germany and Switzerland.
\textbf{(c) Automation performance versus human baseline.}
Robustness over 50/100 consecutive full assembly runs (left), performance per worker-hour relative to a human baseline (centre), and performance per worker-day comparing continuous \textbf{single-arm} operation (24\,h) with an 8-hour human shift (right).
A \textbf{dual-arm} system achieves 150--200\% of human daily throughput.
}
    \label{fig:industrial_learning}
\end{figure}

\textcolor{cor}{The Tactile Skill model was validated across four
real-world industrial assembly workflows: Case~1 (gripper
assembly), Case~2 (bus module assembly), Case~3 (control cabinet
assembly), and Case~4 (hydraulic coupling testing)
(Fig.~\ref{fig:industrial_learning}). Partner companies are
anonymised; they span from small and medium enterprises to
global industry leaders in automation, and names are available
from the authors upon request. Critically, all four workflows
were selected from actual industrial practice, predominantly
executed by trained human personnel due to their contact-rich
and tactile nature --- not purpose-simplified for research.
This distinguishes the present validation from the majority of
prior work, which evaluates on isolated single tasks or
purpose-designed objects from which the physical interaction
effects and tolerances present in real-world manipulation have
been removed, likely representing the frontier of genuinely
unstructured learning. The analogy to human skill development
is instructive: basic force-motion capabilities, while a
meaningful milestone, do not imply the ability to perform at
the level of reliability, speed, and physical complexity
demanded by industrial production --- just as early motor
competence in a child does not translate to the trained
proficiency of an adult professional.} Each assembly was
executed fully autonomously across complete sequences of up to
13 unique contact-rich subtask types, with individual subtask
types appearing up to nine times within a workflow, yielding
up to 23 total subtask executions per assembly run, without
human intervention at any stage. The assemblies span broader
manipulation classes beyond insertion, including screwing,
\textcolor{cor}{visual} precision positioning, and
\textcolor{cor}{coordinated} force--motion.

Across all four assemblies and all subtask types, the \textcolor{cor}{model}
achieves 100\% robustness over 50 consecutive full assembly runs, with
a single exception: one subtask type in the bus module assembly
reaches 98\%, the only failure observed across the entire industrial
validation. Each subtask type is learned in under 30\,min, with
execution times per subtask ranging from approximately 0.8\,s for
simple insertions to 11.0\,s for multi-turn screwing operations
requiring successive regrasps. Within the first learning batch of
trials, several subtask types already yield successful executions,
reflecting the efficiency of the structured parameter initialisation.

Assuming continuous 24-hour operation, a single robot arm's execution speed of
25--33\% of the human rate across all four assemblies
(Table~\ref{tab:industrial_assemblies}) translates to a projected 150--200\% of
human daily throughput relative to a standard 8-hour shift for a dual-arm system
(two arms operating in parallel over three shift-equivalents),
\textcolor{cor}{under the assumption that} the robot incurs no fatigue, shift
changes, or idle time.
Human execution speed is established via \textcolor{cor}{the international gold standard} MTM (Methods-Time
Measurement) analysis~\cite{maynard1948mtm, niebel2003methods},
which decomposes each assembly into fundamental hand motions and
assigns empirically derived normal times.
MTM times are consistent with independently observed
human cycle times, validating the baseline.

The productivity-adjusted robot operational cost accounts for capital
amortisation and the speed differential relative to human workers. At a capital cost of \$60{,}000
amortised over 20{,}000 operating hours and operating at 25\% of human
single-arm execution speed, the productivity-adjusted cost per
human-equivalent output unit is \$12.00\,hr$^{-1}$. This falls below
assembly worker wages in all examined high-wage economies by a factor
of 0.20--0.39, and below minimum wage in Germany and Switzerland.
Manufacturing economies such as China and Mexico approach the
break-even point, indicating that further reductions in robot capital
cost or execution speed improvements would shift the economic balance.
Critically, this economic case is contingent on the 100\% reliability
achieved by the structured \textcolor{cor}{model}: the categorical failure of
end-to-end deep\,RL at sub-millimetre clearances would preclude
continuous autonomous operation and with it the throughput advantage
that drives cost parity.

These results confirm that the Tactile Skill \textcolor{cor}{model} is directly
deployable in industrial practice: reliable autonomous assembly is
achieved without simulation, without human demonstrations, and within a
setup time measured in hours rather than weeks. \textcolor{cor2}{This establishes that the importance of model knowledge is
not confined to controlled laboratory settings, whether basic
or complex, but translates directly into industrial reality.}

%% file: sections/discussion.tex
The central finding of this work is a representational boundary in embodied learning. Across 22 real contact-rich tasks, two deep\,RL algorithms, two control domains, and progressively increasing levels of model knowledge, learning performance improves as physically meaningful structure enters the representation, yet the highest evaluated model-informed deep\,RL levels remain unreliable on the majority of tasks. A structured model that represents embodiment, interaction control, and manipulation-process structure achieves 100\% success across all 22 tasks. The observed transition is therefore not merely a performance gap between algorithms: it identifies information that must be available to the learner for reliable physical interaction.

Within the sample budgets and task family investigated here, this boundary is not closed by additional data, model capacity, or the embodiment and interaction knowledge progressively supplied across Levels~\textsc{II}--\textsc{V}. Across 462,000 real-hardware trials spanning the two analysed algorithms, increasing samples or capacity within a structurally incomplete representation does not supply the missing manipulation-process model. This is not a claim that such structure can never be learned. Rather, reliable physical learning requires it to be represented --- either supplied explicitly or acquired by a learning mechanism capable of identifying it. Because the underlying issue is the availability of task-relevant physical structure to the learner, the result concerns representation design in embodied intelligence rather than connector insertion alone.

\subsection*{The Necessity of Structured Representations}

\textcolor{cor2}{Reliable physical capability is built on transmitted
structured knowledge, not on the scale of unstructured
experience. The first stone tools required a model of material
fracture behaviour without which random striking never
converges to a useful implement; precision metalworking
required the prior structural knowledge of smelting and
controlled heat treatment, \textcolor{cor}{each constituting a
model of physical interaction that enabled the next layer of
capability}; and modern CNC machining, capable of
sub-micrometre tolerances, rests on accumulated layers of
physical models --- kinematics, material science,
thermodynamics, control theory --- none recoverable from
observation of cutting outcomes alone.
\textcolor{cor}{The reliability of each system is a direct
consequence of the structured knowledge encoded at each prior
layer, not of the scale of unstructured experience.} Crucially,
this knowledge is not rediscovered in each generation but
transmitted: formal education conveys number before arithmetic,
grammar before composition, mechanics before engineering,
placing each learner's experience in the correct
representational neighbourhood from the outset. This is
precisely what the Tactile Skill \textcolor{cor}{model}
encodes: prior physical knowledge compressed into a structured
representation that allows the system to start from a
physically informed prior rather than from unstructured
experience.}

The case of large language models confirms rather than
contradicts this view, and is the most direct analogue to the
present debate. Language is not raw data --- it is itself a
structured model of the world, built over millennia of human
cognitive and cultural evolution: grammar encodes relational
structure between concepts, syntax encodes causal and temporal
dependencies. The corpus on which language models train is
therefore the accumulated output of human minds imposing
structure onto the world, and that structure is present in the
data before learning begins. This is exactly what physical
interaction streams lack. Joint states, end-effector poses,
and estimated wrenches do not encode deformation, interference
fit, friction variability, or clamping loads in a form that
unstructured learning can exploit --- the relevant structure
is absent from the input, and scale cannot supply what the
data does not contain.

These observations bear directly on several prominent learning approaches in the field. Imitation learning methods including Diffusion Policy and ACT achieve strong results on visually rich manipulation tasks by learning from human demonstrations — but this success is consistent with, rather than contrary to, the argument presented here. The human demonstrator implicitly encodes physical interaction structure through the demonstration trajectory: they navigate deformation, friction, and clamping correctly, and the learned policy replicates that structured behaviour\textcolor{cor}{, but neither understands nor predicts the underlying physics}. The physical knowledge enters through the demonstration rather than through the model architecture, but it must enter somewhere. Reported results on sub-millimetre-clearance insertion of real industrial components under 
significant physical interaction effects remain absent for these methods. Sim-to-real approaches reduce real-world data requirements but inherit the same structural limitation at the simulation boundary: the residual gap for deformation, friction variability, and material compliance is documented even in the highest-fidelity contact simulation frameworks, and the closest 
prior work at comparable scale explicitly fails on electronic connector insertion~\textcolor{cor}{\cite{tang2023industreal}}. 
Frontier models for robot manipulation represent the current state-of-the-art in scale-based approaches 
and have demonstrated non-trivial generalisation on visually rich tasks; their contact-task performance is evaluated in Barreiros et al.~\cite{barreiros2026lbm},  which explicitly notes limitations on tasks requiring precise force exchange — consistent with the structural argument that visual richness does not substitute for physical interaction structure in the observation stream. Industrial developments point in the same direction. Where frontier models report reliability gains, these stem from closing the loop over real-world experience rather than from pretraining alone~\cite{pi2025pistar06}, and in an open industrial challenge on connector insertion, nine of the ten best teams combined learned perception with classical force-controlled insertion strategies~\cite{intrinsic2026challenge}.
The common thread across all these approaches is that physical interaction structure must be present somewhere in the learning pipeline — in the model architecture, the demonstration, or the simulation — and cannot be recovered from standard observation streams by scale alone.

\textcolor{cor2}{Finally, in the mathematical and physical sense, a model is ideally a compact, clearly defined representation of
explicit knowledge --- flexible in how it is applied and
interpreted across contexts, yet strict in its formal
definition. Applying such explicit, compact models directly
is, by construction, the most efficient approach available,
assuming non-approximated results are desired.}

\subsection*{Mechanistic Account of Task Difficulty}

The physical effect taxonomy introduced in this work provides a mechanistic account of why particular tasks resist end-to-end 
learning. \textcolor{cor}{Fundamental physical effects such as} deformation, interference fit, and friction introduce state-dependent contact conditions \textcolor{cor}{and dynamics} that are not directly observable through joint states or end-effector wrenches. Clamping produces lateral retention forces whose direction and magnitude depend on geometric features below the resolution of standard pose estimation. These effects interact combinatorially: no single effect magnitude
predicts task difficulty in isolation, and tasks combining high
magnitudes across multiple effects do not come close to reliable
performance in any deep\,RL configuration, with success rates
often near zero.

The policy landscape analysis reinforces this interpretation at a structural level. Successful parameter configurations form 
narrow, isolated regions in the 33-dimensional Tactile Skill parameter space, \textcolor{cor}{often} separated by large failure-dominated volumes. This multimodal structure reflects the physical reality that successful insertion strategies are specific, not generic: they must be precisely adapted to the interaction regime of each task. We would like to stress that this observation is likely specific to the Tactile Skill model structure; other representations may exhibit different and possibly superior behaviour. Unconstrained exploration cannot reliably locate these regions within practical trial budgets, while structured initialisation from physical knowledge provides a prior that places candidate solutions in the correct neighbourhood from the outset.

\subsection*{Implications for Learning System Design}
These findings shift the design question from how large a learner must become to what physical structure its representation must make available. That structure need not be hand-coded: it may be supplied analytically, acquired through system identification or self-modelling, inherited from demonstrations, or introduced through simulation. Our result concerns the necessity of the representation, not the particular mechanism by which it is obtained.

A practical implication of these findings is that generalisation across tasks appears to be governed more strongly by physical 
interaction regimes than by geometric similarity. Tasks sharing similar dominant effects tend to exhibit similar learning behaviour, while geometrically similar tasks with different contact regimes may differ substantially in difficulty. This distinction is not captured in most existing manipulation benchmarks, which emphasise geometric variation while simplifying contact physics — a design choice that systematically underestimates the difficulty of real industrial assembly 
and overestimates the generality of learned policies.

The implications of these findings extend beyond insertion tasks to dexterous manipulation more broadly. In-hand reorientation, precision grasping, and fingertip force regulation involve the 
same classes of physical interaction effects identified in this taxonomy — deformation of soft contact surfaces, friction at the fingertip interface, interference fit in precision grasps, 
and clamping loads in pinch configurations. The observation streams available to dexterous learning systems face an analogous structural problem: joint angles, fingertip forces, and 
tactile array readings abstract away the physical interaction structure governing contact at the level of individual fingers and contact patches. Our results suggest that dexterous 
manipulation systems will encounter the same categorical performance boundary unless the physical interaction structure 
of fingertip contact — compliance, friction distributions, deformation geometry — is explicitly encoded in the learning representation rather than inferred from observation alone. 
Reliable dexterous capability likely requires not more data, but richer adaptable physical models embedded in the control and learning representation.

The economic validation across four industrial assembly workflows reinforces the practical consequences of this distinction. Achieving 150--200\% of human daily throughput and operating 
below assembly worker wages across all examined high-wage economies is a direct consequence of the 100\% reliability of 
the structured model under continuous 
operation. A system averaging 46.2\% success simply cannot be deployed in continuous autonomous operation regardless of its speed — reliability is the binding  constraint, and reliability requires structured physical knowledge. Among publicly reported results of frontier models for robot manipulation, only two state success rates above 99\,\% together with a denominator~\cite{sunday2026act2,dyna2025dyna1}; both concern a single repeated folding operation at centimetre-scale tolerance, whereas the workflows here comprise up to 23 contact-rich subtasks per run at clearances below 0.5\,mm.

\subsection*{Limitations}

Several limitations qualify these conclusions. The physical effect magnitudes used in regression analyses were assigned 
by expert consensus among five \textcolor{cor}{expert} engineers; 
direct measurement of these effects during insertion is not accessible with \textcolor{cor}{any state-of-the-art} standard 
sensing, and the assigned scores are therefore approximate. The regression results should be interpreted as capturing qualitative trends rather than precise quantitative mappings. The deep\,RL 
baselines are evaluated over three independent runs per configuration — sufficient given the scale of the experimental campaign, but below what would be needed to fully characterise rare success events. However, the primary interest lies not in the precise quantification of these rare events but in the broader finding that structured model knowledge is beneficial across learning frameworks and, in the contact-rich regime studied here, necessary for any form of meaningful learning. The Tactile Skill parameterisation reflects human-derived model knowledge about insertion tasks --- the structured understanding of approach geometry, contact compliance, and force requirements accumulated through engineering practice --- and was not ablated against alternative structured representations; other parameterisations may yield different and possibly superior learning efficiency or success characteristics, but this does not invalidate the key insight stated above.

All experiments assume known goal poses provided by kinesthetic teaching and rigid fixtures with repeatable placement; robustness to pose uncertainty, upstream manipulation, and fixture variation is not evaluated. All experiments are conducted on the Franka Emika robot, which is the main reference system for control, learning, and manipulation; transfer to systems with different actuation or sensing characteristics would require re-identification of model parameters. However, since the Tactile Skill model framework is grounded in Lagrangian dynamics as well as human-inspired control architectures proven to scale to actual industrial production reality, which describe fundamental motions and forces in generalised form, transfer to systems of comparable capability should in principle be feasible.

A further limitation is the sequential execution of the Tactile Skill phases: the current implementation completes each phase --- approach, contact, and insertion --- before initiating the next, rather than blending them continuously. This phase-discretised execution reduces execution speed and departs from the fluid, overlapping motion characteristic of human manipulation. Unifying these discrete phases into a continuous skill representation is a direction for future work.

Finally, the comparison with frontier models for robot manipulation rests on publicly reported, largely self-reported results that rarely state denominators, success criteria or tolerances~\cite{deepmind2026gr2,figure2026helix25,sunday2026act2,dyna2025dyna1}, and these models target open-ended tasks in unstructured environments rather than the known-goal-pose, fixed-fixture setting studied here.

These limitations do not alter any of the  central conclusions. The categorical performance boundary between even model-informed end-to-end deep\,RL and the structured Tactile Skill model is observed consistently across 22 tasks, two algorithms, two control domains, and four levels of deep\,RL model knowledge. It is robust to the specific choices of algorithm, parameterisation, and experimental conditions within the evaluated range. The implication for learning system design in embodied robotics is direct: for tasks governed by physical interaction effects that are not recoverable from standard observation streams, structured representations are not an optional refinement but a necessary condition. Models matter — not as a constraint on learning, but as the foundation that makes reliable learning possible at all.

%% file: sections/methods.tex
\begin{figure}[!t]
    \centering
    \includegraphics[width=1.0\textwidth]{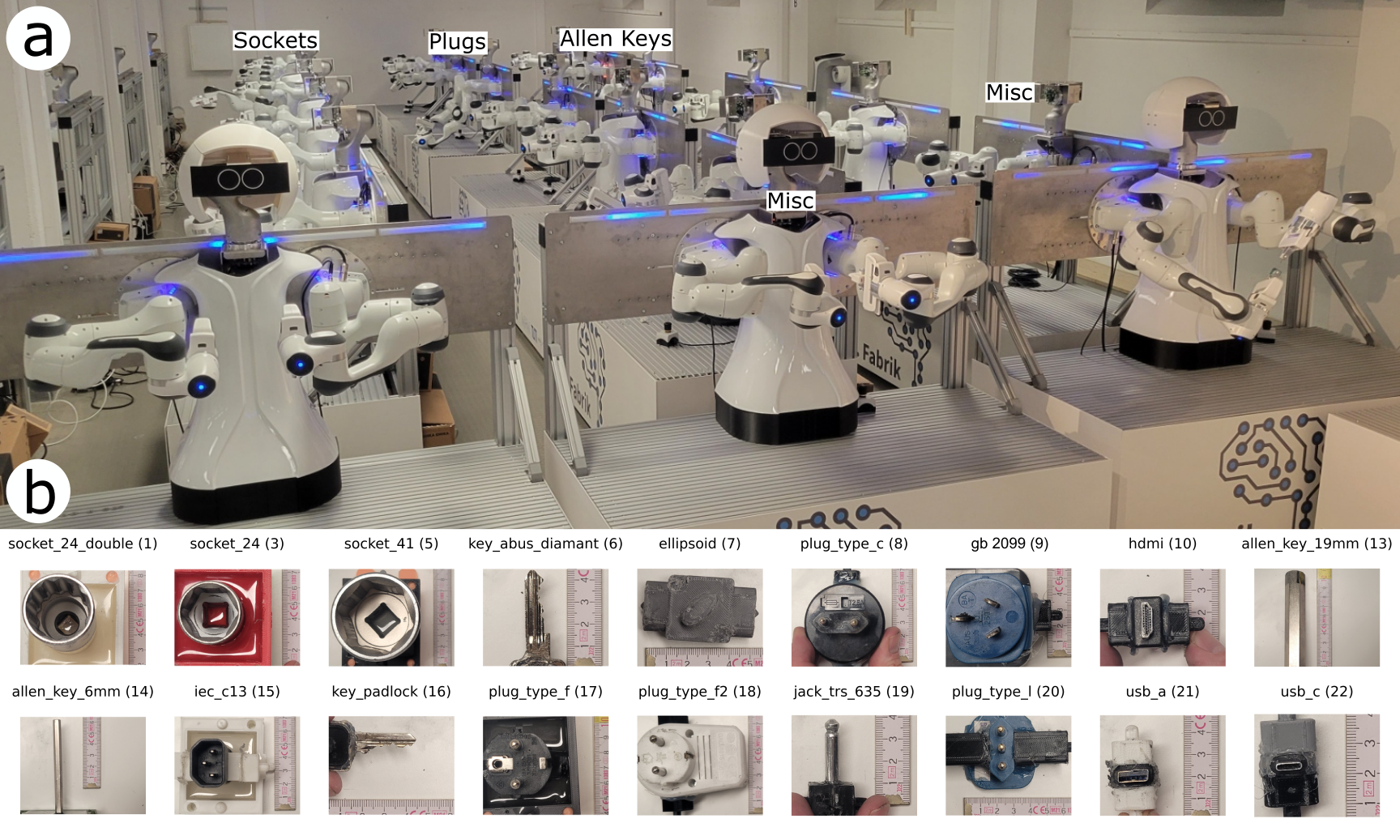}
    \caption{%
    \textbf{a) Experimental setup for the systematic model knowledge study.}
    The robot collective~\cite{collective2024} comprises 25 dual-arm robot setups, of which 22 Franka Emika Panda robotic arms were each dedicated to a single insertion task and operated in parallel without inter-robot communication.
    Each arm was part of a dedicated dual-arm setup mounted on a workbench with a rigid fixture holding the target object, while a parallel-jaw gripper held the insertion part.
    A central coordination PC (not shown) managed model training across all robots.
    Two additional robots were used for the clearance and geometry experiments, and five further robots were used for the industrial assembly validation, resulting in a total of 29 robotic arms across the full study.
    \textbf{b) Exemplary insertion objects from the systematic benchmark, with task indices as used throughout this work.}
    The 18 depicted objects span a broad range of real industrial connectors and mechanical components, with clearances ranging from sub-0.1\,mm to 1.0\,mm and physical interaction regimes covering all five effect categories (P1--P5).
    Task indices 2, 4, 11, and 12 are not shown, as they correspond to geometrically similar variants of depicted objects.}
    \label{fig:collective_objects}
\end{figure}

\subsection*{Hardware and Experimental Setup}

All experiments were conducted on Franka Emika Panda robotic arms~\cite{haddadin2022franka} --- the de facto standard platform in robot manipulation research --- providing 7-degree-of-freedom torque-controlled manipulation with integrated joint torque sensing and accurate estimation of external joint torques $\boldsymbol{\tau}_\mathrm{ext}$. \textcolor{cor2}{The robot is joint-torque controlled at 1\,kHz via its internal real-time interface, with nonlinear dynamics compensation, namely gravity, Coriolis, and friction compensation applied internally --- a further instance of model structure that is universally employed in practice, even by approaches claiming pure end-to-end learning~\cite{brohan2023rt2, black2025pi0, chi2023diffusion}.} Cartesian forces and moments at the end-effector are 
estimated from joint torque measurements using the robot's dynamic model, 
without the use of an external force-torque sensor. A parallel-jaw gripper 
is used to hold insertion objects \textcolor{cor}{firmly}, with only minimal play. All 
objects are fixed in the robot's workspace using a rigid fixture, and the 
robot end-effector pose is initialised to a fixed rendez-vous pose before 
each trial. The goal pose is provided as a fixed Cartesian target, taught once per
task via kinesthetic demonstration; in real-world deployments this pose
can equivalently be obtained from a calibrated vision system or
structured-light scan, requiring no modification to the learning
framework itself. No vision system is used at any stage of learning or execution,
\textcolor{cor}{as the focus is solely on the contact-rich
insertion phase governed by physical interaction effects and
force application.}

\textcolor{cor2}{The systematic model knowledge study employed 22 individual
Franka Emika robotic arms operated in parallel, each dedicated
to a single insertion task without inter-robot communication
(Fig.~\ref{fig:collective_objects}), and required
approximately 70 net days for the core benchmark experiments
alone, and roughly two years including preparation, setup,
testing, and complementary experiments and validation. The
depicted setup, consisting of 50 robot arms in total arranged
in dual-arm configurations, was shared with a concurrent
experimental study on robotic insertion learning; our study
predominantly used the right arm of each robot, the concurrent
study the left arm, with task assignments partially
overlapping between the two studies.} Each robot was connected to a dedicated PC equipped with an Intel Core i5-12600K processor, 16\,GB DDR4 RAM, and 500\,GB NVMe SSD. A central coordination PC with an Intel i9 processor, 64\,GB RAM, 2\,TB SSD, and an NVIDIA RTX 3090 GPU managed model training, while inference during manipulation was executed locally on each robot PC. Two robots were used for the clearance and geometry experiment setup, examining the limits of end-to-end deep reinforcement learning with respect to clearance (0.0 to 2.0\,mm) and shape complexity (triangular, square, and star cross-sections). Five robots were used for the industrial assembly validation, for a total of 29 robots across the work in this paper. All experiments were conducted exclusively on real hardware without any simulation.

\subsection*{Task Benchmark}

The systematic study benchmark consists of 22 tactilely challenging insertion tasks spanning a broad range of real industrial connectors and mechanical components, including hexagonal socket wrenches, \textcolor{cor}{\emph{USB-A}} and \textcolor{cor}{\emph{USB-C}} connectors, 
\textcolor{cor}{\emph{HDMI}} plugs, \textcolor{cor}{\emph{IEC\,C13}} power connectors, European \textcolor{cor}{\emph{Schuko}} power outlets, ellipsoidal connectors, and multi-point socket wrenches of varying size and 
point count (Fig.~\ref{fig:collective_objects}). Tasks were selected to span the full range of physical interaction regimes identified in the physical effect taxonomy, rather than to maximise geometric simplicity. Clearances range from 
sub-0.1\,mm to 1.0\,mm across the benchmark. Nearly all objects are real-world components sourced from consumer electronics, industrial connectors, and mechanical tooling, rather 
than purpose-manufactured or unrealistic 3D-printed test parts.

Each task is characterised by a physical effect profile across five interaction effect categories: pressure (P1), deformation (P2), interference fit (P3), clamping (P4), and friction (P5); 
see the Physical Effect Taxonomy section. Physical 
effect profiles were determined by consensus among five engineering professionals through expert analysis of each task's geometry, material, and contact conditions, and were used strictly to interpret but not to guide the learning 
experiments.

\subsection*{Learning Framework}

We evaluate a spectrum of learning approaches with increasing
degrees of model knowledge, from deep reinforcement learning
in robot joint space \textcolor{cor}{with model-based dynamics
compensation} --- the closest \textcolor{cor}{feasible}
instantiation to end-to-end learning evaluated here --- to a
fully structured Tactile Skill \textcolor{cor}{model}.
Importantly, even the most basic evaluated level is
\textcolor{cor}{clearly} not without model knowledge in the
strict sense: most of the manipulator's dynamics, most
critically gravity compensation \textcolor{cor}{and joint
friction}, are handled by the robot itself and are not learned.
Learning these from scratch \textcolor{cor}{as part of this
study} is infeasible on real \textcolor{cor}{systems} --- it
would cause \textcolor{cor}{severe} hardware damage during uncontrolled motion --- and sim-to-real approaches cannot substitute, as ideal simulation properties do not match real robot dynamics without additional system identification\textcolor{cor}{, which merely shifts the
modelling problem rather than eliminating it}. The 
evaluated deep\,RL levels therefore represent end-to-end learning in the task-relevant sense: no explicit encoding 
of contact physics, kinematics, or 
impedance structure.
\textcolor{cor}{All approaches optimise the same high-level task objective ---
reliable and rapid insertion --- but because the learning architectures differ,
this objective is operationalised through different reward signals. The scalar
reward for the Tactile Skill model is the cycle time in seconds for successful
insertions and a fixed penalty of 5\,s for failed trials, whereas the deep RL
levels receive a dense reward given by the reduction in distance to the goal
pose scaled by a gain (Eq.~\ref{eq:reward_drl}). This difference reflects
the architectures rather than a confound: the structured skill admits a sparse,
trial-level objective, while the generic neural policies require a shaped
per-step signal to learn at all.} This formulation jointly 
optimises success rate and execution speed, with successful policies emerging 
as those that minimise cycle time by first achieving reliable insertion and 
then reducing execution time.

\begin{figure}[!t]
    \centering
    \includegraphics[width=1.0\textwidth]{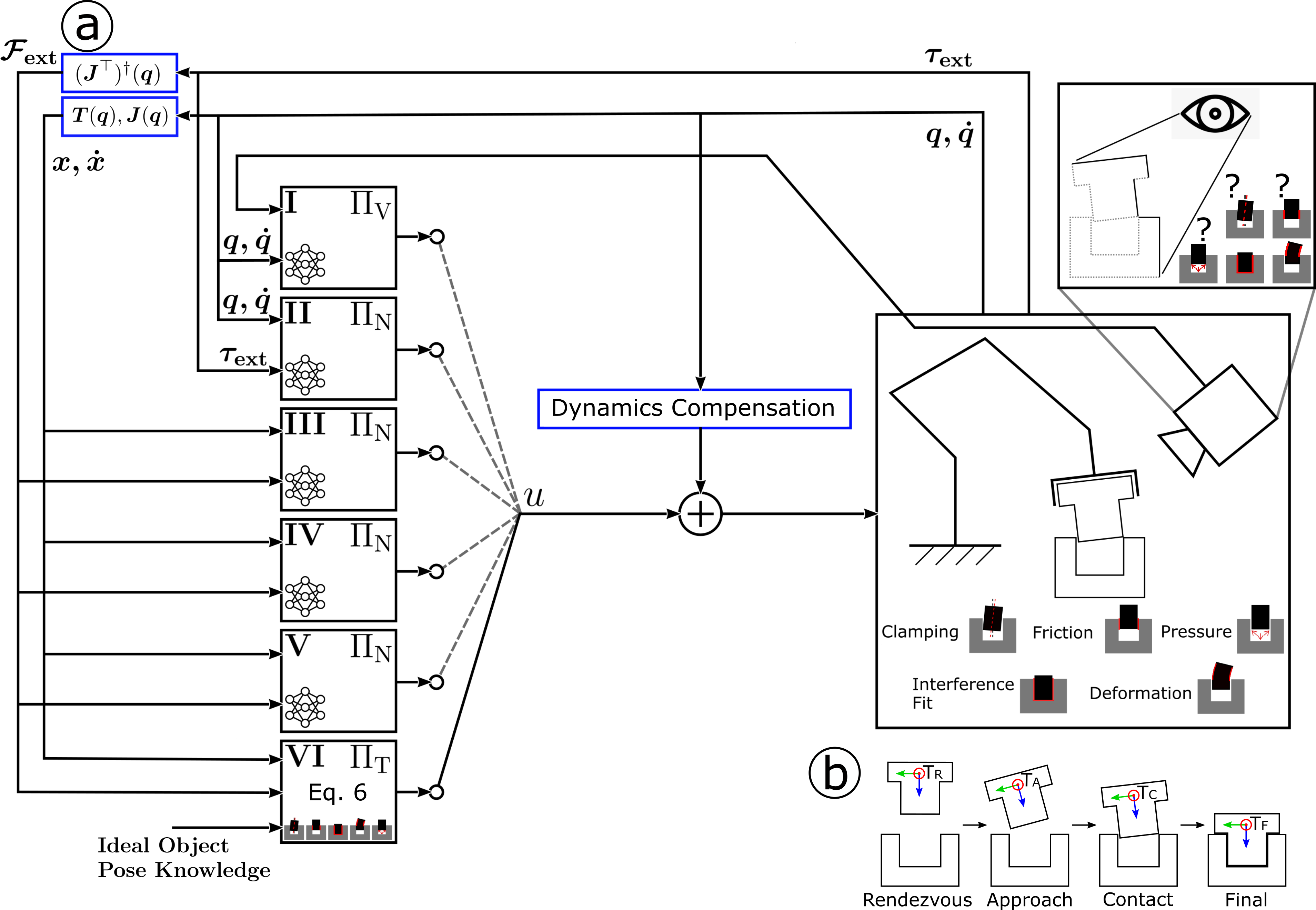}
    \caption{%
  \textbf{(a) Model knowledge hierarchy and learning 
  framework overview.}
  Six learning levels are shown with their state spaces,
  action spaces, and policy structures.
  $\boldsymbol{T}(\boldsymbol{q})$: forward kinematics; 
  $\boldsymbol{J}(\boldsymbol{q})$: robot 
  Jacobian ; $(\boldsymbol{J}^\top)^{\dagger}(\boldsymbol{q})$: pseudoinverse of robot 
  Jacobian transpose; dynamics compensation 
  comprises modelled gravity and Coriolis terms 
  (Eq.~\ref{eq:impedance}).
  $\Pi_\mathrm{V}$: vision-based neural network policy; 
  $\Pi_\mathrm{N}$: neural network policy; 
  $\Pi_\mathrm{T}$: Tactile Skill policy.
  Level~\textsc{I} (vision and joint state) is not 
  evaluated: visual observation cannot resolve the 
  physical interaction effects governing insertion — 
  deformation, friction, interference fit, clamping, 
  and pressure leave no reliable visual signature 
  (top-right inset).
  Note that even Level~\textsc{I} implicitly relies 
  on model knowledge: gravity compensation and robot 
  dynamics are handled by the manipulator and not 
  learned.
  Levels~\textsc{II}--\textsc{V} are deep\,RL 
  approaches with increasing model knowledge: 
  interaction joint space (\textsc{II}), interaction 
  task space (\textsc{III}), action-range scaling 
  (\textsc{IV}), and adaptive variance reduction 
  (\textsc{V}); Level~\textsc{VI} is the structured 
  Tactile Skill.
  Blue boxes highlight embodiment-specific knowledge 
  — kinematics, Jacobians, gravity compensation, 
  Coriolis terms — often treated as implicit 
  infrastructure in deep learning \textcolor{cor}{without acknowledgement, despite being universally employed by approaches claiming end-to-end learning}.
  \textbf{(b) Tactile Skill insertion 
  primitive: phase decomposition.}
  The skill is parameterised by 
  $\boldsymbol{\theta} \in \mathbb{R}^{33}$ and 
  executed across three phases.
  \textit{Approach}: lateral and rotational offset 
  $(\Delta x, \Delta y, \Delta\chi, \Delta\phi)$ 
  from rendez-vous to goal pose.
  \textit{Contact}: compliant impedance contact with 
  learned velocity~$\dot{z}_d$ and 
  stiffness~$\boldsymbol{K}_{xyz}$, $\boldsymbol{K}_{\phi\chi\psi}$.
  \textit{Insertion}: 6D impedance-controlled 
  insertion with learned stiffness, oscillatory 
  force policy (amplitude, frequency, phase), and 
  feedforward force~$\boldsymbol{F}_{ff}$.%
  \vspace{1cm}
}
    \label{fig:hypothesis}
\end{figure}

The four considered deep\,RL \textcolor{cor}{levels of model
knowledge} and their configurations are summarised in
Table~\ref{tab:levels_transposed}. Levels~\textsc{II}--\textsc{V}
are each evaluated in two control domains: the force domain,
in which the policy commands desired end-effector wrenches or
joint torques, and the velocity domain, in which the policy
commands desired Cartesian or joint velocities
(Fig.~\ref{fig:hypothesis}). \textcolor{cor}{Note that in both
domains the policy outputs feed into highly optimized
model-based controllers.}
Each level adds a specific form of model knowledge motivated
by a known limitation of the prior level (see also
Supplementary Information table~\ref{tab:model-knowledge-layers}). Starting from
Level~\textsc{II}, which operates in joint space,
Level~\textsc{III} adds task-space kinematics because
joint-space policies cannot exploit Cartesian contact
geometry; Level~\textsc{IV} adds action scaling because
unscaled Cartesian actions are insufficient to modulate the
forces required for physical insertion, both in terms of
magnitude and axis-specific sensitivity; 
Level~\textsc{V} adds adaptive variance 
reduction because fixed exploration 
prevents convergence once partial success 
is achieved; Level~\textsc{VI} replaces 
the neural network policy entirely with 
explicit physical structure because the 
remaining interaction effects are not 
recoverable from the observation stream.

The transition from Level~\textsc{V} to Level~\textsc{VI} is
necessarily architectural rather than a single-variable
algorithmic increment. Levels~\textsc{II}--\textsc{V}
progressively provide a generic neural policy with increasing embodiment
and interaction knowledge, yet retain a
function-approximation formulation in which the structure of
the manipulation process itself must be discovered from
experience. Level~\textsc{VI} introduces an explicit process
representation comprising discrete approach, contact, and
insertion phases, compliant force--motion regulation, and
physically meaningful policy parameters. The transition
therefore constitutes the final step in the model knowledge
hierarchy: once the invariant physical structure of the task
is explicitly represented, learning can focus exclusively on
the task-dependent quantities that remain unknown.

\begin{table}[h]
\centering
\caption{Overview of the four considered DRL learning levels (both PPO and SAC) and their key properties. State spaces, action spaces, and variance schedules are denoted for each level.
$\mathcal{P}_\mathcal{D}$ denotes the neural network policy; $f_s$ the 
action scaling vector; $s_\sigma(n_t)$ the trial-dependent variance schedule.}
\label{tab:levels_transposed}
\small
\setlength{\tabcolsep}{4pt}
\renewcommand{\arraystretch}{1.1}

\begin{tabularx}{\linewidth}{l l l l l}
\hline
 & \textbf{II} & \textbf{III} & \textbf{IV} & \textbf{V}\\
\hline

\textbf{Name} &
Interaction &
Interaction &
Int. Task Space&
Int. Task Space\\

\textbf{} &
Joint Space &
Task Space &
+ Scaling &
Scaling + Var.
 \\

\textbf{State} &
$[\boldsymbol{q},\,\dot{\boldsymbol{q}},\,\boldsymbol{\tau}_\text{ext}] \in \mathbb{R}^{3 \times 7}$&
$[\boldsymbol{x},\,\dot{\boldsymbol{x}},\,\boldsymbol{\mathcal{F}}_\text{ext}] \in \mathbb{R}^{3 \times 6}$  &
$[\boldsymbol{x},\,\dot{\boldsymbol{x}},\,\boldsymbol{\mathcal{F}}_\text{ext}] \in \mathbb{R}^{3 \times 6}$  &
$[\boldsymbol{x},\,\dot{\boldsymbol{x}},\,\boldsymbol{\mathcal{F}}_\text{ext}] \in \mathbb{R}^{3 \times 6}$  \\

\textbf{Action} &
$\sim \boldsymbol{\mathcal{P}}_\mathcal{D}\in\mathbb{R}^7$ &
$\sim\boldsymbol{\mathcal{P}}_\mathcal{D}\in \textcolor{cor}{se(3)}$ &
$\sim \boldsymbol{f}_s\boldsymbol{\mathcal{P}}_\mathcal{D}\in \textcolor{cor}{se(3)}$ &
$\sim \boldsymbol{f}_s\boldsymbol{\mathcal{P}}_\mathcal{D}\in \textcolor{cor}{se(3)}$\\

\textbf{Scaling} &
--- &
--- &
$\boldsymbol{f}_s\in\mathbb{R}^6$ &
$\boldsymbol{f}_s\in\mathbb{R}^6$\\

\textbf{Variance} &
$0.1 \boldsymbol{I}$ &
$0.1 \boldsymbol{I}$ &
$0.1 \boldsymbol{I}$ &
$s_\sigma(n_t) \boldsymbol{I}$\\

\hline
\end{tabularx}
\end{table}

\subsubsection*{Level II — Interaction Joint Space}

This baseline level represents the most basic 
evaluated configuration: no task-space 
transformation, no explicit contact model, 
and no impedance structure are encoded in 
the learning representation. Physical 
interaction structure must therefore be 
inferred entirely from joint-space 
observations. Note that, as with all 
evaluated levels, the manipulator's 
internal dynamics compensation — gravity, 
Coriolis, and friction terms — remains 
active and is not subject to learning. \textcolor{cor}{One could argue that this constitutes an unfair
advantage for end-to-end approaches; however, these components
are prerequisites for any meaningful and safe physical interaction
in the real world.} The state is $\boldsymbol{s} = [\boldsymbol{q}, \dot{\boldsymbol{q}}, 
\boldsymbol{\tau}_\text{ext}] \in \mathbb{R}^{3 \times 7}$, where 
$\boldsymbol{q} \in \mathbb{R}^7$ are joint positions, 
$\dot{\boldsymbol{q}} \in \mathbb{R}^7$ are joint velocities, and 
$\boldsymbol{\tau}_\text{ext} \in \mathbb{R}^7$ are estimated external joint 
torques. The desired action is either a desired joint torque or desired joint velocity command 
$\boldsymbol{a} \sim \boldsymbol{\mathcal{P}}_\mathcal{D}(\boldsymbol{s}, \boldsymbol{\sigma}) 
\in \mathbb{R}^7$. Joint positions and velocities are normalised by the joint 
limits of the Franka robot. This level uses no explicit information about 
the robot kinematics its Cartesian workspace, or end-effector interaction state. The sampling variance $\boldsymbol{\sigma} = 0.1 \boldsymbol{I}$ 
follows the default initialisation of the 
Stable-Baselines3 
implementation~\cite{raffin2021stable} 
and is fixed throughout training.

\subsubsection*{Level III — Interaction Task Space}

This level encodes knowledge of the robot's 
forward kinematics and Jacobian by moving \textcolor{cor}{the learning problem}
to a Cartesian state and action 
representation. The robot state is 
$\boldsymbol{s} = [\boldsymbol{x}, 
\dot{\boldsymbol{x}}, \boldsymbol{\mathcal{F}}_\text{ext}]$ 
 $\in \mathbb{R}^{3 \times 6}$, where 
$\boldsymbol{x} \in \mathbb{R}^6$ is the 
end-effector Cartesian pose, 
$\dot{\boldsymbol{x}} \in \mathbb{R}^6$ the 
Cartesian velocity, and 
$\boldsymbol{\mathcal{F}}_\text{ext} \in \mathbb{R}^6$ 
the estimated external Cartesian wrench. 
The action $\boldsymbol{a} \in \textcolor{cor}{se(3)}$ 
is either a desired Cartesian velocity 
$\dot{\boldsymbol{x}}_d$ in the velocity 
domain, or a desired feedforward wrench 
$\boldsymbol{\mathcal{F}}_{ff}$ in the force domain.

\subsubsection*{Level IV — Interaction Task Space with Action Scaling}

This level augments Level III with an expert-determined, task-specific 
action scaling vector $\boldsymbol{f}_s \in \mathbb{R}^6$. The scaled action is 
$\boldsymbol{a} \sim \boldsymbol{f}_s\,\boldsymbol{\mathcal{P}}_\mathcal{D}(\boldsymbol{s}, 
\boldsymbol{\sigma}) \in \textcolor{cor}{se(3)}$. The scaling emphasises Cartesian 
directions physically meaningful for the given insertion task and suppresses 
less relevant directions. Values for $\boldsymbol{f}_s$ were determined through expert 
model knowledge and refined empirically, chosen to be large enough to permit task 
completion while remaining conservative enough to prevent excessive contact 
forces. This balance is particularly important in the force domain, where 
excessive scaling can lead to kinetic energy accumulation and safety limit 
violations upon impact.

\subsubsection*{Level V — Interaction Task Space with Action Scaling and 
Sampling Variation}
This level further augments Level IV with an adaptive sampling variance 
schedule. The variance evolves as $\boldsymbol{\sigma} = s_\sigma(n_t) \boldsymbol{I}$, 
where $n_t$ is the number of successful trials observed. Concretely, 
$\boldsymbol{\sigma} = w\boldsymbol{\sigma}_\text{init} \cdot 
\max(1 - 0.01 \cdot n_\text{success},\, 0) \boldsymbol{I}$, implementing an implicit 
curriculum in which exploration is progressively reduced as the policy 
accumulates successful experience and capped at zero to prevent negative 
variance. This level encodes the highest degree of model knowledge 
achievable within the deep RL paradigm examined here.

\subsubsection*{Level~\textsc{VI} --- Tactile Skill}

The Tactile Skill \textcolor{cor}{model} departs from 
the deep\,RL paradigm entirely. 
Rather than learning a neural network 
policy, it learns a parameter vector 
$\boldsymbol{\theta} = (\boldsymbol{\theta}_\pi, 
\boldsymbol{\theta}_c) \in \mathbb{R}^{33}$ 
that parameterises a tactile skill — the 
combination of a tactile policy and a 
tactile controller — following the GGTWreP 
framework~\cite{Johannsmeier2025ProcessCentricTaxonomy}. \textcolor{cor}{The framework itself constitutes a higher
level of model knowledge, and its superior performance is
best understood as the continuation of the same trend of
increasing model knowledge rather than as a paradigm shift
separate from the preceding levels.}

The \textit{tactile policy} encodes and 
generates coordinated twist and wrench 
commands
\begin{equation}\label{eq:tactile_policy}
    \boldsymbol{\pi}_d = 
    [\dot{\boldsymbol{x}}_d^\top, 
    \boldsymbol{f}_d^\top]^\top\textcolor{cor}{,}
\end{equation}
to drive the tactile controller, based 
on the percept vector $\boldsymbol{\Omega}$ 
and policy parameters $\boldsymbol{\theta}_\pi$. 
In the insertion skill, $\boldsymbol{\theta}_\pi$ 
encodes desired Cartesian motion commands 
$(\dot{\boldsymbol{x}}_d, \ddot{\boldsymbol{x}}_d)$, 
oscillatory wrench commands $(\mathbf{a}, \xi)$, 
and desired feedforward force $\boldsymbol{F}_{ff}$ 
— defining what the robot should do at 
each phase of the insertion.

The \textit{tactile controller} is driven by the tactile
policy~(\ref{eq:tactile_policy}) and informed by the
percept vector $\boldsymbol{\Omega}$. It is parameterised 
by $\boldsymbol{\theta}_c$ and generates 
effort-level commands through an impedance 
control architecture that simultaneously 
regulates motion and force
\begin{equation}
    \boldsymbol{\mathcal{F}}_\text{ext} = 
    \boldsymbol{M}_x(\boldsymbol{q})
    \ddot{\boldsymbol{x}} + 
    \boldsymbol{D}_d(
    \boldsymbol{M}_x(\boldsymbol{q}), 
    \boldsymbol{\theta}_{c})
    \dot{\tilde{\boldsymbol{x}}} + 
    \boldsymbol{K}_d(\boldsymbol{\theta}_{c})
    \tilde{\boldsymbol{x}} + 
    \boldsymbol{u}_f(\boldsymbol{\Omega}_t,t)\textcolor{cor}{,}
\end{equation}
where $\tilde{\boldsymbol{x}} = 
\boldsymbol{x} - \boldsymbol{x}_d$ is 
the Cartesian pose error, 
$\boldsymbol{M}_x(\boldsymbol{q})$ the 
Cartesian mass matrix, 
$\boldsymbol{K}_d(\boldsymbol{\theta}_c)$ 
the desired positive definite stiffness 
matrix parameterised by 
$\boldsymbol{\theta}_c$, and 
$\boldsymbol{D}_d$ the desired damping 
matrix, derived from the stiffness 
parameters $\boldsymbol{\theta}_c$ and 
the inertia matrix 
$\boldsymbol{M}_x(\boldsymbol{q})$ 
via critical damping
\begin{equation}
    \boldsymbol{D}_d = 2\sqrt{
    \boldsymbol{M}_x(\boldsymbol{q})\,
    \boldsymbol{K}_d(\boldsymbol{\theta}_c)
    }.
\end{equation}
For the insertion skill, $\boldsymbol{u}_f$ 
combines the oscillatory wrench policy 
and the constant feedforward force
\begin{equation}\label{eq:policy}
    \boldsymbol{u}_f(\boldsymbol{\Omega}_t, t) 
    = \mathrm{diag}(\boldsymbol{A})\,
    \boldsymbol{s}(t)
    + \boldsymbol{F}_{ff},
\end{equation}
where $\boldsymbol{s}(t) \in \mathbb{R}^6$ with
$[\boldsymbol{s}(t)]_i = \sin(2\pi \boldsymbol{f}_i t + \boldsymbol{\phi}_i)$,
$i = 1, \ldots, 6$
and $\boldsymbol{A}$, $\boldsymbol{f}$, 
and $\boldsymbol{\phi}$ are the learned 
oscillation amplitude, frequency, and 
phase shift vectors respectively, and 
$\boldsymbol{F}_{ff}$ the constant 
feedforward force — both constituting 
policy parameters $\boldsymbol{\theta}_\pi$. 
\textcolor{cor}{Finally, the} resulting action is
\begin{equation}
    \boldsymbol{a} = 
    \boldsymbol{\mathcal{F}} = 
    \boldsymbol{g}(\boldsymbol{\theta}_\pi, \boldsymbol{\theta}_c, \boldsymbol{\Omega})\textcolor{cor}{.}
\end{equation}
The key property of this parameterisation 
is that $\boldsymbol{\theta}$ implicitly 
captures the physical interaction effects 
present in each task without requiring 
their direct observation
\begin{equation}\label{eq:effects}
    \boldsymbol{\theta} = 
    \boldsymbol{h}(e_{P1}, e_{P2}, e_{P3}, 
    e_{P4}, e_{P5})\textcolor{cor}{,}
\end{equation}
where $e_{P1}$--$e_{P5}$ denote the 
normalised magnitudes of pressure, 
deformation, interference fit, clamping, 
and friction respectively. The structured 
parameterisation is sufficient to address 
all evaluated physical effect combinations, 
a property not achievable by any of the 
evaluated deep\,RL configurations.

The skill is decomposed into three 
sequential phases $s_{\pi,1}$, 
$s_{\pi,2}$, $s_{\pi,3}$, each governed 
by a distinct subset of 
$\boldsymbol{\theta}$ 
(Table~\ref{tab:params}, 
Fig.~\ref{fig:hypothesis}).

\textbf{Approach phase} ($s_{\pi,1}$)\textbf{.}
The robot moves from its rendez-vous pose 
to an approach pose. The policy parameters 
$\boldsymbol{\theta}_\pi^{(1)} = 
(\Delta x, \Delta y, \Delta\chi, 
\Delta\phi)$ define a lateral offset 
$(\Delta x, \Delta y)$ limited to 
$\pm5$\,mm and an orientation offset 
$(\Delta\chi, \Delta\phi)$ limited to 
$\pm0.17$\,rad, relative to the goal 
pose. No controller parameters are 
active in this phase.

\textbf{Contact phase} ($s_{\pi,2}$)\textbf{.}
The robot moves into contact under 
impedance control. The policy parameters 
$\boldsymbol{\theta}_\pi^{(2)} = 
(\dot{z}_d, \ddot{z}_d)$ 
govern the contact approach via desired 
downward velocity $\dot{z}_d \in 
(0, 0.5]$\,m/s and acceleration 
$\ddot{z}_d \in (0, 1]$\,m/s$^2$. 
The controller parameters 
$\boldsymbol{\theta}_c^{(2)} = 
(K_{xyz}, K_{\phi\chi\psi})$ define 
compliant behaviour: translational 
stiffness $K_{xyz} \in (0, 2000]$\,N/m 
and rotational stiffness 
$K_{\phi\chi\psi} \in (0, 200]$\,Nm/rad 
allow the end-effector to accommodate 
geometric misalignment while 
establishing stable contact.

\textbf{Insertion phase} ($s_{\pi,3}$)\textbf{.}
The robot executes the insertion under 
full \textcolor{cor}{task level} Cartesian
impedance control. The controller 
parameters $\boldsymbol{\theta}_c^{(3)} 
= \boldsymbol{K}$ comprise the full Cartesian 
stiffness $K_x, K_y, K_z \in 
(0, 2000]$\,N/m and $K_\phi, K_\chi, 
K_\psi \in (0, 200]$\,Nm/rad, primarily 
addressing deformation and interference 
fit effects (P2, P3). The policy 
parameters $\boldsymbol{\theta}_\pi^{(3)}$ 
comprise three components: the nominal 
trajectory $(\dot{\boldsymbol{x}}_d, 
\ddot{\boldsymbol{x}}_d)$ with 
translational velocity $\dot{x}_d \in 
(0, 0.5]$\,m/s, acceleration 
$\ddot{x}_d \in (0, 1]$\,m/s$^2$, 
rotational velocity $\dot{\phi}_d \in 
(0, 0.25]$\,rad/s, and acceleration 
$\ddot{\phi}_d \in (0, 0.5]$\,rad/s$^2$; 
the oscillatory wrench policy 
$(\mathbf{a}, \xi)$ with amplitude 
$A_x, A_y \in (0, 10]$\,N, $A_\phi, 
A_\chi \in (0, 5]$\,Nm, frequency 
$f_x, f_y \in (0, 2.5]$\,Hz, $f_\phi, 
f_\chi \in (0, 1]$\,Hz, and phase shift 
$\phi_x, \phi_y, \phi_\phi, \phi_\chi 
\in [0, 2\pi)$\,rad, primarily addressing 
clamping and friction (P4, P5); and the 
feedforward force $\boldsymbol{F}_{ff} = 
(F_x, F_y, F_z)^\top$, extending the 
GGTWreP $\mathbf{f}_\text{push}$ to all 
three translational axes. The combination of tactile 
controller $\boldsymbol{\theta}_c$ and 
tactile policy $\boldsymbol{\theta}_\pi$ 
enables the robot to implicitly address 
all five physical interaction effect 
classes, consistent with 
Eq.~\ref{eq:effects}.

\begin{table}[H]
\centering
\caption{\textcolor{cor}{Learning} parameters of the 
Tactile Skill 
$\boldsymbol{\theta} = 
(\boldsymbol{\theta}_\pi, 
\boldsymbol{\theta}_c) \in 
\mathbb{R}^{33}$, organised by phase 
following GGTWreP 
nomenclature~\cite{Johannsmeier2025ProcessCentricTaxonomy}. 
$\boldsymbol{\theta}_\pi$: tactile policy 
parameters (desired motion and wrench 
commands); $\boldsymbol{\theta}_c$: 
tactile controller parameters.}
\label{tab:params}
\small
\begin{tabular}{lllcl}
\hline
\textbf{Phase} & 
\textbf{Type} &
\textbf{Parameter} & 
\textbf{Dim.} & 
\textbf{Range} \\
\hline
\multicolumn{5}{l}{%
  \cellcolor{gray!10}%
  \textit{Tactile Policy Parameters 
  $\boldsymbol{\theta}_\pi$}} \\
\hline
$s_{\pi,1}$ Approach  
& $\boldsymbol{\theta}_\pi$ 
& Lateral offset $\Delta x$               
& $1$ & $\pm5$\,mm \\
& & Lateral offset $\Delta y$               
& $1$ & $\pm5$\,mm \\
& & Rotational offset $\Delta\phi$          
& $1$ & $\pm0.17$\,rad \\
& & Rotational offset $\Delta\chi$          
& $1$ & $\pm0.17$\,rad \\
\hline
$s_{\pi,2}$ Contact   
& $\boldsymbol{\theta}_\pi$
& Downward velocity $\dot{z}_d$           
& $1$ & $(0,\,0.5]$\,m/s \\
& & Downward acceleration $\ddot{z}_d$      
& $1$ & $(0,\,1]$\,m/s$^2$ \\
\hline
$s_{\pi,3}$ Insertion 
& $\boldsymbol{\theta}_\pi$ 
& Translational velocity $\dot{x}_d$      
& $1$ & $(0,\,0.5]$\,m/s \\
& & Translational acceleration $\ddot{x}_d$ 
& $1$ & $(0,\,1]$\,m/s$^2$ \\
& & Rotational velocity $\dot{\phi}_d$      
& $1$ & $(0,\,0.25]$\,rad/s \\
& & Rotational acceleration $\ddot{\phi}_d$ 
& $1$ & $(0,\,0.5]$\,rad/s$^2$ \\
& & Oscillation amplitude $\boldsymbol{A}$ 
  ($\mathbf{a}$)        
& $4$ & 
$A_x,A_y\in(0,\,10]$\,N \\
& & & &
$A_\phi,A_\chi\in(0,\,5]$\,Nm \\
& & Oscillation frequency $\boldsymbol{f}$ 
  ($\xi$)        
& $4$ & 
$f_x,f_y\in(0,\,2.5]$\,Hz \\
& & & &
$f_\phi,f_\chi\in(0,\,1]$\,Hz \\
& & Oscillation phase $\boldsymbol{\phi}$   
& $4$ & 
$\phi_x,\phi_y,\phi_\phi,\phi_\chi
\in[0,\,2\pi)$\,rad \\
& & Feedforward force $\boldsymbol{F}_{ff}$ 
& $3$ & 
$F_x,F_y\in(0,\,20]$\,N \\
& & & &
$F_z\in(0,\,20]$\,N$^\dagger$ \\
\hdashline
\multicolumn{5}{l}{%
  \cellcolor{gray!10}%
  \textit{Tactile Controller Parameters 
  $\boldsymbol{\theta}_c$}} \\
\hline
$s_{\pi,2}$ Contact   
& $\boldsymbol{\theta}_c$ 
& Translational stiffness $K_{xyz}$       
& $1$ & $(0,\,2000]$\,N/m \\
& & Rotational stiffness $K_{\phi\chi\psi}$ 
& $1$ & $(0,\,200]$\,Nm/rad \\
\hline
$s_{\pi,3}$ Insertion 
& $\boldsymbol{\theta}_c$ 
& Full 6D stiffness $\boldsymbol{K}_x$      
& $6$ & 
$K_x,K_y,K_z\in(0,\,2000]$\,N/m \\
& & & & 
$K_\phi,K_\chi,K_\psi\in(0,\,200]$\,Nm/rad \\
\hline
\multicolumn{5}{l}{$^\dagger$ Expanded 
to $(0,\,80]$\,N for Schuko outlet 
tasks.}
\end{tabular}
\end{table}

\subsection*{Control Architecture 
of Levels~\textsc{II}--\textsc{V}}

The two controller \textcolor{cor}{types} implemented and evaluated in this work are a
velocity-domain impedance controller and a force-domain feedforward force controller,
corresponding to rows \emph{PD $\wedge$ IK $\wedge$ $\hat{\boldsymbol{c}} ~\wedge ~\hat{\boldsymbol{g}}$ $\wedge$ $\textcolor{cor}{\hat{\boldsymbol{M}}}(\boldsymbol{q})$}
and \emph{$\textcolor{cor}{\boldsymbol{J}} \wedge \hat{\boldsymbol{c}} ~\wedge ~\hat{\boldsymbol{g}}$ $\wedge$ \textcolor{cor}{$\hat{\boldsymbol{M}}(\boldsymbol{q})$}}
in Table~\ref{tab:model_tree}, respectively.
Neither controller is end-to-end learned: both rely on substantial \textcolor{cor}{physical} model
knowledge that is \textcolor{cor}{deduced from model-based system identification}.

\paragraph{Task-space velocity domain.}
The velocity-domain controller is a Cartesian impedance controller
with \textcolor{cor}{model-based} dynamic compensation:
\begin{equation}
{\color{cor2}\boldsymbol{\tau}_d =
\boldsymbol{J}^\top(\boldsymbol{q})\left[
  \boldsymbol{K}_x \left(\boldsymbol{x}_0 + \int \dot{\boldsymbol{x}}_d(t)\,\mathrm{d}t -
  \boldsymbol{x}\right)
  - \boldsymbol{D}_x \dot{\boldsymbol{x}}\right]
  + \hat{\boldsymbol{c}}(\boldsymbol{q},\dot{\boldsymbol{q}})
  + \hat{\boldsymbol{g}}(\boldsymbol{q})},
\label{eq:impedance}
\end{equation}
\textcolor{cor}{where $\boldsymbol{J}(\boldsymbol{q})$ is the robot Jacobian, whose
transpose maps the commanded Cartesian wrench into joint torques, and
$\boldsymbol{K}_x$ and $\boldsymbol{D}_x$ denote the Cartesian stiffness and damping
matrices, with $\boldsymbol{D}_x = 2(\hat{\boldsymbol{M}}_x\boldsymbol{K}_x)^{1/2}$
chosen for critical damping.
$\hat{\boldsymbol{M}}_x(\boldsymbol{q})$ is the modelled operational-space inertia
matrix, $\hat{\boldsymbol{c}}(\boldsymbol{q},\dot{\boldsymbol{q}})$ the modelled
Coriolis and centrifugal torques, and $\hat{\boldsymbol{g}}(\boldsymbol{q})$ the modelled
gravity torques.
The learned policy $\pi_{\dot{\boldsymbol{x}}}$
contributes only the desired end-effector velocity
$\dot{\boldsymbol{x}}_d$; all remaining terms are computed from
the robot model. This controller therefore requires the robot's forward kinematics and
Jacobian $\boldsymbol{J}(\boldsymbol{q})$, the operational-space inertia
$\hat{\boldsymbol{M}}_x(\boldsymbol{q}) =
(\boldsymbol{J}\hat{\boldsymbol{M}}^{-1}(\boldsymbol{q})\boldsymbol{J}^\top)^{-1}$
--- itself requiring both the joint-space inertia $\hat{\boldsymbol{M}}(\boldsymbol{q})$
and $\boldsymbol{J}(\boldsymbol{q})$ --- and the Coriolis and gravity models
$\hat{\boldsymbol{c}}(\boldsymbol{q},\dot{\boldsymbol{q}})$,
$\hat{\boldsymbol{g}}(\boldsymbol{q})$.
Additionally, an external wrench estimate
$\hat{\boldsymbol{F}}_{\mathrm{ext}}$, obtained from a
momentum-based observer that itself requires $\hat{\boldsymbol{M}}(\boldsymbol{q})$ and
$\hat{\boldsymbol{g}}(\boldsymbol{q})$, is needed during learning to provide contact
feedback to the policy, though it is not part of the controller itself.}

\paragraph{Task-space force domain.}
The force-domain controller maps a learned task-space wrench
\textcolor{cor}{$\boldsymbol{\pi}_{\mathcal{F}}$} directly to
a desired task-space force, augmented with dynamic compensation:
\begin{equation}
\textcolor{cor2}{\boldsymbol{\tau}_d =
\boldsymbol{J}^\top(\boldsymbol{q})
  \boldsymbol{\pi}_{\mathcal{F}}
  + \hat{\boldsymbol{c}}(\boldsymbol{q},\dot{\boldsymbol{q}})
  + \hat{\boldsymbol{g}}(\boldsymbol{q})},
\label{eq:wrench}
\end{equation}
\textcolor{cor}{which is likewise mapped to joint torques.}
The policy provides only the desired wrench; \textcolor{cor}{all
remaining model terms} are supplied by the robot model. Without this compensation the robot cannot support its own
weight, as shown by the uncompensated rows in
Table~\ref{tab:model_tree}. \textcolor{cor}{Ultimately, both
domains require comparable levels of model knowledge; the
primary difference lies in which components are encoded in
the controller versus delegated to the policy.}

The evaluated joint-space controllers follow an analogous structure (see Table~\ref{tab:model_tree}).
Across all configurations, this hierarchy illustrates how physical knowledge is 
progressively shifted from the \textcolor{cor}{model-based} controller to the learned policy as model 
knowledge decreases. Crucially, none of the evaluated configurations are end-to-end: 
every policy operates within a structured control layer that explicitly encodes 
kinematic and dynamic robot properties. Gravity, Coriolis effects, and contact 
compliance are never learned from data --- they are \textcolor{cor}{absolute} prerequisites without 
which the robot cannot safely operate. \textcolor{cor}{It is worth noting that this applies universally:
to our knowledge, no existing work operates under a different
regime. This is particularly true for approaches using the
Franka Emika robot, the de facto reference platform in robot
manipulation research, where model-based dynamics compensation
is always present.}

\subsection*{Deep Reinforcement Learning
Procedure and Network Architecture}
For the deep reinforcement learning (DRL) conditions, each
experiment consists of 500 training trials per task. Three
independent experiments were conducted for every combination
of task, control domain, algorithm, and model knowledge level.
Three algorithms were evaluated in total --- PPO, SAC, and
DDPG --- amounting to 693,000 executed trials across the full
campaign. DDPG did not yield statistically meaningful success
rates across the benchmark and is therefore excluded from the
main analysis; the 462,000 trials conducted under PPO and SAC
form the basis of all reported results. Each experiment was
initialised with a different random seed. Both algorithms
perform 1,000 initial exploration steps without policy updates
before learning begins.

It is important to note that even the lowest evaluated model
knowledge levels are not end-to-end in the strict sense.
Beyond the robot dynamics compensation discussed above, the
experimental design itself encodes substantial prior
knowledge: the reward function provides a dense progress
signal rather than a sparse success indicator, termination
criteria constrain exploration to physically meaningful
regions of the workspace, and all hyperparameters reflect
community knowledge accumulated across years of deep\,RL
research --- none of which would be feasible without this
implicit model knowledge.

\paragraph{Network Architecture and Optimisation}
Proximal Policy Optimisation (PPO)~\cite{schulman2017proximal}
and Soft Actor-Critic (SAC)~\cite{haarnoja2018soft} --- the
two most widely adopted algorithms in robot manipulation
research, selected to enable direct comparison with prior
work; DDPG~\cite{lillicrap2015continuous} was additionally
evaluated but excluded from the headline analysis due to statistically non-meaningful success rates --- use multilayer perceptron policies with identical capacity: three hidden layers with 256 units each. Actor 
and critic networks are trained 
using the Adam 
optimiser~\cite{kingma2015adam} 
with a learning rate of 
$3\times10^{-4}$ and batch size 
of 256. No output activation 
function is applied in either case. 
All network architecture and 
optimisation hyperparameters are 
summarised in Supplementary Information 
Table~\ref{tab:net_arch}.

PPO uses a \texttt{tanh} activation 
function and collects rollout 
batches of 512 transitions per 
update. SAC uses \texttt{ReLU} 
activation and an off-policy replay 
buffer of $10^6$ transitions. Both 
algorithms use a discount factor 
$\gamma = 0.99$. SAC employs a 
trainable entropy coefficient 
initialised at $\alpha_\text{init} 
= 0.2$, while PPO uses a fixed 
entropy coefficient of $10^{-3}$ 
and generalised advantage 
estimation (GAE) with 
$\lambda = 0.95$~\cite{schulman2015gae}.

\paragraph{Reward Function and 
Termination Criteria}

The reward function and termination 
criteria encode three forms of 
prior knowledge about the insertion 
task that are not learned but 
explicitly engineered.

First, the reward at time step $t$ 
is a dense shaped signal defined 
as the reduction in Euclidean 
distance to the goal
\begin{equation}
r_t = 100\left(\left\| 
\boldsymbol{p}_{t-1} - 
\boldsymbol{p}_g \right\|_2 - 
\left\| \boldsymbol{p}_t - 
\boldsymbol{p}_g \right\|_2\right)\textcolor{cor}{,}
\label{eq:reward_drl}
\end{equation}
where $\boldsymbol{p}_t$, 
$\boldsymbol{p}_{t-1}$, and 
$\boldsymbol{p}_g \in \mathbb{R}^3$ 
denote the current, previous, and 
goal end-effector positions 
respectively. This formulation 
encodes prior knowledge that 
Cartesian progress toward the goal 
is the task-relevant signal; a 
truly assumption-free system would 
receive only a binary success 
indicator. The scaling factor of 
100 and the success bonus of $+25$ 
awarded upon reaching the goal 
within 2\,mm are further 
engineering choices that shape 
the reward landscape.

Second, the 2\,mm success tolerance 
encodes physical knowledge of what 
constitutes a completed insertion 
— this threshold is not learned 
but set from \textcolor{cor}{expert} knowledge 
of the task.

Third, a trial is terminated if the 
maximum execution time of 5\,s is 
exceeded or if the end-effector 
leaves a $\pm5$\,cm cube around 
either the initial or goal pose. 
Both criteria \textcolor{cor}{correspond to}
physical task knowledge: the spatial 
termination cube constrains 
exploration to the physically 
meaningful workspace region around 
the insertion axis, preventing the 
policy from discovering degenerate 
solutions in unreachable or 
task-irrelevant configurations; the 
time limit reflects prior knowledge 
of typical human insertion duration 
and ensures that trials consuming 
the full time budget without 
progress are terminated rather than 
allowed to accumulate uninformative 
experience. Both would be absent 
in a purely end-to-end system 
operating without task-specific 
constraints.

\paragraph{Evaluation Protocol}
The final 100 trials of each 
500-trial learning run constitute 
the evaluation set; 
Success rate $\bar{s}$ is computed 
as the fraction of successful trials 
in these final 100 executions, 
averaged over three independent 
experiments
\begin{equation}
    \bar{s} = \frac{1}{3}\sum_{k=1}^{3} 
    \frac{1}{100}\sum_{t=401}^{500} 
    \mathbbm{1}[\text{success}_{t,k}]\textcolor{cor}{.}
\end{equation}
Mean cycle time $\bar{c}$ is reported 
over successful trials only
\begin{equation}
    \bar{c} = \frac{\sum_{t,k} 
    c_{t,k} \cdot 
    \mathbbm{1}[\text{success}_{t,k}]}
    {\sum_{t,k} 
    \mathbbm{1}[\text{success}_{t,k}]}\textcolor{cor}{,}
\end{equation}
where $c_{t,k}$ is the cycle time 
of trial $t$ in experiment $k$. 
Failed trials contribute to the 
success rate but not to the cycle 
time metric.

\subsection*{Tactile Skill Learning Procedure}

The parameter vector $\boldsymbol{\theta} \in \mathbb{R}^{33}$ is optimised by a black-box solver based on~\cite{voigt2020multilevel}. The 
solver maintains a multimodal Gaussian distribution over the parameter space 
from which candidate parameter vectors are proposed. A Support Vector Machine 
approximates the feasible success region in parameter space, filtering 
proposed samples to retain only those estimated to lie within the success 
region prior to execution. This two-component architecture separates 
exploration via the multimodal Gaussian from exploitation of discovered 
success regions via the SVM, enabling efficient search in the 33-dimensional 
parameter space \textcolor{cor}{(for a single arm-gripper scenario)} without requiring gradient information.

All 33 parameters are proposed simultaneously for each trial. The scalar 
reward — cycle time for successful insertions, or 5\,s for failed trials — 
is returned to the solver after each trial. The solver is updated every 10 
trials in a batch regime. Each learning run consists of 160 trials with no early stopping
criterion and is repeated 10 times per task to assess run-to-run
variance. Of the 220 scheduled runs, 9 were incomplete and excluded,
leaving 211 runs and 33,760 usable trials out of the 35,200 scheduled
across all 22 tasks. A task is considered successfully learned
\textcolor{cor2}{if} the framework achieves three consecutive
successful insertions with identical parameters, without any
intervening failure. \textcolor{cor2}{Across the full campaign,
only 4 of 211 full-length runs failed to find a solution, all
concentrated on the two objects with the lowest per-trial
success rates (Schuko outlets and abus-e30). Every task was
solved in the majority of its runs.} This binary criterion
determines whether a task has been solved and forms the basis
for the benchmark results reported across all 22 tasks.
\textcolor{cor2}{We note that this tasks-solved criterion
differs from the trial-averaged success rate reported for the
deep\,RL baselines.}

The parameter space is \textcolor{cor}{the same} across 
all 22 tasks with one task-specific 
adaptation: the feedforward force range 
$F_z$ is expanded from $(0, 20]$\,N 
to $(0, 80]$\,N for Schuko outlet tasks 
(plug Type F), reflecting 
the substantially higher axial force 
required to overcome the spring-loaded 
clamping contacts of this connector 
class.

\subsection*{Physical Effect Taxonomy}\label{sec:physical_effect_taxonomy}

We argue that a structured taxonomy 
of physical effects governs 
contact-rich insertion tasks, 
organised into three concentric 
layers: geometry, material, and 
physical interaction effects 
(Fig.~\ref{fig:main}). As no 
consensus exists on the 
classification of complex non-convex 
geometries, the systematic model 
knowledge study focuses exclusively 
on physical interaction effects. 
A basic examination of geometric 
importance is considered separately 
in the clearance and geometry 
experiment \textcolor{cor}{in Sec. \emph{Limits of Deep Reinforcement Learning}}, which uses 
purpose-manufactured parts with 
simple, controlled cross-sections.

The physical interaction effects 
layer captures emergent contact 
phenomena arising from the 
interaction between geometry, 
material, and insertion dynamics. 
Pressure~(P1) refers to the 
distributed normal force acting 
across the contact interface during 
insertion. Deformation~(P2) 
describes elastic shape change of 
one or both components under 
contact loading, modifying the 
effective clearance and force 
distribution. Interference 
fit~(P3) refers to insertion 
conditions characterised by very 
tight clearances, spring-loaded 
force barriers, or slight geometric 
overlap, producing a force barrier 
that must be overcome for 
successful insertion. Clamping~(P4) 
describes lateral retention forces 
exerted by housing walls or local 
geometric features on the inserted 
component, as present for example 
in phone jacks and key insertions. 
Friction~(P5) refers to \textcolor{cor}{locally} tangential 
resistance forces at the contact 
interface that oppose insertion 
motion and are sensitive to 
approach angle and surface 
condition.

Neurophysiological studies of 
fingertip force control during 
dexterous manipulation provide 
biological grounding for the 
physical effect taxonomy. Tactile 
afferent signals encode normal 
force distribution, contact 
friction, and object compliance 
as distinct fundamental quantities 
regulated by predictive motor 
commands during object 
manipulation~\cite{johansson2009coding}, 
consistent with the identification 
of pressure~(P1), deformation~(P2), 
and friction~(P5) as primary 
physical interaction effects in 
the present taxonomy. The remaining 
effects — interference fit~(P3) 
and clamping~(P4) — are specific 
to constrained insertion tasks and 
do not arise in free-object 
manipulation, reflecting the 
additional physical complexity of 
contact-rich assembly relative to 
everyday dexterous manipulation.

These five effects interact 
combinatorially and nonlinearly; 
their joint effect on task 
difficulty \textcolor{cor}{could not} be predicted from 
any additive combination of 
individual effects, motivating the 
systematic experimental evaluation 
described below.

To enable quantitative analysis of 
how each effect influences task 
difficulty, each of the five physical 
interaction effects was assigned a 
normalised magnitude score in 
$[0, 1]$ per task, determined by 
consensus among five robotics 
engineers holding at minimum a 
Master's degree in mechanical 
engineering or robotics. All 
annotators had hands-on experience 
conducting robot manipulation 
experiments under impedance control; several additionally 
had direct experience with real 
industrial assembly processes and 
components. All five had \textcolor{cor}{expert-level} familiarity 
with contact mechanics, material 
deformation, friction, and tolerance 
fits, and direct experimental 
exposure to the evaluated tasks 
acquired over the course of the 
two-year study. Scores reflect the relative intensity of each 
effect based on geometry, material properties, and contact conditions, and 
were used exclusively for post-hoc regression analysis against observed 
performance metrics.

\subsection*{\textcolor{cor}{Policy Landscape Analysis}}

\textcolor{cor}{To characterise the geometry of successful parameter regions in
the $33$-dimensional Tactile Skill parameter space, the per-task learning
trials were projected onto a one-dimensional ordering using a Hilbert
space-filling curve. Each parameter $\theta_i$ was min--max normalised to
$[0,1]$ independently per task, scaled by that task's observed range across all
trials, with constant parameters mapped to $0$; indices are therefore
comparable within, but not across, tasks.
Each normalised coordinate was quantised to $2^{4}=16$ levels and mapped to a
Hilbert index by standard bit-interleaving with Gray-code decoding. Trials
sorted by this index yield the success/failure structure visualised in
Fig.~\ref{fig:hilbert}. The Hilbert ordering preserves locality in one
direction --- points adjacent in the index are close in parameter space --- but
not conversely; together with the coarse quantisation and per-task scaling, the
ordering therefore reflects within-parameterisation structure rather than a
metric embedding, and because the skill is phase-discretised, the observed
multimodality is in part a property of this parameterisation and its phase
boundaries.}

\textcolor{cor}{To confirm that the observed success/failure separation is a
property of the data rather than an artifact of the projection, it was
re-measured under systematic perturbations of the ordering, quantified by a
Wald--Wolfowitz runs-test $z$-statistic on the labels sorted by the $1$D index
(more negative $z$ indicates longer contiguous success/failure runs; one-sided
significance $p=\Phi(z)$). Narrow, multimodal structure was significant for $14$
of $22$ tasks ($p<0.001$; mean $z=-4.20$), and where present it was stable
across Hilbert resolutions $p=3\text{--}7$ (mean within-task SD of $z=0.70$),
agreed closely between the Hilbert curve and an alternative Z-order (Morton)
curve (mean $|\Delta z|=0.95$; $14/22$ significant under both), and was retained
under a median of $92\%$ of $30$ random dimension-to-axis orderings
(Supplementary Fig.~\ref{fig:hilbert_sensitivity}). The multimodality is
therefore not an artifact of the chosen resolution, space-filling curve, or
dimension ordering; the remaining eight tasks lack \emph{detectable} structure
--- owing to too few successful trials or more diffuse landscapes --- rather
than contradicting the finding.}

\subsection*{Permutation-Based 
Feature Importance Analysis}

To identify which \textcolor{cor}{elements} of
$\boldsymbol{\theta}$ govern
insertion success and execution
efficiency, a feature importance
analysis based on parameter
permutation was conducted
independently per task\textcolor{cor}{, on the 18 of 22
benchmark tasks with sufficient successful trials to train a
task-specific classifier; four tasks were excluded owing to too few successful insertions to
support reliable cross-validation}. For each
\textcolor{cor}{element} $\theta_i \in
\boldsymbol{\theta}$, its values
are randomly permuted across
trials, breaking the association
between that parameter and task
outcome while leaving all other
parameters unchanged. The
resulting decrease in classifier
performance quantifies the
contribution of $\theta_i$ to
predicting the outcome of interest.

Two binary classification targets
were defined. For success
prediction, trials are labelled
as successful or failed. For cost
prediction, only successful trials
are considered; cycle times are
binarised into low-cost (lowest
33rd percentile) and high-cost
(remaining 67\%) classes, applied
independently per task to account
for the wide variation in absolute
cycle times across the \textcolor{cor}{18 analysed}
tasks. \textcolor{cor}{The cost target is therefore imbalanced
33/67 by construction; F1 scoring and stratified splitting are used to account for this, so cost importances are
not directly comparable in magnitude to those of the balanced
success target.}

For each classification target,
an SVM with radial basis function
kernel was trained per task using
randomised hyperparameter search
(\texttt{RandomizedSearchCV},
$n_\text{iter} = 50$, 5-fold
cross-validation, F1 scoring)
over the following ranges:
regularisation parameter
$C \in [10^{-3}, 10^3]$ and kernel
bandwidth $\gamma \in
[10^{-4}, 10^1]$, both sampled
log-uniformly. Features were
standardised to zero mean and unit
variance prior to training. An
80/20 stratified train--test split
was used with fixed random seed
for reproducibility. Permutation
importance was computed on the
held-out test set over
$n_\text{repeats} = 30$
permutations per parameter,
reported as mean decrease in F1
score, following the permutation
importance
method~\cite{fisher2019all}
as implemented in
scikit-learn~\cite{pedregosa2011scikit}.
\textcolor{cor}{Small negative importance values arise when a parameter
carries no predictive signal for the target, so that permuting
it marginally improves held-out F1 through reduced overfitting;
these are interpreted as effectively zero importance. Because
permutation importance distributes attribution arbitrarily among
correlated parameters, importances are interpreted at the level
of the physically meaningful parameter groups (approach offset,
contact, insertion stiffness, oscillatory policy, and
feedforward force) rather than individual dimensions, and a low
individual importance may reflect redundancy with a correlated
parameter rather than irrelevance.}

A complementary failure analysis
was conducted by defining two
failure-specific targets: trials
with no success (failure regardless
of cost) and trials with high cost
given failure. Permutation
importance for these targets
identifies which parameters drive
failure, independently of the
success-oriented analysis\textcolor{cor}{, using the identical
per-task SVM, 5-fold cross-validated hyperparameter search, and
held-out permutation procedure described above}.

\subsection*{Component Damage,  Replacement Protocol and Experiment Order}

\textcolor{cor}{Experiments were executed in loose order of the examined model
knowledge levels, from Level~\textsc{II} to Level~\textsc{VI}, and within each
level the velocity domain preceded the force domain. The algorithm order within
each level and domain was SAC, PPO, then DDPG, although this order was adjusted
for individual combinations where scheduling required. For each
level--domain--algorithm combination, the three independent runs were executed
consecutively, with all 22 tasks running in parallel on their dedicated robots.
When an unrecoverable robot fault or component damage occurred during a run, the
affected task's run was repeated in full at a later time, so that every reported
run corresponds to a complete, uninterrupted execution.}

\textcolor{cor}{Components and robots were inspected before and after every run
for faults and damage. Components were assessed with respect to insertability
and wear---specifically, whether the task's assumed physical interaction
characteristics were still present---and components that were sufficiently
degraded to alter these characteristics were replaced with new parts. Fresh
components were therefore not introduced on a fixed per-condition schedule but
on a condition-triggered basis, when inspection indicated that wear had begun to
change the task. Each trial was additionally recorded by a task-focused camera,
allowing gross destruction events to be identified during and after execution.}

\textcolor{cor}{When damage was identified, either from post-run inspection or,
in ambiguous cases, from examination of the trial recordings, all trials back to
the last inspection at which the component was confirmed intact were discarded
and repeated.}

\subsection*{Clearance and Geometry Experiment}

To characterise the limits of end-to-end deep reinforcement learning as a function of geometric difficulty, a dedicated experiment was conducted using two Franka robotic arms and purpose-manufactured peg-and-hole test parts with controlled cross-sections and clearances. Three cross-sectional shapes (triangular, square, and star, Supplementary Fig.~\ref{fig:clearance_shapes}) were evaluated at four clearance levels (0.0, 0.5, 1.0, and 2.0 mm), yielding twelve distinct geometric conditions. Each condition was trained using SAC in the Interaction Joint Space velocity domain (Level II) over 500 trials. Three independent runs were conducted per condition to ensure statistical reliability.
Unlike the main benchmark, which involves real world components subject to complex physical interaction effects, this experiment isolates the influence of clearance and rotational symmetry on deep RL performance in a controlled setting. The reward function and experimental setup were otherwise identical to those used in the main deep RL experiment.

\subsection*{Task Profile Datasheet}

To support mechanistic interpretation of learned policies,
each task is characterised by a structured task profile
combining the physical effect taxonomy with the learned skill
execution behaviour (Supplementary Information).  The profile is generated automatically 
after each completed learning run and comprises three components.
 
The physical effect profile summarises the normalised magnitudes of the five 
interaction effects (P1--P5) assigned by expert consensus, providing a 
compact fingerprint of the task's physical interaction regime.
 
The trajectory visualisation shows the end-effector path of the best
solution found across the learning runs, projected onto the
\textcolor{cor}{$(x,y)$}, $(x,z)$, and $(y,z)$ planes, separately for
each of the three skill phases. In each panel, the trajectory segment
belonging to the respective phase is shown in full colour, and contact
forces along the trajectory are shown as vectors. Colour coding
distinguishes the parameters most important for success (red) from
those most important for execution cost (orange), as determined by the
permutation-based importance analysis. This visualisation reveals the
spatial structure of the best solution found --- its orientation and
relationship to the contact geometry --- and provides direct evidence
of how the learned approach, contact, and insertion strategies relate
to the physical interaction effects of the task.
 
The parameter summary shows the normalised values of the learned parameter 
vector $\boldsymbol{\theta}$ for the best-performing trial, organised by phase. 
Together with the observations panel, which provides an automatically 
generated natural-language summary of parameter sensitivities derived from 
the importance analysis, the profile constitutes a self-contained record of 
the learned skill that supports both human inspection and cross-task 
comparison. Task profiles for all 22 benchmark tasks are provided in the 
Supplementary Information.

\subsection*{Energy Consumption}
The total energy consumption per task 
is approximately 128.8\,kJ for one 
Tactile Skill learning run and 435.5\,kJ 
for a single deep\,RL run — a factor of 
$3.4\times$ lower, reflecting the 
substantially reduced trial count 
required by the structured 
parameterisation. Robot operation 
dominates total consumption across 
both frameworks; computational costs 
for training and inference are 
negligible by comparison 
(Supplementary Information 
Table~\ref{tab:energy}).

\subsection*{Industrial Assembly Validation}

\begin{figure}[!t]
    \centering
    \includegraphics[width=1.0\textwidth]{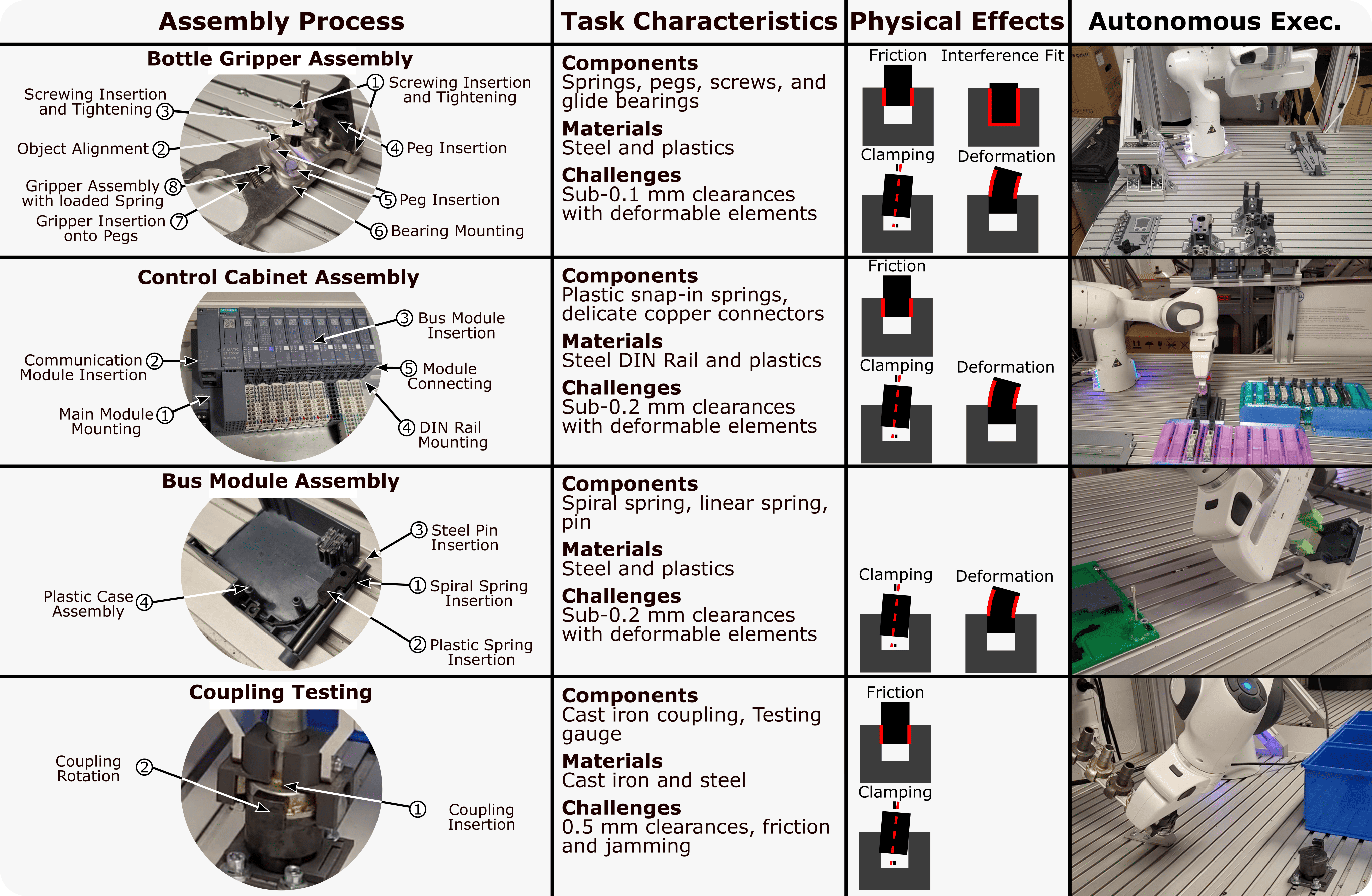}
\caption{%
  \textbf{Industrial assembly validation across four manufacturing partners.}
  For each workflow, the assembly process, \textcolor{cor}{specific task characteristics,} dominant physical interaction
  effects (P1--P5), and fully autonomous robot execution are shown.
}
    \label{fig:assemblies}
\end{figure}

To the authors' knowledge, no prior work has demonstrated tactile manipulation automation at comparable  scale, reliability, execution speed, and real-world assembly complexity.
The Tactile Skill \textcolor{cor}{model} was applied to four
real-world industrial assembly tasks in collaboration with German
industry partners\textcolor{cor}{, denoted Case~1--4 in the
following.}  These assemblies (Fig.~\ref{fig:assemblies}) were selected to represent 
the breadth of contact-rich manufacturing challenges encountered in practice, 
spanning a wide range of part counts, materials, tolerances, and physical 
interaction regimes --- including rigid, flexible, and spring-loaded components. 
All assemblies were executed fully autonomously on a single Franka robotic arm 
without human intervention, using 3D-printed fixtures and, for screwing 
operations, a contact-activated cordless screwdriver. The framework detects 
execution errors in near-real time from learned tactile signatures of successful 
task behaviour, enabling automatic retry without operator intervention; retry 
cycles are included in all reported cycle times (see 
Fig.~\ref{fig:industrial_learning}). 
Non-contact motions, grasp poses, and start and goal poses were
kinesthetically taught by a human operator; all contact-rich subtasks
were learned autonomously by the Tactile Skill \textcolor{cor}{model} framework.
Reliability was assessed over 50 
consecutive full assembly runs per task. Across a two-arm configuration operating continuously for 24\,h
\textcolor{cor}{(projected by linear scaling from measured single-arm cycle
times; see Productivity Analysis)}, the framework achieves an
\textcolor{cor}{estimated} 150--200\% of human daily throughput and 400\% of
human productivity per unit labour cost. Human 
baseline cycle times were recorded by certified technicians experienced in the 
respective assembly tasks under normal working conditions, and serve as the 
reference for all throughput and productivity comparisons.

\textcolor{cor}{\textbf{Bottle Gripper Package Assembly.}}
The gripper packages assembled in this task are installed in
high-throughput bottling lines, where they grip and guide
bottles through filling and capping stations. Each package
comprises 12 parts --- including stainless steel and plastic
components, a spring, roller bearings, and precision clamps
--- assembled across 13 unique (17 in total) contact-rich
subtasks: 8 insertions, 2 screwing operations, and 3 pressing
steps, with part tolerances below 0.05\,mm. The primary
challenges are the extremely tight tolerances combined with
simultaneous multi-constraint contact: the spring must be
compressed and held against its stored elastic energy while
the robot simultaneously inserts into two bearings without
losing grip, any significant force deviation causing the spring to escape
or the assembly to jam. Reliable engagement of tight screw
threads and the handling of deformable plastic bearings add
further difficulty. Setup required 172 kinesthetically
programmed poses ($\sim$8\,h) and approximately 3\,h of
learning time. Non-contact motions are handled by programmed
trajectories; only contact-rich steps are learned. Robot
cycle time is approximately 4\,min\,20\,s versus
approximately 1\,min\,5\,s for a human worker, with 100\%
reliability over 50 runs.

\textcolor{cor}{\textbf{Control Cabinet Assembly.}} This task
involves assembling groups of bus modules used in industrial
control cabinets, each consisting of a base module and one or
more plug-in modules with sensitive pin contacts and plastic
housings, with part tolerances below 0.2\,mm. The primary
challenge is force sensitivity: pin contacts bend permanently
under lateral forces of only a few newtons, and snap-fit
housings crack under excessive insertion force, making every
failure irreversible. Reliable assembly therefore demands
precise force control throughout, with little margin for
exploratory contact. The assembly involves 5 unique subtasks
(23 in total), requiring 18 programmed poses ($\sim$60\,min)
and approximately 60\,min of learning time. Robot cycle time
is approximately 5\,min\,50\,s with 100\% reliability.

\textcolor{cor}{\textbf{Bus Module Assembly.}} This task
requires assembling a bus module consisting of two flexible
plastic shells, an aluminium pin, and two springs --- with
nearly all parts being compliant. The primary challenge is
deformability: both springs deform significantly during
assembly and must be manipulated in a controlled manner, as
even small unintended forces or deformations can cause a spring to escape
and project several metres due to the released elastic
energy. Additionally, the highly deformable housing requires
two hooks to be inserted simultaneously into small apertures,
demanding precise bilateral force coordination. The
prevalence of deformable components introduces contact
dynamics not well captured by rigid-body models, making this
task a representative challenge for tactile learning. The
assembly requires 9 learned subtasks and 29 programmed poses
($\sim$90\,min), with approximately 100\,min of learning
time. Robot cycle time is approximately 1\,min\,15\,s versus
approximately 20\,s for a human worker, with $\geq$98\%
reliability.

\textcolor{cor}{\textbf{Hydraulic Coupling Tactile Testing.}}
Rather than a pure assembly task, this application automates
the quality inspection of hydraulic couplings. Couplings are
fed to the robot, which grasps each part, compliantly inserts
it into a test gauge, rotates it, and evaluates its
correctness based on measured forces and motions before
sorting it into the appropriate bin. While the simplest of
the four cases in terms of workflow complexity, the task
still demands precise tactile force sensing to distinguish
between coupling variants and to align correctly with the
test gauge despite geometric variability across parts. The
task requires 2 learned subtasks and 6 programmed poses
($\sim$15\,min), with approximately 20\,min of learning time.
Robot cycle time is approximately 15\,s versus approximately
5\,s for a human worker, with 100\% reliability.

Links to videos of the respective automation results are provided in the 
Supplementary Information.

\subsection*{\textcolor{cor}{Productivity} Analysis}
Robot operational costs were estimated from a unit hardware cost
of approximately USD\,40,000, based on a 2022 robot price of
USD\,30,000 including gripper and controller, scaled by a
conservative ancillary cost multiplier of 1.33 to account for
mounting hardware, safety documentation, and consumables. This
multiplier is intentionally \textcolor{cor}{kept} conservative
relative to full industrial cobot deployments, which typically
require additional integration infrastructure such as vision
systems, custom tooling, and cell
integration~\cite{evsint2026cobot}; none of these were required
in this work. The \textcolor{cor}{used robot} platform has since
been succeeded by the Franka Research~3.
Power consumption \textcolor{cor}{was measured to be}
200\,W continuous (80\,W controller, 120\,W robot arm), yielding
an energy cost of approximately USD\,800 over the robot lifetime
at USD\,0.20\,kWh$^{-1}$, consistent with average industrial
electricity prices across high-wage economies \textcolor{cor}{(2023--2025
averages; EU average: USD\,0.222\,kWh$^{-1}$; Germany:
USD\,0.284\,kWh$^{-1}$; North America:
USD\,0.161\,kWh$^{-1}$)~\cite{globalpetrolprices2025}}. Energy
cost is negligible relative to hardware cost.
\textcolor{cor}{A setup and personnel budget of USD\,20,000 was
assumed, corresponding to approximately 100 engineer-hours at a
blended rate of USD\,100\,h$^{-1}$ for mechanical integration,
electrical commissioning, and safety documentation, plus
USD\,10,000 for miscellaneous tooling, cabling, and contingency.
This reflects a minimal but realistic deployment scenario without
custom cell integration.} This yields a total estimated system
cost of approximately USD\,60,000.
Productivity per unit labour cost was computed as the ratio of
daily output to hourly labour cost, benchmarked against national
minimum wages~\cite{oecd2024wages} and assembly worker wages in
several countries (Fig.~\ref{fig:industrial_learning}).
Throughput comparisons assume a two-arm parallel configuration
operating continuously for 24\,h, scaled linearly from
single-arm measurements. Under this assumption, a single robot
arm operating 24\,h achieves approximately 75\% of human daily
throughput over an 8-hour shift; a two-arm configuration
achieves approximately 150--200\% of human daily throughput.

\subsection*{Methods-Time Measurement Analysis and Human Baseline}
\label{sec:mtm}

Methods-Time Measurement (MTM) is a predetermined \textcolor{cor}{internationally standardized} motion time system
that decomposes manual work into fundamental human motions and assigns
empirically derived standard time values to each~\cite{maynard1948mtm,
niebel2003methods}. Times are expressed in Time Measurement Units (TMU):

\begin{equation}
  1\ \text{TMU} = 0.036\ \text{s} = 6 \times 10^{-4}\ \text{min}
  \label{eq:tmu}
\end{equation}

\noindent For each motion, the corresponding TMU value is added; in the case
of simultaneous actions, only the longest duration counts toward the total.
Complete motion-level tables for all assemblies are provided in the
supplementary data file.

To validate the MTM estimates, assembly times were measured \textcolor{cor}{independently} with certified
skilled workers, each performing five repetitions per assembly. Both measured
mean times and MTM estimates are reported in
Table~\ref{tab:industrial_assemblies}.
Measured times were consistent with MTM predictions, deviating by at most
4\% above or 8\% below the predicted values.

\subsection*{Experimental Time Expenditure}

The full study comprised three experimental campaigns conducted over a total timespan of approximately two years (Table~\ref{tab:timetable}). Preparation tasks are largely non-recurring per campaign; core experiment tasks persisted throughout each campaign's runtime. Campaign~1 (Tactile Skill learning) involved 22 robots across 22
objects, with 35,200 scheduled trials (of which 33,760 were used in
the evaluation) over 3~days of active robot runtime, requiring
approximately 767\,h of preparation and 172\,h of core experiment time. Campaign 2 (Deep RL) involved the same 22 robots and objects across 693,000 trials over 68 days of active robot runtime, with approximately 755\,h of preparation and 15,470\,h of core experiment time — the dominant cost of the full study. Campaign 3 (Clearance and geometry experiment) involved 2 robots across 12 objects with 18,000 trials over 10 days, requiring approximately 50\,h of preparation and 88\,h of core experiment time. Across all three campaigns, total preparation time was approximately 1,572\,h and total core experiment time approximately 15,730\,h.

Prior to the experimental campaigns, the 
robot collective required a one-time 
infrastructure setup investment of 
approximately 1,890\,h 
(Table~\ref{tab:infrastructure}), covering 
software systems and 
scripting, and individual robot setup 
(${\sim}1{,}390$\,h) — and hardware 
installation, including robot tables, 
cabling, object production, and mounting 
(${\sim}500$\,h).

\subsection*{Limitations and Scope}

The experimental results are obtained under a set of controlled assumptions that constrain the generality of the conclusions.

\textbf{Known goal pose.} All experiments assume a fixed, pre-taught goal pose per task. The framework does not \textcolor{cor}{require} explicit object localisation or pose estimation, and robustness to pose uncertainty is not evaluated. In practical deployments, such pose information may be provided by an external perception system, but such integration is not assessed here.

\textcolor{cor2}{\textbf{Single robot platform.} All experiments are conducted on the Franka Emika
robot~\cite{haddadin2022franka}. The Franka Emika Panda provides integrated joint torque sensing and torque-level control, making it well suited to this study and enabling experimentation at the required scale. While this introduces
embodiment-specific parameters that may influence absolute
performance values, to our knowledge no other platform
provides comparable characteristics for contact-rich
manipulation research. Its wide availability in laboratories
worldwide further supports reproducibility, and its
comparatively low cost was what made experimentation at this
scale feasible. Addressing the implications for platform
generalisation nonetheless remains an important direction for
future work.}

\textcolor{cor}{\textbf{Fixed fixture assumption.} Objects are rigidly fixtured
with repeatable placement; upstream manipulation stages such as
bin-picking, part feeding, or in-hand reorientation are not
considered. We acknowledge that this does not generalise
straightforwardly to unfixtured or dual-arm manipulation
scenarios. However, the physical interaction effects examined
here are intrinsic to the contact task itself and present
regardless of the manipulation platform or upstream pipeline.
The principle that structured model knowledge is necessary for
reliable contact-rich manipulation should therefore hold in
broader settings, though the specific policy structure and
parameters would likely differ.}

\textbf{Parameterisation design.} The 33-dimensional Tactile Skill parameterisation is derived from the GGTWreP framework~\cite{Johannsmeier2025ProcessCentricTaxonomy} and encodes structured knowledge about insertion tasks. Alternative parameterisations\textcolor{cor}{, including
Dynamic Movement Primitives~\cite{schaal2006dmp},
Probabilistic Movement Primitives~\cite{paraschos2013promp},
and neural policy representations such as Diffusion
Policy~\cite{chi2023diffusion},} are not evaluated and may
yield different learning efficiency or success characteristics;
\textcolor{cor}{however, this does not alter the central
finding that structured model knowledge improves learning
performance and is a prerequisite for feasible learning of
meaningful manipulation in the physical world.}

\textbf{Statistical power of deep reinforcement learning baselines.} Each deep RL configuration is evaluated over three independent runs per task--algorithm--level combination. This corresponds to approximately 693{,}000 executed trials over 68 \textcolor{cor}{net} days of continuous robot operation across 22 robotic systems. Given the scale of the experimental campaign, additional replications were impractical at this scale; the categorical success boundary, however, is evident within individual tasks and does not depend on aggregate statistical power. Results are therefore interpreted at the aggregate level across tasks and algorithms.

\textbf{Physical effect annotation.} 
The P1--P5 physical effect magnitudes 
used in regression analyses are based 
on expert annotation by five \textcolor{cor}{well trained} robotics 
engineers with direct experience of 
the evaluated tasks. No formal 
inter-rater reliability metrics were 
computed, reflecting the inherent 
difficulty of objectively measuring 
physical interaction effects during 
contact-rich insertion — quantities 
such as deformation magnitude, 
interference fit force, and clamping 
load are process-dependent phenomena 
that resist standardised measurement 
under \textcolor{cor}{actual} real-world task conditions, 
consistent with the observation that 
robot performance on physical 
interaction tasks is fundamentally 
process-specific and not reducible 
to static capability 
metrics~\cite{kirschner2025treeofrobots}. 
Accordingly, the regression results 
should be interpreted as capturing 
qualitative trends in task difficulty 
rather than exact quantitative 
mappings between effect magnitude 
and policy performance. Direct measurement of individual 
effect magnitudes during insertion 
is not feasible under real-world 
task conditions, as contact forces 
reflect the combined and inseparable 
contribution of all five effects 
simultaneously.

\textcolor{cor2}{\textbf{Explicit versus learned models.} This study does not establish that no future architecture
could acquire the relevant physical models from experience.
Recurrent policies, active system-identification methods, or
self-modelling approaches that learn robot morphology,
kinematics, and dynamics directly from 
proprioception~\cite{diazledezma2023selfdiscovery,
kwiatkowski2019taskagnostic} may learn aspects of the
embodiment, interaction, and process structure that are
explicitly encoded here. Such a result would be consistent
with, not contrary to, our conclusion: reliable contact-rich
learning requires these models to be represented, whether
they are supplied in advance or acquired during learning.}

%% file: sections/supplementary.tex
\section*{Supplementary Information}

This document provides supplementary 
figures and tables referenced in the 
main text, followed by task profile 
datasheets for all 22 benchmark tasks.

\begin{table}[h]
\centering
\caption{Symbol glossary for the 
model knowledge hierarchy 
(Table~\ref{tab:model_tree}).}
\label{tab:symbol_glossary}
\small
\begin{tabular}{lll}
\hline
\textbf{Symbol} & 
\textbf{Description} & 
\textbf{Unit} \\
\hline
\multicolumn{3}{l}{
  \textit{Robot state}} \\
$\boldsymbol{q}$ & 
  Joint positions & rad \\
$\dot{\boldsymbol{q}}$ & 
  Joint velocities & rad\,s$^{-1}$ \\
$\ddot{\boldsymbol{q}}$ & 
  Joint accelerations 
  (not available) & 
  rad\,s$^{-2}$ \\
$\boldsymbol{x}$ & 
  Cartesian end-effector pose & 
  m, rad \\
$\dot{\boldsymbol{x}}$ & 
  Cartesian end-effector 
  velocity & 
  m\,s$^{-1}$, rad\,s$^{-1}$ \\
\hline
\multicolumn{3}{l}{
  \textit{Forces and torques}} \\
$\boldsymbol{\tau}_d$ & 
  Desired joint torques 
  (control output) & N\,m \\
$\hat{\boldsymbol{\tau}}_\mathrm{ext}$ & 
  Estimated external joint torques 
  via momentum observer & N\,m \\
$\hat{\boldsymbol{\mathcal{F}}}_\mathrm{ext}$ & 
  Estimated external Cartesian 
  wrench & N, N\,m \\
\hline
\multicolumn{3}{l}{
  \textit{Dynamic model terms}} \\
$\boldsymbol{M}(\boldsymbol{q})$ & 
  Joint-space inertia matrix & 
  kg\,m$^2$ \\
$\hat{\boldsymbol{c}}(\boldsymbol{q}, 
  \dot{\boldsymbol{q}})$ & 
  Estimated Coriolis and 
  centrifugal terms & N\,m \\
$\hat{\boldsymbol{g}}(\boldsymbol{q})$ & 
  Estimated gravity compensation 
  torques & N\,m \\
$\boldsymbol{r}$ & 
  Residual unmodelled dynamics & 
  N\,m \\
\hline
\multicolumn{3}{l}{
  \textit{Kinematics}} \\
$\boldsymbol{J}(\boldsymbol{q})$ & 
  Geometric Jacobian matrix & — \\
$\boldsymbol{J}^\top$ & 
  Jacobian transpose & — \\
$\boldsymbol{J}^\dagger$ & 
  Moore--Penrose pseudoinverse, 
  $\boldsymbol{J}^\top
  (\boldsymbol{J}
  \boldsymbol{J}^\top)^{-1}$ 
  & — \\
$\boldsymbol{T}(\boldsymbol{q})$ & 
  Forward kinematics mapping 
  to Cartesian pose & — \\
IK & 
  Inverse kinematics & — \\
\hline
\multicolumn{3}{l}{
  \textit{Policies}} \\
$\boldsymbol{\pi}_{\dot{\boldsymbol{q}}}$ & 
  Joint velocity policy & 
  rad\,s$^{-1}$ \\
$\boldsymbol{\pi}_{\boldsymbol{\tau}}$ & 
  Joint torque policy & N\,m \\
$\boldsymbol{\pi}_{\dot{\boldsymbol{x}}}$ & 
  Cartesian twist policy & 
  m\,s$^{-1}$, rad\,s$^{-1}$ \\
$\boldsymbol{\pi}_{\boldsymbol{\mathcal{F}}}$ & 
  Cartesian wrench policy & 
  N, N\,m \\
\hline
\multicolumn{3}{l}{
  \textit{Control}} \\
FBC$(\cdot)$ & 
  Feedback controller; a PD 
  controller is one realisation & — \\
\hline
\end{tabular}
\end{table}
\newpage
\textcolor{cor}{\subsection*{Definitions}}
\textcolor{cor}{\textbf{Definition 1} (Tactile Skill). \textit{A tactile skill integrates (1) a
\textbf{tactile policy}, which encodes coordinated desired wrench and twist
commands, and drives (2) a \textbf{tactile controller} that simultaneously
regulates and tracks compliance and contact force by generating inputs to
(3) a \textbf{tactile platform} that represents the low-level joint dynamics
integrating the actual physical system and delivering the percept vector
$\boldsymbol{\Omega}$ to measure the performance $\mathcal{Q}$ that is used
to (4) \textbf{learn} the policy and control parameters $\boldsymbol{\theta}_\pi$
and $\boldsymbol{\theta}_c$, respectively.}}

\bigskip

\textcolor{cor}{\textbf{Definition 2} (Task). \textit{A task process $p$ is
specified by the process operations needed to achieve the process objective
and the process requirements. It comprises four process states: initial state
$s_0$, error state $s_e$, final state $s_1$, and policy state $s_\pi$. The
process requirements are formalized by transitions $\delta \in \Delta$
connecting the process operations and three boundary conditions:}
\begin{itemize}
    \item \textit{The \textbf{precondition}
    $\mathcal{C}_\mathrm{pre}(\mathcal{O}) =
    c_{1,\mathrm{pre}}(\mathcal{O}) \wedge \cdots \wedge
    c_{n,\mathrm{pre}}(\mathcal{O})$
    checks whether the process is ready to start and switches from
    $s_0$ to $s_\pi$.}
    \item \textit{The \textbf{error condition}
    $\mathcal{C}_\mathrm{err}(\mathcal{O}) =
    c_{1,\mathrm{err}}(\mathcal{O}) \vee \cdots \vee
    c_{n,\mathrm{err}}(\mathcal{O})$
    is triggered by an irreversible failure, immediately terminating
    the process and entering the error state $s_e$ from $s_0$ or
    $s_\pi$.}
    \item \textit{The \textbf{success condition}
    $\mathcal{C}_\mathrm{suc}(\mathcal{O}) =
    c_{1,\mathrm{suc}}(\mathcal{O}) \wedge \cdots \wedge
    c_{n,\mathrm{suc}}(\mathcal{O})$
    indicates successful execution of the process, triggering a switch
    from $s_0$ or $s_\pi$ to $s_1$.}
\end{itemize}
}
\textcolor{cor}{\subsection*{Figures and Tables}}

\begin{table*}[t]
\centering
\caption{\textbf{Layers of model knowledge in contact-rich manipulation
learning.} The three model-knowledge layers---embodiment, interaction and
process---are properties of the control \emph{representation} and define the
model-knowledge hierarchy evaluated in this work (Levels~\textsc{II}--\textsc{VI}).
Each layer specifies a distinct class of physical structure that may be supplied
explicitly or acquired by a learning mechanism, but that cannot be absent from a
representation capable of reliable contact-rich manipulation. The
search/learning mechanism is listed separately and is not itself model
knowledge.}
\label{tab:model-knowledge-layers}
\small
\setlength{\tabcolsep}{6pt}
\renewcommand{\arraystretch}{1.25}
\begin{tabularx}{\textwidth}{@{} l >{\raggedright\arraybackslash}p{4.4cm} X c @{}}
\toprule
\textbf{Layer} & \textbf{Knowledge represented} & \textbf{Representative elements} &
\textbf{Level(s)} \\
\midrule
\textbf{Embodiment} &
Physical structure and dynamics of the robotic system &
Kinematics and Jacobian $\boldsymbol{J}(\boldsymbol{q})$; inertia
$\hat{\boldsymbol{M}}(\boldsymbol{q})$; gravity, Coriolis and friction
compensation $\hat{\boldsymbol{\gamma}},\hat{\boldsymbol{\mu}}$; external wrench
estimate $\hat{\boldsymbol{\mathcal{F}}}_{\mathrm{ext}}$ &
\textsc{II}--\textsc{VI} \\
\addlinespace[2pt]
\textbf{Interaction} &
Relationship between robot motion and exchanged effort &
Joint- vs.\ task-space representation; effort/flow variables (wrench/twist);
impedance control; physically meaningful action ranges and scaling &
\textsc{III}--\textsc{VI} \\
\addlinespace[2pt]
\textbf{Process} &
Temporal and physical organisation of the manipulation task &
Approach, contact and insertion phases; phase transitions; compliant
force--motion regulation; feedforward force; oscillatory search &
\textsc{VI} \\
\midrule
\multicolumn{4}{@{}p{\dimexpr\textwidth-2\tabcolsep\relax}@{}}{%
\footnotesize\emph{Search/learning mechanism} (not model knowledge): identifies
the task-dependent parameters that remain once the structure is fixed---PPO,
SAC and DDPG for the deep-RL levels; the multimodal-Gaussian sampler with an SVM
feasibility model for the Tactile~Skill.} \\
\bottomrule
\end{tabularx}
\end{table*}

\definecolor{secgray}{gray}{0.82}
\definecolor{colTested}{RGB}{180,220,180}   
\definecolor{colPossible}{RGB}{200,218,240}
\definecolor{colNo}{RGB}{240,200,200}          

\definecolor{lvl0}{RGB}{245,245,245}
\definecolor{lvl1}{RGB}{220,230,245}
\definecolor{lvl2}{RGB}{195,215,240}
\definecolor{lvl3}{RGB}{170,200,235}
\definecolor{lvl4}{RGB}{145,185,230}
\definecolor{lvl5}{RGB}{120,170,225}
\definecolor{lvl6}{RGB}{95,155,220}
 

\renewcommand{\arraystretch}{1.25}

\begin{sidewaystable}
\centering
\scriptsize
\captionsetup{margin={0.0cm, -0cm}}
\setlength{\tabcolsep}{2.5pt}
\caption{Model knowledge hierarchy for contact-rich manipulation \textcolor{cor}{reinforcement} learning.
\textit{Assumptions}: torque-controlled robot; $\textcolor{cor}{\boldsymbol{q}}$, $\dot{\textcolor{cor}{\boldsymbol{q}}}$, $\textcolor{cor}{\boldsymbol{\tau}}_J$ available;
$\ddot{\textcolor{cor}{\boldsymbol{q}}}$ not available; $\hat{\textcolor{cor}{\boldsymbol{\tau}}}_\mathrm{ext}$ \textcolor{cor}{requires}  $\textcolor{cor}{\boldsymbol{M}}(\textcolor{cor}{\boldsymbol{q}})$.
$\textcolor{cor}{\boldsymbol{J}}^\dagger := \textcolor{cor}{\boldsymbol{J}^\top(\boldsymbol{JJ}^\top)^{-1}}$ (Moore--Penrose pseudoinverse).
\colorbox{colTested}{\rule{0pt}{1.0ex}\  } tested in this work;\;
\hspace*{-0.13cm}\colorbox{colPossible}{\rule{0pt}{1.0ex}\  } feasible, not tested;\;
\hspace*{-0.13cm}\colorbox{colNo}{\rule{0pt}{1.0ex\ }} not feasible.
Blue shading of \textbf{Modeling}: darker $=$ more model knowledge. A--D mark infeasible preliminary levels, excluded from the methods; I--V are the levels discussed therein. All variables, operators, and abbreviations are defined in
Table~\ref{tab:symbol_glossary}.}
\label{tab:model_tree}

\begin{adjustbox}{max width=\linewidth}
\begin{tabular}{L{2.6cm}!{\color{black!25}\vrule} L{0.6cm}!{\color{black!25}\vrule} L{2.15cm}!{\color{black!25}\vrule} L{2.6cm}!{\color{black!25}\vrule} L{4.8cm}!{\color{black!25}\vrule} L{0.7cm}!{\color{black!25}\vrule} L{4.0cm}!{\color{black!25}\vrule} L{1.2cm}!{\color{black!25}\vrule} L{3.8cm}}
\toprule
\textbf{Name} & \textbf{Level} & \textbf{Modeling} & \textbf{Control paradigm} & \textbf{Controller} & \textbf{Policy} & \textbf{Implicitly req. learning objective} & \textbf{Meas.} & \textbf{Physically feasible} \\
\midrule
\rowcolor{secgray}\multicolumn{9}{l}{\textbf{Joint-Space}} \\
\arrayrulecolor{black!25}\hline\arrayrulecolor{black}

Torque &
A &
\hspace*{-0.13cm}\colorbox{white}{\parbox{2.2cm}{\RaggedRight ---}} &
Joint torque policy &
$\boldsymbol{\tau}_d = \boldsymbol{\pi}_{\boldsymbol{\tau}}$ &
$\boldsymbol{\pi}_{\boldsymbol{\tau}}$ &
Dynamics ($\boldsymbol{c}$, $\boldsymbol{g}$); residual $\boldsymbol{r}$; $\boldsymbol{J}(\boldsymbol{q})$ for task-space motion &
$\boldsymbol{q},\dot{\boldsymbol{q}}$ &
\hspace*{-0.13cm}\colorbox{colNo}{\parbox{3.6cm}{\RaggedRight No. Cannot support itself; hardware and object damage.}} \\
\arrayrulecolor{black!25}\hline\arrayrulecolor{black}

Velocity &
A+ &
\hspace*{-0.13cm}\colorbox{lvl1}{\parbox{2.2cm}{\RaggedRight FBC }} &
Joint vel.\ policy &
$\boldsymbol{\tau}_d = \mathrm{FBC}\!\left(\int\boldsymbol{\pi}_{\dot{\boldsymbol{q}}}\,dt\right)$ &
$\boldsymbol{\pi}_{\dot{\boldsymbol{q}}}$ &
Dynamics ($\boldsymbol{c}$, $\boldsymbol{g}$); residual $\boldsymbol{r}$; $\boldsymbol{J}(\boldsymbol{q})$ for task-space motion &
$\boldsymbol{q},\dot{\boldsymbol{q}}$ &
\hspace*{-0.13cm}\colorbox{colNo}{\parbox{3.6cm}{\RaggedRight No. No gravity compensation; robot cannot support its own weight; \textcolor{cor}{inevitable} hardware damage.}} \\
\arrayrulecolor{black!25}\hline\arrayrulecolor{black}

Torque Force Feedback &
B &
\hspace*{-0.13cm}\colorbox{lvl2}{\parbox{2.2cm}{\RaggedRight$\hat{\boldsymbol{M}}(\boldsymbol{q})$}} &
Joint \textcolor{cor}{torque}\ policy $+$ $\hat{\boldsymbol{\tau}}_\mathrm{ext}$ &
$\boldsymbol{\tau}_d = \boldsymbol{\pi}_{\boldsymbol{\tau}}$ &
$\boldsymbol{\pi}_{\boldsymbol{\tau}}$ &
Dynamics ($\boldsymbol{c}$, $\boldsymbol{g}$); residual $\boldsymbol{r}$; $\boldsymbol{J}(\boldsymbol{q})$ for task-space motion &
$\boldsymbol{q},\dot{\boldsymbol{q}},\hat{\boldsymbol{\tau}}_\mathrm{ext}$ &
\hspace*{-0.13cm}\colorbox{colNo}{\parbox{3.6cm}{\RaggedRight No. Immediate collapse; $\boldsymbol{M}(\boldsymbol{q})$ observer cannot substitute for compensation.}} \\
\arrayrulecolor{black!25}\hline\arrayrulecolor{black}

Velocity Force Feedback &
B &
\hspace*{-0.13cm}\colorbox{lvl2}{\parbox{2.2cm}{\RaggedRight FBC $\wedge$ $\hat{\boldsymbol{M}}(\boldsymbol{q})$}} &
Joint vel.\ policy $\wedge$ $\hat{\boldsymbol{\tau}}_\mathrm{ext}$ &
$\boldsymbol{\tau}_d = \mathrm{FBC}\!\left(\int\boldsymbol{\pi}_{\dot{\boldsymbol{q}}}\,dt\right)$ &
$\boldsymbol{\pi}_{\dot{\boldsymbol{q}}}$ &
Dynamics ($\boldsymbol{c}$, $\boldsymbol{g}$); residual $\boldsymbol{r}$; $\boldsymbol{J}(\boldsymbol{q})$ for task-space motion &
$\boldsymbol{q},\dot{\boldsymbol{q}},\hat{\boldsymbol{\tau}}_\mathrm{ext}$ &
\hspace*{-0.13cm}\colorbox{colNo}{\parbox{3.6cm}{\RaggedRight No. Uncompensated gravity/Coriolis; collapses regardless of inertia observer.}} \\
\arrayrulecolor{black!25}\hline\arrayrulecolor{black}

Torque Compensated &
I &
\hspace*{-0.13cm}\colorbox{lvl3}{\parbox{2.2cm}{\RaggedRight $\hat{\boldsymbol{c}}~\wedge~\hat{\boldsymbol{g}}$}} &
Joint torque policy &
$\boldsymbol{\tau}_d = \boldsymbol{\pi}_{\boldsymbol{\tau}} + \hat{\boldsymbol{c}} + \hat{\boldsymbol{g}}$ &
$\boldsymbol{\pi}_{\boldsymbol{\tau}}$ &
Residual $\boldsymbol{r}$; $\boldsymbol{J}(\boldsymbol{q})$ for task-space motion &
$\boldsymbol{q},\dot{\boldsymbol{q}}$ &
\hspace*{-0.13cm}\colorbox{colPossible}{\parbox{3.6cm}{\RaggedRight \textbf{Possible}, not tested.}} \\
\arrayrulecolor{black!25}\hline\arrayrulecolor{black}

Velocity Compensated &
I &
\hspace*{-0.13cm}\colorbox{lvl3}{\parbox{2.2cm}{\RaggedRight FBC $\wedge$ $\hat{\boldsymbol{c}}~\wedge~\hat{\boldsymbol{g}}$}} &
Joint vel.\ policy &
$\boldsymbol{\tau}_d = \mathrm{FBC}\!\left(\int\boldsymbol{\pi}_{\dot{\boldsymbol{q}}}\,dt\right) + \hat{\boldsymbol{c}} + \hat{\boldsymbol{g}}$ &
$\boldsymbol{\pi}_{\dot{\boldsymbol{q}}}$ &
Residual $\boldsymbol{r}$; $\boldsymbol{J}(\boldsymbol{q})$ for task-space motion &
$\boldsymbol{q},\dot{\boldsymbol{q}}$ &
\hspace*{-0.13cm}\colorbox{colPossible}{\parbox{3.6cm}{\RaggedRight \textbf{Possible}, not tested. Self-supporting; standard impedance-control config.}} \\
\arrayrulecolor{black!25}\hline\arrayrulecolor{black}

\textbf{Torque Compensated Force Feedback} &
II &
\hspace*{-0.13cm}\colorbox{lvl4}{\parbox{2.2cm}{\RaggedRight $\hat{\boldsymbol{c}}~\wedge~\hat{\boldsymbol{g}}$ $\wedge$ $\hat{\boldsymbol{M}}(\boldsymbol{q})$}} &
Joint torque policy $\wedge$ $\hat{\boldsymbol{\tau}}_\mathrm{ext}$ &
$\boldsymbol{\tau}_d = \boldsymbol{\pi}_{\boldsymbol{\tau}} + \hat{\boldsymbol{c}} + \hat{\boldsymbol{g}}$ &
$\boldsymbol{\pi}_{\boldsymbol{\tau}}$ &
Residual $\boldsymbol{r}$ &
$\boldsymbol{q},\dot{\boldsymbol{q}},\hat{\boldsymbol{\tau}}_\mathrm{ext}$ &
\hspace*{-0.13cm}\colorbox{colTested}{\parbox{3.6cm}{\RaggedRight \textbf{Yes}, tested.}} \\
\arrayrulecolor{black!25}\hline\arrayrulecolor{black}
\textbf{Velocity Compensated Force Feedback} &
II &
\hspace*{-0.13cm}\colorbox{lvl4}{\parbox{2.2cm}{\RaggedRight FBC $\wedge$ $\hat{\boldsymbol{c}}~\wedge~\hat{\boldsymbol{g}}$ $\wedge$ $\hat{\boldsymbol{M}}(\boldsymbol{q})$}} &
Joint \textcolor{cor}{vel.} policy $\wedge$ $\hat{\boldsymbol{\tau}}_\mathrm{ext}$ &
$\boldsymbol{\tau}_d = \mathrm{FBC}\!\left(\int\boldsymbol{\pi}_{\dot{\boldsymbol{q}}}\,dt\right) + \hat{\boldsymbol{c}} + \hat{\boldsymbol{g}}$ &
$\boldsymbol{\pi}_{\dot{\boldsymbol{q}}}$ &
Residual $\boldsymbol{r}$; $\boldsymbol{J}(\boldsymbol{q})$ for task-space motion &
$\boldsymbol{q},\dot{\boldsymbol{q}},\hat{\boldsymbol{\tau}}_\mathrm{ext}$ &
\hspace*{-0.13cm}\colorbox{colTested}{\parbox{3.6cm}{\RaggedRight \textbf{Yes}, tested. Full compensation $+$ ext.-force feedback.}} \\
\arrayrulecolor{black!25}\hline\arrayrulecolor{black}

\midrule
\rowcolor{secgray}\multicolumn{9}{l}{\textbf{Cartesian-Space}} \\
\arrayrulecolor{black!25}\hline\arrayrulecolor{black}

Wrench &
C &
\hspace*{-0.13cm}\colorbox{lvl1}{\parbox{2.2cm}{\RaggedRight $\boldsymbol{J}$}} &
Wrench policy &
$\boldsymbol{\tau}_d = \boldsymbol{J}^\top(\boldsymbol{q})\,\boldsymbol{\pi}_{\boldsymbol{\mathcal{F}}}$ &
$\boldsymbol{\pi}_{\boldsymbol{\mathcal{F}}}$ &
Dynamics ($\boldsymbol{c}$, $\boldsymbol{g}$); residual $\boldsymbol{r}$; $\boldsymbol{M}(\boldsymbol{q})$ for $\hat{\boldsymbol{\mathcal{F}}}_\mathrm{ext}$ &
$\boldsymbol{x},\dot{\boldsymbol{x}}$ &
\hspace*{-0.13cm}\colorbox{colNo}{\parbox{3.6cm}{\RaggedRight No. No compensation; robot cannot sustain itself.}} \\
\arrayrulecolor{black!25}\hline\arrayrulecolor{black}

Twist &
C &
\hspace*{-0.13cm}\colorbox{lvl1}{\parbox{2.2cm}{\RaggedRight FBC $\wedge$ IK $\wedge~\boldsymbol{J}$}} &
Twist policy &
$\boldsymbol{\tau}_d = \boldsymbol{J}^\top(\boldsymbol{q})\,\mathrm{FBC}\!\left(\int\boldsymbol{\pi}_{\dot{\boldsymbol{x}}}\,dt\right)$ &
$\boldsymbol{\pi}_{\dot{\boldsymbol{x}}}$ &
Dynamics ($\boldsymbol{c}$, $\boldsymbol{g}$); residual $\boldsymbol{r}$; $\boldsymbol{M}(\boldsymbol{q})$ for $\hat{\boldsymbol{\mathcal{F}}}_\mathrm{ext}$ &
$\boldsymbol{x},\dot{\boldsymbol{x}}$ &
\hspace*{-0.13cm}\colorbox{colNo}{\parbox{3.6cm}{\RaggedRight No. Collapse without gravity/Coriolis; same as joint-level FBC.}} \\
\arrayrulecolor{black!25}\hline\arrayrulecolor{black}

Wrench Force Feedback &
D &
\hspace*{-0.13cm}\colorbox{lvl2}{\parbox{2.2cm}{\RaggedRight $\boldsymbol{J}$ $\wedge$ $\hat{\boldsymbol{M}}(\boldsymbol{q})$}} &
Wrench policy $\wedge$ $\hat{\boldsymbol{\mathcal{F}}}_\mathrm{ext}$ &
$\boldsymbol{\tau}_d = \boldsymbol{J}^\top(\boldsymbol{q})\,\boldsymbol{\pi}_{\boldsymbol{\mathcal{F}}}$ &
$\boldsymbol{\pi}_{\boldsymbol{\mathcal{F}}}$ &
Dynamics ($\boldsymbol{c}$, $\boldsymbol{g}$); residual $\boldsymbol{r}$; &
$\boldsymbol{x},\dot{\boldsymbol{x}},\hat{\boldsymbol{\mathcal{F}}}_\mathrm{ext}$ &
\hspace*{-0.13cm}\colorbox{colNo}{\parbox{3.6cm}{\RaggedRight No. Uncompensated gravity/Coriolis; wrench policy unstabilisable.}} \\
\arrayrulecolor{black!25}\hline\arrayrulecolor{black}

Twist Force Feedback &
D &
\hspace*{-0.13cm}\colorbox{lvl2}{\parbox{2.2cm}{\RaggedRight FBC $\wedge$ IK $\wedge~\boldsymbol{J}$ $\wedge$ $\hat{\boldsymbol{M}}(\boldsymbol{q})$}} &
Twist policy $\wedge$ $\hat{\boldsymbol{\mathcal{F}}}_\mathrm{ext}$ &
$\boldsymbol{\tau}_d = \boldsymbol{J}^\top(\boldsymbol{q})\,\mathrm{FBC}\!\left(\int\boldsymbol{\pi}_{\dot{\boldsymbol{x}}}\,dt\right)$ &
$\boldsymbol{\pi}_{\dot{\boldsymbol{x}}}$ &
Dynamics ($\boldsymbol{c}$, $\boldsymbol{g}$); residual $\boldsymbol{r}$; &
$\boldsymbol{x},\dot{\boldsymbol{x}},\hat{\boldsymbol{\mathcal{F}}}_\mathrm{ext}$ &
\hspace*{-0.13cm}\colorbox{colNo}{\parbox{3.6cm}{\RaggedRight No. Collapse regardless of ext.-force observer.}} \\
\arrayrulecolor{black!25}\hline\arrayrulecolor{black}

Wrench Compensated &
E &
\hspace*{-0.13cm}\colorbox{lvl3}{\parbox{2.2cm}{\RaggedRight $\boldsymbol{J}$ $\wedge$ $\hat{\boldsymbol{c}}~\wedge~\hat{\boldsymbol{g}}$}} &
Wrench policy &
$\boldsymbol{\tau}_d = \boldsymbol{J}^\top(\boldsymbol{q})\,\boldsymbol{\pi}_{\boldsymbol{\mathcal{F}}} + \hat{\boldsymbol{c}} + \hat{\boldsymbol{g}}$ &
$\boldsymbol{\pi}_{\boldsymbol{\mathcal{F}}}$ &
Residual $\boldsymbol{r}$; $\boldsymbol{M}(\boldsymbol{q})$ for $\hat{\boldsymbol{\mathcal{F}}}_\mathrm{ext}$ &
$\boldsymbol{x},\dot{\boldsymbol{x}}$ &
\hspace*{-0.13cm}\colorbox{colPossible}{\parbox{3.6cm}{\RaggedRight \textbf{Possible}, not tested. Self-supporting wrench policy.}} \\
\arrayrulecolor{black!25}\hline\arrayrulecolor{black}

Twist Compensated &
E &
\hspace*{-0.13cm}\colorbox{lvl3}{\parbox{2.2cm}{\RaggedRight FBC $\wedge$ IK $\wedge~\boldsymbol{J}$ $\wedge$ $\hat{\boldsymbol{c}}~\wedge~\hat{\boldsymbol{g}}$}} &
Twist policy &
$\boldsymbol{\tau}_d = \boldsymbol{J}^\top(\boldsymbol{q})\,\mathrm{FBC}\!\left(\int\boldsymbol{\pi}_{\dot{\boldsymbol{x}}}\,dt\right) + \hat{\boldsymbol{c}} + \hat{\boldsymbol{g}}$ &
$\boldsymbol{\pi}_{\dot{\boldsymbol{x}}}$ &
Residual $\boldsymbol{r}$; $\boldsymbol{M}(\boldsymbol{q})$ for $\hat{\boldsymbol{\mathcal{F}}}_\mathrm{ext}$ &
$\boldsymbol{x},\dot{\boldsymbol{x}}$ &
\hspace*{-0.13cm}\colorbox{colPossible}{\parbox{3.6cm}{\RaggedRight \textbf{Possible}, not tested. Task-space analogue of joint-space FBC$+\hat{\boldsymbol{c}}+\hat{\boldsymbol{g}}$.}} \\
\arrayrulecolor{black!25}\hline\arrayrulecolor{black}

\textbf{Wrench Compensated Force Feedback} &
III - V &
\hspace*{-0.13cm}\colorbox{lvl4}{\parbox{2.2cm}{\RaggedRight $\boldsymbol{J}$ $\wedge$ $\hat{\boldsymbol{c}}~\wedge~\hat{\boldsymbol{g}}$ $\wedge$ $\hat{\boldsymbol{M}}(\boldsymbol{q})$}} &
Wrench policy $\wedge$ $\hat{\boldsymbol{\mathcal{F}}}_\mathrm{ext}$ &
$\boldsymbol{\tau}_d = \boldsymbol{J}^\top(\boldsymbol{q})\,\boldsymbol{\pi}_{\boldsymbol{\mathcal{F}}} + \hat{\boldsymbol{c}} + \hat{\boldsymbol{g}}$ &
$\boldsymbol{\pi}_{\boldsymbol{\mathcal{F}}}$ &
Residual $\boldsymbol{r}$ &
$\boldsymbol{x},\dot{\boldsymbol{x}},\hat{\boldsymbol{\mathcal{F}}}_\mathrm{ext}$ &
\hspace*{-0.13cm}\colorbox{colTested}{\parbox{3.6cm}{\RaggedRight \textbf{Yes}, tested.}} \\
\arrayrulecolor{black!25}\hline\arrayrulecolor{black}
\textbf{Twist Compensated Force Feedback} &
III - V &
\hspace*{-0.13cm}\colorbox{lvl4}{\parbox{2.2cm}{\RaggedRight FBC $\wedge$ IK $\wedge~\boldsymbol{J}$ $\wedge$ $\hat{\boldsymbol{c}}~\wedge~\hat{\boldsymbol{g}}$ $\wedge$ $\hat{\boldsymbol{M}}(\boldsymbol{q})$}} &
Twist policy $\wedge$ $\hat{\boldsymbol{\mathcal{F}}}_\mathrm{ext}$ &
$\boldsymbol{\tau}_d = \boldsymbol{J}^\top(\boldsymbol{q})\,\mathrm{FBC}\!\left(\int\boldsymbol{\pi}_{\dot{\boldsymbol{x}}}\,dt\right) + \hat{\boldsymbol{c}} + \hat{\boldsymbol{g}}$ &
$\boldsymbol{\pi}_{\dot{\boldsymbol{x}}}$ &
Residual $\boldsymbol{r}$ &
$\boldsymbol{x},\dot{\boldsymbol{x}},\hat{\boldsymbol{\mathcal{F}}}_\mathrm{ext}$ &
\hspace*{-0.13cm}\colorbox{colTested}{\parbox{3.6cm}{\RaggedRight \textbf{Yes}, tested.}} \\
\arrayrulecolor{black!25}\hline\arrayrulecolor{black}

\bottomrule
\end{tabular}
\end{adjustbox}  
\end{sidewaystable}

\begin{figure}[!h]
    \centering
    \includegraphics[width=1.0\textwidth]{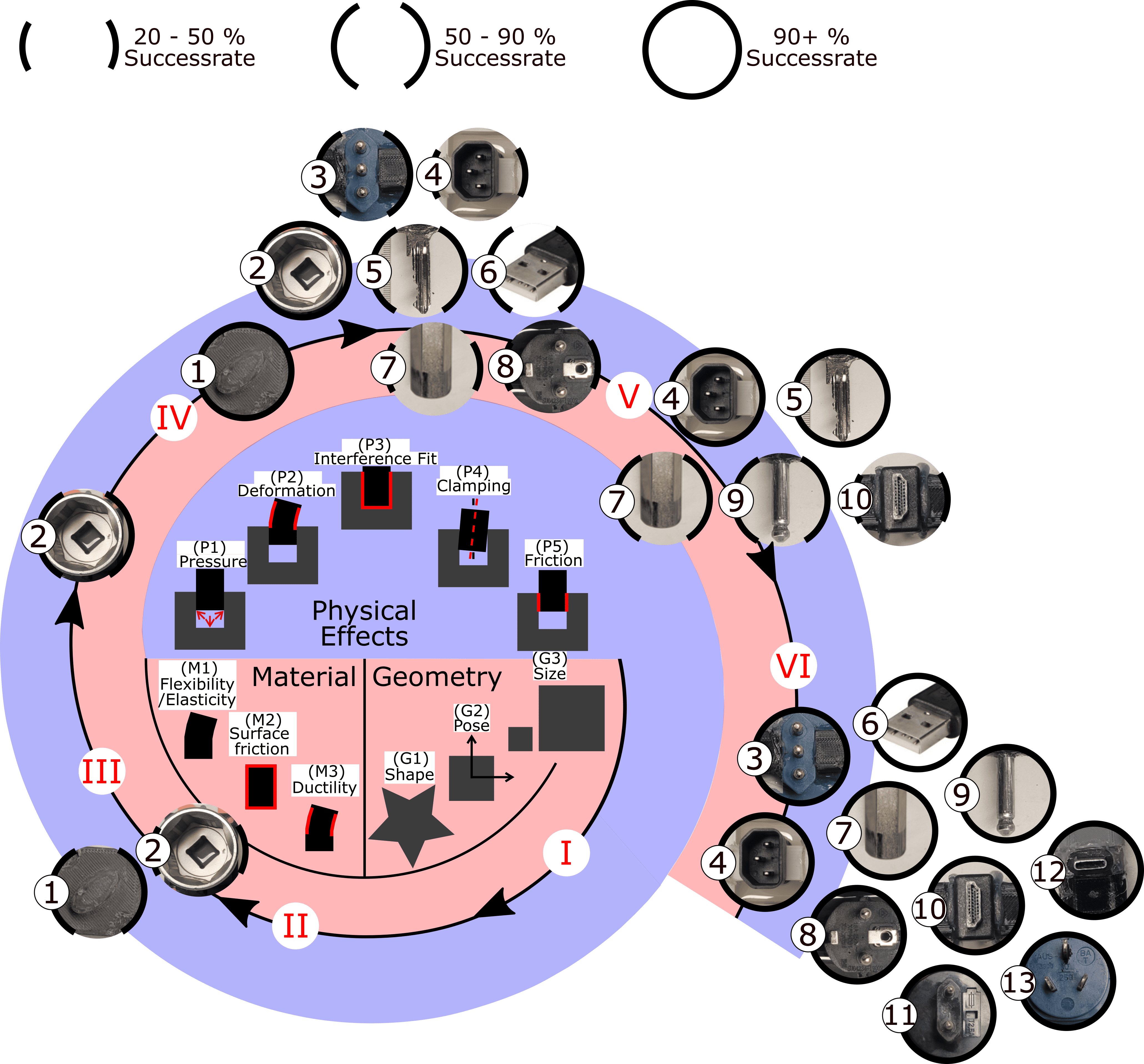}
\textcolor{cor}{\caption{%
  \textbf{Representative benchmark tasks and their learning
  outcomes.}
  Each object is shown with a ring indicating the best
  deep\,RL success rate achieved across all evaluated levels
  and algorithms: 1/3 ring ($20$--$50\%$), 2/3 ring
  ($50$--$90\%$), and full ring ($\geq90\%$). Red Roman
  numerals indicate the model knowledge level
  (Levels~\textsc{II}--\textsc{VI}) at which the shown
  performance is achieved. Objects:
  1)~3D-printed ellipsoid,
  2)~6-point socket,
  3)~Type-l plug,
  4)~IEC-C13 connector,
  5)~Padlock key,
  6)~USB-A,
  7)~Hex key,
  8)~Schuko plug,
  9)~6.35\,mm audio jack,
  10)~HDMI,
  11)~Europlug,
  12)~USB-C,
  13)~GB-2099 plug.
  Tasks with high physical effect magnitudes predominantly
  achieve only low success rates under deep\,RL, while
  geometrically simpler tasks with lower effect profiles
  reach moderate to high success rates. All tasks achieve
  100\% success with the Tactile Skill model
  (Level~\textsc{VI}), independent of physical effect profile.%
}}
    \label{fig:main_objects}
\end{figure}

\begin{figure}[!h]
    \centering
    \includegraphics[width=1.0\textwidth]{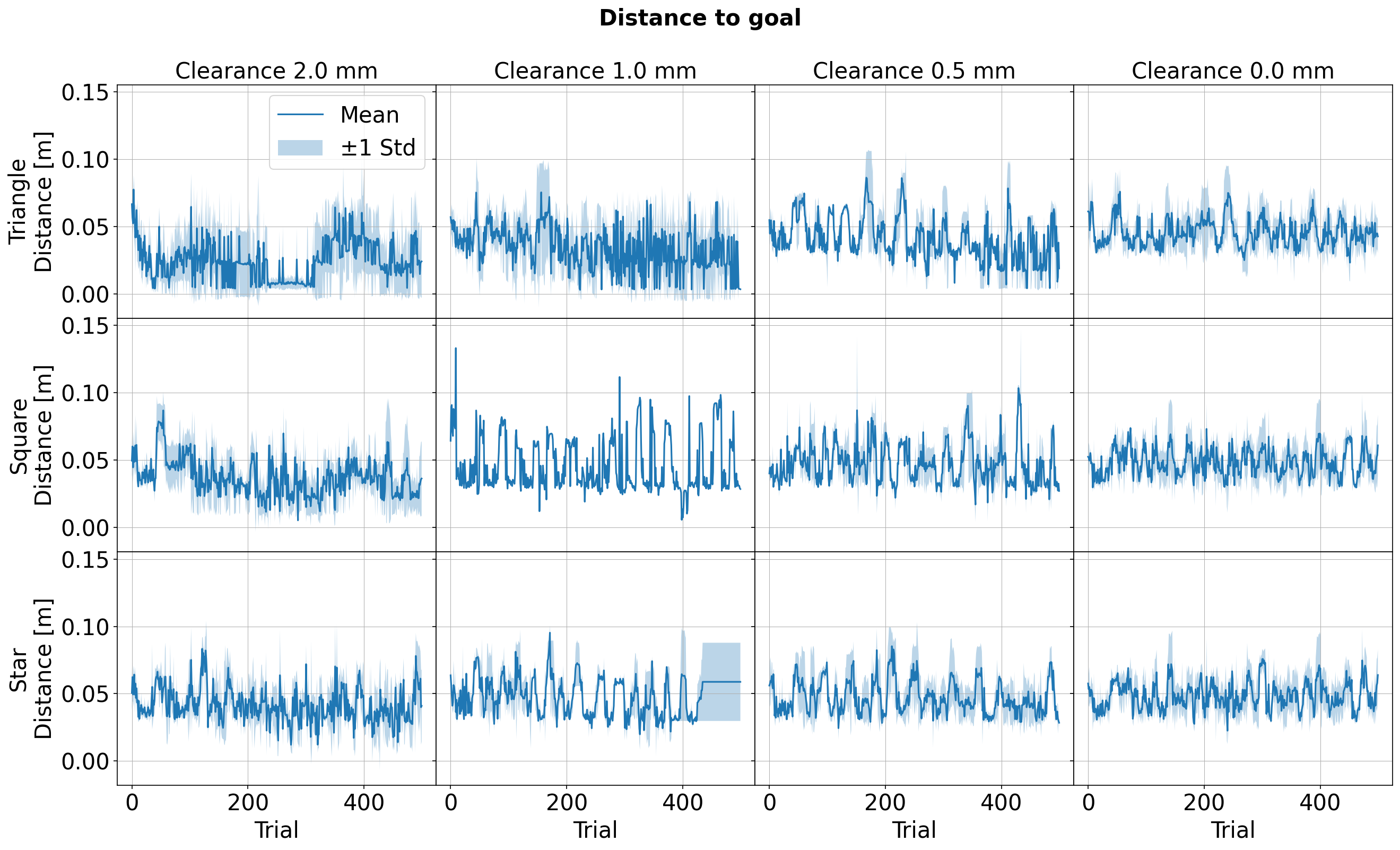}
\caption{%
\textbf{Distance-to-goal trajectories over 500 training
trials for the same shape--clearance conditions as
Fig.~\ref{fig:drl_cg_success}.}
Mean final Euclidean distance to the goal position
$\pm1$\,std is shown per trial, averaged across three
independent runs. For the triangular shape, a
statistically meaningful reduction in distance to goal is
observed for up to 1.0\,mm clearance, with learning rate
appearing correlated with decreasing clearance.
For the square and star, a marginal reduction
is visible at 2.0\,mm; below this threshold the trajectories
show no systematic improvement and are consistent with exploratory noise.
}
    \label{fig:drl_cg_distance}
\end{figure}

\begin{figure}[t]
  \centering
  \includegraphics[width=\textwidth]{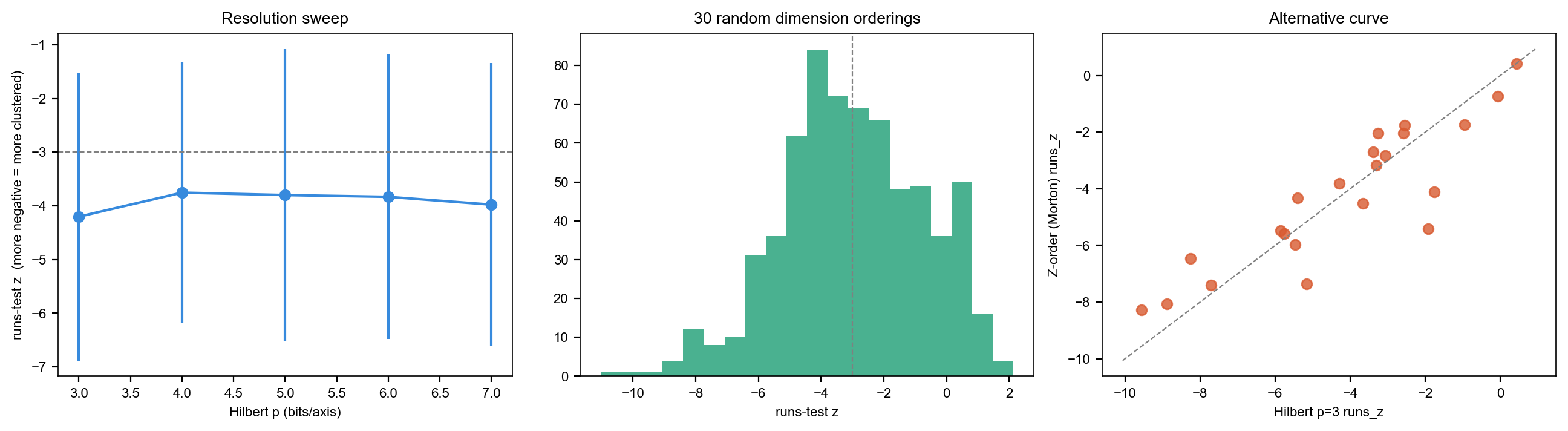}
  \caption{\textcolor{cor}{\textbf{Robustness of the policy-landscape
    separation to the 1D projection.} Separation is measured by the runs-test
    $z$ (more negative $=$ more clustered; dashed line at $z=-3$).
    \textbf{Left:} resolution sweep---mean $z$ across objects is stable for
    Hilbert $p=3\text{--}7$. \textbf{Centre:} distribution of $z$ over $30$
    random dimension orderings across all objects, the majority remaining
    clustered. \textbf{Right:} Z-order (Morton) versus Hilbert $z$ per object,
    lying close to the identity (dashed), showing the two space-filling curves
    agree. Together these indicate the multimodal success structure is not an
    artifact of the chosen ordering.}}
  \label{fig:hilbert_sensitivity}
\end{figure}

\begin{figure}[!t]
    \centering
    \includegraphics[width=1.0\textwidth]{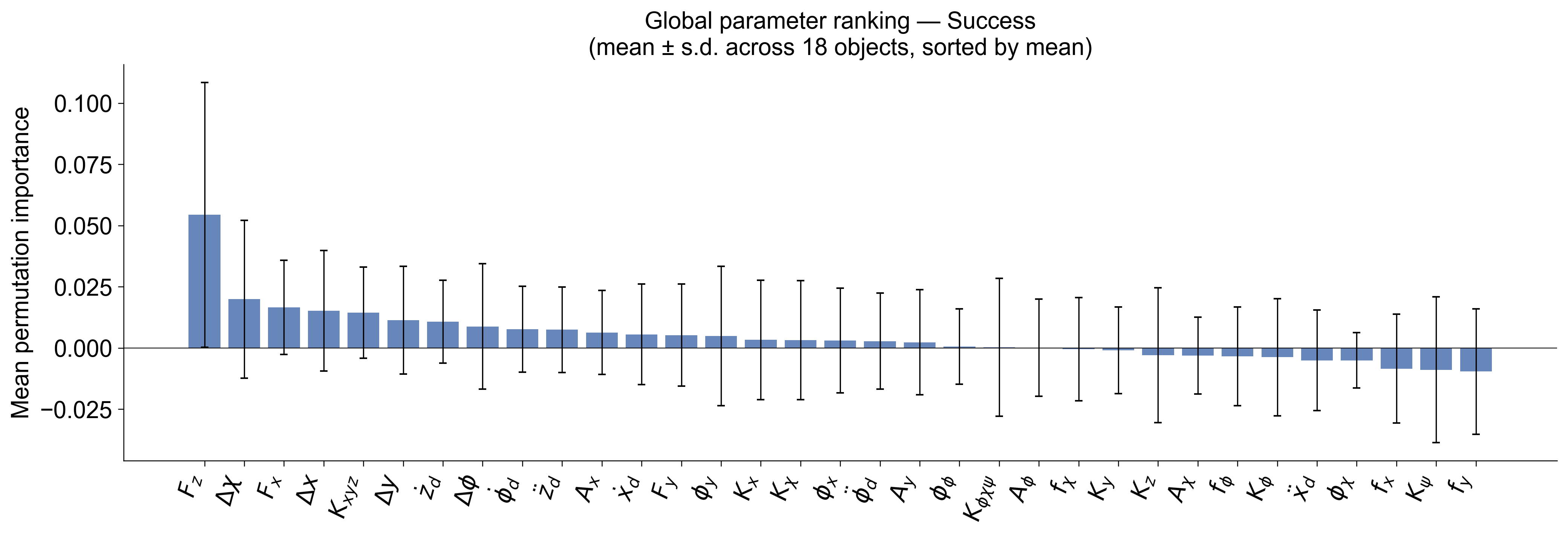}
\caption{%
  \textbf{Global permutation-based 
  feature importance ranking for 
  insertion success across 18 tasks.}
  Each bar shows the mean decrease 
  in F1 score upon permutation of 
  parameter $\theta_i \in 
  \boldsymbol{\theta}$, reflecting 
  its importance for predicting 
  insertion success; results are 
  averaged across 18 tasks and 
  independent runs and sorted by 
  mean importance; error bars 
  indicate $\pm1$\,s.d.\ across 
  tasks.
  Feedforward force 
  magnitude~$F_\mathrm{z}$ is the dominant 
  predictor of insertion success, 
  reflecting its role in overcoming 
  force barriers, friction, and 
  clamping loads; lateral approach 
  offset~$\Delta\chi$ and constant 
  lateral force~$F_\mathrm{x}$ are the next 
  most important parameters.%
}
    \label{fig:ranking_success}
\end{figure}

\begin{figure}[!t]
    \centering
    \includegraphics[width=1.0\textwidth]{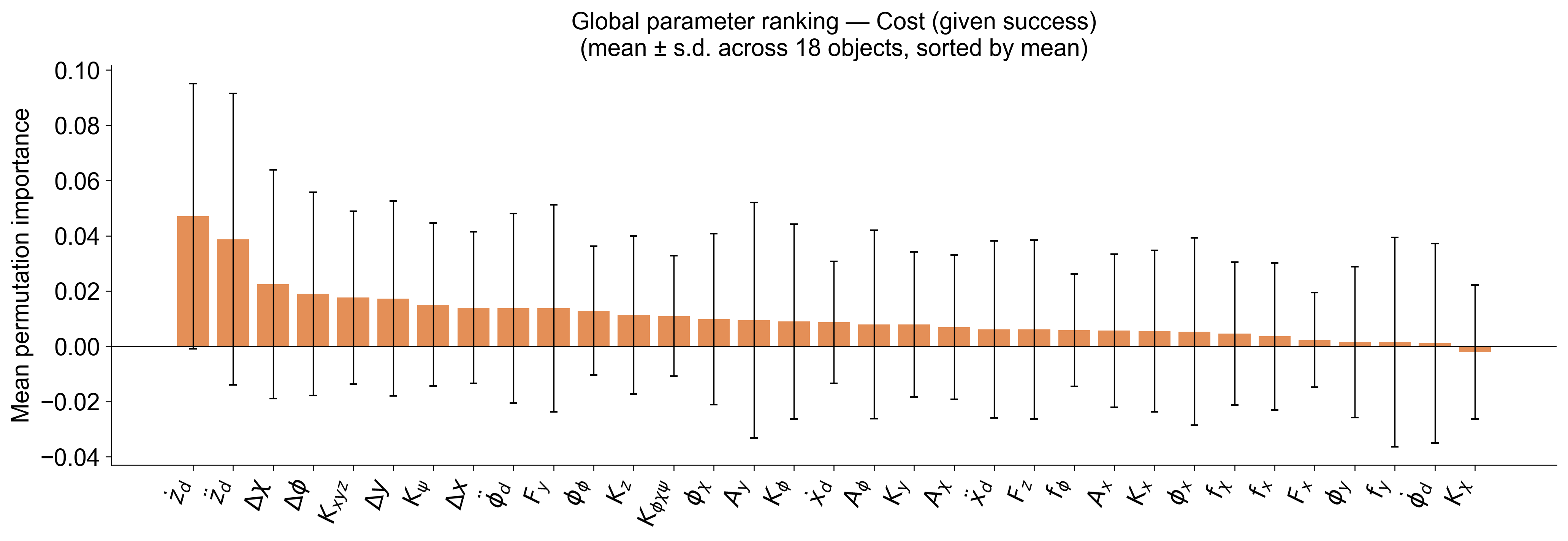}
\caption{%
  \textbf{Global permutation-based 
  feature importance ranking for 
  execution cost given success 
  across 18 tasks.}
  Each bar shows the mean decrease 
  in F1 score upon permutation of 
  parameter $\theta_i \in 
  \boldsymbol{\theta}$, reflecting 
  its importance for predicting 
  execution cost; results are 
  averaged across 18 tasks and 
  independent runs and sorted by 
  mean importance; error bars 
  indicate $\pm1$\,s.d.\ across 
  tasks.
  Contact approach 
  velocity~$\dot{\mathrm{z}}_d$ is the 
  dominant predictor of execution 
  speed, reflecting that the rate 
  of initial contact establishment 
  is the primary determinant of 
  cycle time once a successful 
  insertion strategy is in place.%
}
    \label{fig:ranking_cost}
\end{figure}

\begin{figure}[!t]
    \centering
    \includegraphics[width=1.0\textwidth]{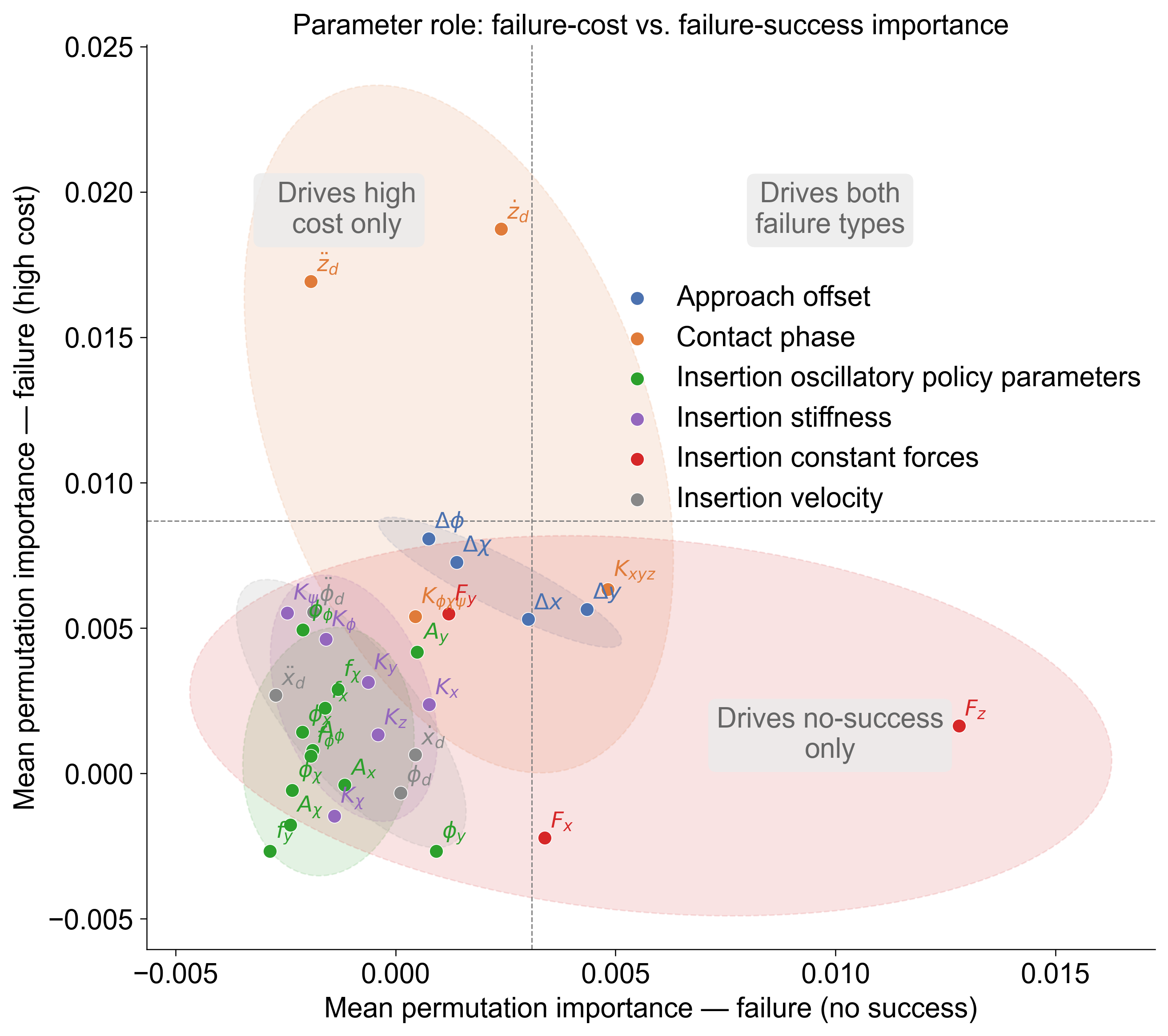}
    \caption{%
  \textbf{Per-task permutation-based feature importance for insertion
  failure across a representative subset of benchmark tasks.}
  Each point represents one Tactile Skill parameter, plotted \textcolor{cor}{for} its
  mean permutation importance for failure given no success ($x$-axis)
  against failure given high cost ($y$-axis).
  Feedforward force~$F_\mathrm{z}$ remains the strongest failure predictor:
  insufficient axial force is the primary failure mode across tasks.
  The remaining failure-driving parameters differ from the success
  rankings, indicating that failure arises from distinct parameter
  combinations rather than uniformly low values of success-critical
  parameters.%
}
    \label{fig:scatter_failure}
\end{figure}

\begin{figure}[!t]
    \centering
    \includegraphics[width=1.0\textwidth]{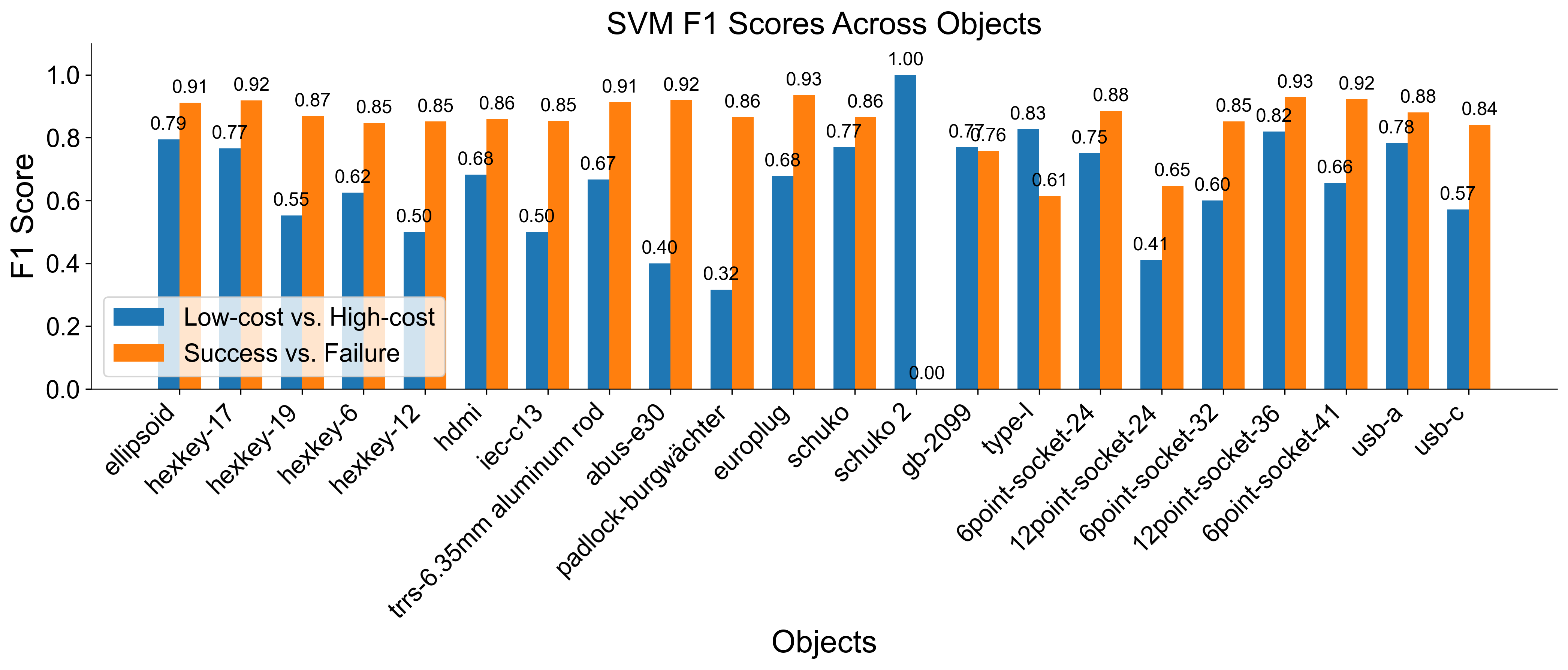}
    \caption{%
  \textbf{SVM classifier accuracy for success/failure and cost
  classification across all 22 benchmark tasks with sufficient
  successful trials.}
  Orange bars show F1 score for success vs.\ failure classification;
  blue bars show F1 score for low-cost vs.\ high-cost classification,
  both computed on the Tactile
  Skill trial data.
  Success vs.\ failure classification achieves F1 scores of
  0.77--1.00 across tasks, indicating that the parameter space contains
  learnable structure that the SVM reliably captures.
  Cost classification is substantially more variable (0.32--0.93),
  reflecting the weaker and more task-specific relationship between
  parameter values and execution speed given success.
  Tasks are ordered by task index as used throughout this work.%
}
    \label{fig:svm_f1}
\end{figure}

\begin{table}[htbp]
  \centering
  \caption{Insertion Speed [Ex.\,s$^{{-1}}$] -- Best of PPO / SAC per task}
  \label{tab:speed_best}
  \small
  \setlength{\tabcolsep}{4pt}
  \begin{tabular}{lrrrrrrrr}
    \toprule
         & \multicolumn{3}{c}{Force Domain} & \multicolumn{4}{c}{Velocity Domain} & VI. Tactile \\
        \cmidrule(lr){2-4} \cmidrule(lr){5-8} \cmidrule(lr){9-9}
        Object [Ex.\,s$^{{-1}}$] & II & III & IV & II & III & IV & V & Skill \\ \midrule
        1. 12point-socket-24 & {--} & {--} & 0.89 & {--} & {--} & 0.88 & 0.62 & 1.01 \\
        2. 12point-socket-36 & 0.69 & 0.78 & 1.03 & {--} & {--} & 1.19 & 0.41 & 0.96 \\
        3. 6point-socket-24 & 1.26 & 3.01 & 0.82 & 1.61 & 1.51 & 1.01 & 0.96 & 0.85 \\
        4. 6point-socket-32 & 0.91 & 0.45 & 1.61 & {--} & 0.91 & 0.98 & 1.22 & 0.98 \\
        5. 6point-socket-41 & 0.38 & 0.33 & 0.88 & {--} & 0.80 & 0.87 & 0.68 & 0.95 \\
        6. abus-e30 & {--} & {--} & {--} & {--} & 0.54 & 0.72 & 0.54 & 0.74 \\
        7. ellipsoid & 0.53 & 0.44 & 1.85 & 0.68 & {--} & 1.27 & 1.13 & 1.12 \\
        8. europlug & {--} & {--} & 1.71 & {--} & {--} & {--} & 0.67 & 0.99 \\
        9. gb-2099 & {--} & {--} & 0.72 & {--} & {--} & 0.49 & 0.48 & 1.16 \\
        10. hdmi & {--} & {--} & {--} & 1.44 & 0.72 & 0.90 & 0.66 & 1.23 \\
        11. hexkey-12 & {--} & {--} & 0.59 & 0.46 & 0.52 & {--} & 0.39 & 1.05 \\
        12. hexkey-17 & {--} & {--} & {--} & 0.34 & 1.07 & 1.36 & 1.26 & 1.26 \\
        13. hexkey-19 & {--} & {--} & {--} & {--} & {--} & 1.06 & {--} & 0.95 \\
        14. hexkey-6 & {--} & {--} & {--} & 0.82 & 0.88 & {--} & {--} & 1.21 \\
        15. iec-c13 & {--} & 0.53 & 0.71 & {--} & 0.67 & 0.88 & 0.91 & 0.88 \\
        16. padlock-burgw\"{a}chter & {--} & {--} & 2.79 & {--} & {--} & 0.47 & 0.91 & 1.01 \\
        17. schuko & {--} & {--} & 0.66 & {--} & {--} & 1.06 & 0.74 & 0.84 \\
        18. schuko 2 & {--} & {--} & 0.92 & {--} & {--} & 0.48 & 0.67 & 0.77 \\
        19. trrs-6.35 mm aluminum rod & {--} & {--} & 0.64 & 0.44 & 0.82 & {--} & 0.76 & 1.06 \\
        20. type-l & {--} & {--} & {--} & {--} & {--} & 0.74 & 0.84 & 1.08 \\
        21. usb-a & 1.09 & 1.14 & 1.08 & 0.73 & 1.23 & 0.70 & 1.16 & 1.12 \\
        22. usb-c & {--} & 1.68 & 0.34 & 1.10 & 1.67 & 0.69 & 0.61 & 1.25 \\
        \midrule
        \textbf{Mean} & \textbf{0.22} & \textbf{0.38} & \textbf{0.78} & \textbf{0.35} & \textbf{0.51} & \textbf{0.72} & \textbf{0.71} & \textbf{1.02} \\
        \textcolor{cor}{\textit{Adj.\ mean$^\dagger$}} & \textit{0.81} & \textit{1.04} & \textit{1.08} & \textit{0.85} & \textit{0.94} & \textit{0.88} & \textit{0.78} & \textit{1.02} \\
    \\\bottomrule
  \end{tabular}
  \smallskip\par\noindent\footnotesize{$^\dagger$Adjusted mean: computed over tasks with at least one successful trial only; {--} tasks excluded.}
\end{table}

\begin{table}[htbp]
  \centering
  \caption{Success Rate [\%] -- Proximal Policy Optimization}
  \label{tab:success_ppo}
  \small
  \setlength{\tabcolsep}{4pt}
  \begin{tabular}{lrrrrrrrr}
    \toprule
         & \multicolumn{3}{c}{Force Domain} & \multicolumn{4}{c}{Velocity Domain} & VI. Tactile \\
        \cmidrule(lr){2-4} \cmidrule(lr){5-8} \cmidrule(lr){9-9}
        Object [\%] & II & III & IV & II & III & IV & V & Skill \\ \midrule
        1. 12point-socket-24 & 0.0 & 0.0 & 14.3 & 0.0 & 0.0 & 14.0 & 17.0 & \textcolor{green!60!black}{100.0} \\
        2. 12point-socket-36 & 23.0 & 70.0 & 15.7 & 0.0 & 0.0 & 59.7 & \textcolor{green!60!black}{100.0} & \textcolor{green!60!black}{100.0} \\
        3. 6point-socket-24 & 1.3 & 0.3 & 22.0 & 7.3 & 6.3 & 46.0 & 29.0 & \textcolor{green!60!black}{100.0} \\
        4. 6point-socket-32 & 1.3 & 0.0 & \textcolor{blue}{95.7} & 0.0 & 0.3 & 88.0 & \textcolor{blue}{98.0} & \textcolor{green!60!black}{100.0} \\
        5. 6point-socket-41 & 32.3 & 16.7 & 82.0 & 0.0 & 0.3 & 86.0 & \textcolor{blue}{94.7} & \textcolor{green!60!black}{100.0} \\
        6. abus-e30 & 0.0 & 0.0 & 0.0 & 0.0 & 0.3 & 80.7 & 63.0 & \textcolor{green!60!black}{100.0} \\
        7. ellipsoid & 52.0 & 68.7 & \textcolor{green!60!black}{99.0} & 1.0 & 0.0 & \textcolor{green!60!black}{99.0} & \textcolor{green!60!black}{100.0} & \textcolor{green!60!black}{100.0} \\
        8. europlug & 0.0 & 0.0 & 62.0 & 0.0 & 0.0 & 0.0 & 11.7 & \textcolor{green!60!black}{100.0} \\
        9. gb-2099 & 0.0 & 0.0 & 5.3 & 0.0 & 0.0 & 25.7 & 0.0 & \textcolor{green!60!black}{100.0} \\
        10. hdmi & 0.0 & 0.0 & 0.0 & 0.3 & 0.7 & 0.0 & 0.3 & \textcolor{green!60!black}{100.0} \\
        11. hexkey-12 & 0.0 & 0.0 & 35.0 & 0.3 & 4.3 & 0.0 & 0.7 & \textcolor{green!60!black}{100.0} \\
        12. hexkey-17 & 0.0 & 0.0 & 0.0 & 0.3 & 0.3 & 0.0 & 30.7 & \textcolor{green!60!black}{100.0} \\
        13. hexkey-19 & 0.0 & 0.0 & 0.0 & 0.0 & 0.0 & 0.0 & 0.0 & \textcolor{green!60!black}{100.0} \\
        14. hexkey-6 & 0.0 & 0.0 & 0.0 & 2.3 & 1.7 & 0.0 & 0.0 & \textcolor{green!60!black}{100.0} \\
        15. iec-c13 & 0.0 & 2.3 & 11.3 & 0.0 & 0.3 & 16.3 & \textcolor{blue}{97.7} & \textcolor{green!60!black}{100.0} \\
        16. padlock-burgw\"{a}chter & 0.0 & 0.0 & 0.3 & 0.0 & 0.0 & 1.0 & 23.0 & \textcolor{green!60!black}{100.0} \\
        17. schuko & 0.0 & 0.0 & 27.7 & 0.0 & 0.0 & 0.3 & 33.3 & \textcolor{green!60!black}{100.0} \\
        18. schuko 2 & 0.0 & 0.0 & 6.0 & 0.0 & 0.0 & 3.0 & 0.3 & \textcolor{green!60!black}{100.0} \\
        19. trrs-6.35 mm aluminum rod & 0.0 & 0.0 & 19.0 & 0.0 & 0.0 & 0.0 & 40.7 & \textcolor{green!60!black}{100.0} \\
        20. type-l & 0.0 & 0.0 & 0.0 & 0.0 & 0.0 & 48.3 & 66.0 & \textcolor{green!60!black}{100.0} \\
        21. usb-a & 1.7 & 2.3 & 82.0 & 1.0 & 2.0 & 7.7 & 15.0 & \textcolor{green!60!black}{100.0} \\
        22. usb-c & 0.0 & 0.0 & 1.7 & 2.0 & 13.0 & 0.7 & 1.0 & \textcolor{green!60!black}{100.0} \\
        \midrule
        \textbf{Mean} & \textbf{5.1} & \textbf{7.3} & \textbf{26.3} & \textbf{0.7} & \textbf{1.3} & \textbf{26.2} & \textbf{37.4} & \textbf{100.0} \\
    \\\bottomrule
  \end{tabular}
\end{table}

\begin{table}[htbp]
  \centering
  \caption{Success Rate [\%] -- Soft Actor-Critic}
  \label{tab:success_sac}
  \small
  \setlength{\tabcolsep}{4pt}
  \begin{tabular}{lrrrrrrrr}
    \toprule
         & \multicolumn{3}{c}{Force Domain} & \multicolumn{4}{c}{Velocity Domain} & VI. Tactile \\
        \cmidrule(lr){2-4} \cmidrule(lr){5-8} \cmidrule(lr){9-9}
        Object [\%] & II & III & IV & II & III & IV & V & Skill \\ \midrule
        1. 12point-socket-24 & 0.0 & 0.0 & 0.0 & 0.0 & 0.0 & 4.7 & 27.3 & \textcolor{green!60!black}{100.0} \\
        2. 12point-socket-36 & 19.3 & 35.3 & 6.7 & 0.0 & 0.0 & 66.0 & \textcolor{blue}{93.0} & \textcolor{green!60!black}{100.0} \\
        3. 6point-socket-24 & 0.0 & 0.0 & 36.7 & 7.3 & 1.3 & 8.3 & 55.7 & \textcolor{green!60!black}{100.0} \\
        4. 6point-socket-32 & 1.0 & 0.7 & \textcolor{blue}{96.7} & 0.0 & {--} & 68.3 & \textcolor{blue}{96.0} & \textcolor{green!60!black}{100.0} \\
        5. 6point-socket-41 & 32.7 & 20.0 & 63.7 & 0.0 & 0.3 & 65.7 & 77.3 & \textcolor{green!60!black}{100.0} \\
        6. abus-e30 & 0.0 & 0.0 & 0.0 & 0.0 & 0.0 & 89.7 & \textcolor{blue}{92.3} & \textcolor{green!60!black}{100.0} \\
        7. ellipsoid & 66.3 & 69.7 & \textcolor{green!60!black}{99.3} & 0.0 & 0.0 & \textcolor{green!60!black}{99.7} & \textcolor{green!60!black}{100.0} & \textcolor{green!60!black}{100.0} \\
        8. europlug & 0.0 & 0.0 & 43.0 & 0.0 & 0.0 & 0.0 & 32.0 & \textcolor{green!60!black}{100.0} \\
        9. gb-2099 & 0.0 & 0.0 & 0.0 & 0.0 & 0.0 & 2.7 & 8.3 & \textcolor{green!60!black}{100.0} \\
        10. hdmi & 0.0 & 0.0 & 0.0 & 0.0 & 0.3 & 24.7 & 0.7 & \textcolor{green!60!black}{100.0} \\
        11. hexkey-12 & 0.0 & 0.0 & 9.7 & 1.3 & 1.7 & 0.0 & 5.0 & \textcolor{green!60!black}{100.0} \\
        12. hexkey-17 & 0.0 & 0.0 & 0.0 & 0.3 & 1.0 & 0.3 & 83.3 & \textcolor{green!60!black}{100.0} \\
        13. hexkey-19 & 0.0 & 0.0 & 0.0 & 0.0 & 0.0 & 0.7 & 0.0 & \textcolor{green!60!black}{100.0} \\
        14. hexkey-6 & 0.0 & 0.0 & 0.0 & 0.7 & 1.7 & 0.0 & 0.0 & \textcolor{green!60!black}{100.0} \\
        15. iec-c13 & 0.0 & 0.0 & 3.3 & 0.0 & 0.0 & 20.7 & 53.0 & \textcolor{green!60!black}{100.0} \\
        16. padlock-burgw\"{a}chter & 0.0 & 0.0 & 0.0 & 0.0 & 0.0 & 2.3 & {--} & \textcolor{green!60!black}{100.0} \\
        17. schuko & 0.0 & 0.0 & {--} & 0.0 & 0.0 & 12.7 & 20.7 & \textcolor{green!60!black}{100.0} \\
        18. schuko 2 & 0.0 & 0.0 & 1.0 & 0.0 & 0.0 & 1.3 & 1.7 & \textcolor{green!60!black}{100.0} \\
        19. trrs-6.35 mm aluminum rod & 0.0 & 0.0 & 15.0 & 0.3 & 0.3 & 0.0 & 73.3 & \textcolor{green!60!black}{100.0} \\
        20. type-l & 0.0 & 0.0 & 0.0 & 0.0 & 0.0 & 20.7 & 40.7 & \textcolor{green!60!black}{100.0} \\
        21. usb-a & 0.0 & 0.0 & 64.0 & 0.3 & 0.0 & 0.0 & 23.0 & \textcolor{green!60!black}{100.0} \\
        22. usb-c & 0.0 & 0.3 & {--} & 5.3 & 11.0 & 4.3 & 0.7 & \textcolor{green!60!black}{100.0} \\
        \midrule
        \textbf{Mean} & \textbf{5.4} & \textbf{5.7} & \textbf{20.0} & \textbf{0.7} & \textbf{0.8} & \textbf{22.4} & \textbf{40.2} & \textbf{100.0} \\
    \\\bottomrule
  \end{tabular}
\end{table}
 
\begin{table}[htbp]
  \centering
  \caption{Insertion Speed [Ex.\,s$^{-1}$] -- Proximal Policy Optimization}
  \label{tab:speed_ppo}
  \small
  \setlength{\tabcolsep}{4pt}
  \begin{tabular}{lrrrrrrrr}
    \toprule
         & \multicolumn{3}{c}{Force Domain} & \multicolumn{4}{c}{Velocity Domain} & VI. Tactile \\
        \cmidrule(lr){2-4} \cmidrule(lr){5-8} \cmidrule(lr){9-9}
        Object [Ex.\,s$^{{-1}}$] & II & III & IV & II & III & IV & V & Skill \\ \midrule
        1. 12point-socket-24 & {--} & {--} & 0.89 & {--} & {--} & 0.88 & 0.60 & 1.01 \\
        2. 12point-socket-36 & 0.69 & 0.78 & 1.03 & {--} & {--} & 0.97 & 0.41 & 0.96 \\
        3. 6point-socket-24 & 1.26 & 3.01 & 0.89 & 1.61 & 1.51 & 1.01 & 1.04 & 0.85 \\
        4. 6point-socket-32 & 0.91 & {--} & 2.39 & {--} & 0.91 & 0.98 & 1.22 & 0.98 \\
        5. 6point-socket-41 & 0.42 & 0.35 & 0.88 & {--} & 0.80 & 0.87 & 0.68 & 0.95 \\
        6. abus-e30 & {--} & {--} & {--} & {--} & 0.54 & 0.58 & 0.44 & 0.74 \\
        7. ellipsoid & 0.99 & 0.63 & 1.92 & 0.68 & {--} & 1.25 & 1.13 & 1.12 \\
        8. europlug & {--} & {--} & 1.71 & {--} & {--} & {--} & 0.65 & 0.99 \\
        9. gb-2099 & {--} & {--} & 0.72 & {--} & {--} & 0.49 & {--} & 1.16 \\
        10. hdmi & {--} & {--} & {--} & 1.44 & 0.72 & {--} & 2.16 & 1.23 \\
        11. hexkey-12 & {--} & {--} & 0.59 & 0.47 & 0.52 & {--} & 0.79 & 1.05 \\
        12. hexkey-17 & {--} & {--} & {--} & 0.34 & 0.75 & {--} & 1.61 & 1.26 \\
        13. hexkey-19 & {--} & {--} & {--} & {--} & {--} & {--} & {--} & 0.95 \\
        14. hexkey-6 & {--} & {--} & {--} & 0.82 & 0.88 & {--} & {--} & 1.21 \\
        15. iec-c13 & {--} & 0.53 & 0.71 & {--} & 0.67 & 0.85 & 0.91 & 0.88 \\
        16. padlock-burgw\"{a}chter & {--} & {--} & 2.79 & {--} & {--} & 0.47 & 0.91 & 1.01 \\
        17. schuko & {--} & {--} & 0.66 & {--} & {--} & 0.67 & 0.74 & 0.84 \\
        18. schuko 2 & {--} & {--} & 0.92 & {--} & {--} & 0.48 & 0.56 & 0.77 \\
        19. trrs-6.35 mm aluminum rod & {--} & {--} & 0.64 & {--} & {--} & {--} & 0.94 & 1.06 \\
        20. type-l & {--} & {--} & {--} & {--} & {--} & 0.74 & 0.84 & 1.08 \\
        21. usb-a & 1.09 & 1.14 & 1.08 & 0.73 & 1.23 & 0.70 & 0.56 & 1.12 \\
        22. usb-c & {--} & {--} & 0.34 & 1.73 & 1.67 & 0.45 & 0.61 & 1.25 \\
        \midrule
        \textbf{Mean} & \textbf{0.24} & \textbf{0.29} & \textbf{0.83} & \textbf{0.36} & \textbf{0.46} & \textbf{0.52} & \textbf{0.76} & \textbf{1.02} \\
        \textcolor{cor}{\textit{Adj.\ mean$^\dagger$}} & \textit{0.89} & \textit{1.07} & \textit{1.13} & \textit{0.98} & \textit{0.93} & \textit{0.76} & \textit{0.88} & \textit{1.02} \\
    \\\bottomrule
  \end{tabular}
  \smallskip\par\noindent\footnotesize{$^\dagger$Adjusted mean: computed over tasks with at least one successful trial only; {--} tasks excluded.}
\end{table}
 
\begin{table}[htbp]
  \centering
  \caption{Insertion Speed [Ex.\,s$^{-1}$] -- Soft Actor-Critic}
  \label{tab:speed_sac}
  \small
  \setlength{\tabcolsep}{4pt}
  \begin{tabular}{lrrrrrrrr}
    \toprule
         & \multicolumn{3}{c}{Force Domain} & \multicolumn{4}{c}{Velocity Domain} & VI. Tactile \\
        \cmidrule(lr){2-4} \cmidrule(lr){5-8} \cmidrule(lr){9-9}
        Object [Ex.\,s$^{{-1}}$] & II & III & IV & II & III & IV & V & Skill \\ \midrule
        1. 12point-socket-24 & {--} & {--} & {--} & {--} & {--} & 0.97 & 0.62 & 1.01 \\
        2. 12point-socket-36 & 0.37 & 0.46 & 1.07 & {--} & {--} & 1.19 & 1.00 & 0.96 \\
        3. 6point-socket-24 & {--} & {--} & 0.82 & 1.11 & 1.07 & 1.26 & 0.96 & 0.85 \\
        4. 6point-socket-32 & 0.35 & 0.45 & 1.61 & {--} & {--} & 1.06 & 1.07 & 0.98 \\
        5. 6point-socket-41 & 0.38 & 0.33 & 0.58 & {--} & 0.61 & 0.93 & 0.63 & 0.95 \\
        6. abus-e30 & {--} & {--} & {--} & {--} & {--} & 0.72 & 0.54 & 0.74 \\
        7. ellipsoid & 0.53 & 0.44 & 1.85 & {--} & {--} & 1.27 & 1.43 & 1.12 \\
        8. europlug & {--} & {--} & 1.46 & {--} & {--} & {--} & 0.67 & 0.99 \\
        9. gb-2099 & {--} & {--} & {--} & {--} & {--} & 0.59 & 0.48 & 1.16 \\
        10. hdmi & {--} & {--} & {--} & {--} & 1.04 & 0.90 & 0.66 & 1.23 \\
        11. hexkey-12 & {--} & {--} & 0.50 & 0.46 & 0.81 & {--} & 0.39 & 1.05 \\
        12. hexkey-17 & {--} & {--} & {--} & 0.38 & 1.07 & 1.36 & 1.26 & 1.26 \\
        13. hexkey-19 & {--} & {--} & {--} & {--} & {--} & 1.06 & {--} & 0.95 \\
        14. hexkey-6 & {--} & {--} & {--} & 0.49 & 0.60 & {--} & {--} & 1.21 \\
        15. iec-c13 & {--} & {--} & 0.55 & {--} & {--} & 0.88 & 0.78 & 0.88 \\
        16. padlock-burgw\"{a}chter & {--} & {--} & {--} & {--} & {--} & 0.47 & {--} & 1.01 \\
        17. schuko & {--} & {--} & {--} & {--} & {--} & 1.06 & 1.03 & 0.84 \\
        18. schuko 2 & {--} & {--} & 0.80 & {--} & {--} & 1.00 & 0.67 & 0.77 \\
        19. trrs-6.35 mm aluminum rod & {--} & {--} & 0.58 & 0.44 & 0.82 & {--} & 0.76 & 1.06 \\
        20. type-l & {--} & {--} & {--} & {--} & {--} & 0.87 & 0.61 & 1.08 \\
        21. usb-a & {--} & {--} & 1.09 & 1.22 & {--} & {--} & 1.16 & 1.12 \\
        22. usb-c & {--} & 1.68 & {--} & 1.10 & 1.47 & 0.69 & 0.53 & 1.25 \\
        \midrule
        \textbf{Mean} & \textbf{0.07} & \textbf{0.15} & \textbf{0.50} & \textbf{0.24} & \textbf{0.34} & \textbf{0.74} & \textbf{0.69} & \textbf{1.02} \\
        \textcolor{cor}{\textit{Adj.\ mean$^\dagger$}} & \textit{0.41} & \textit{0.67} & \textit{0.99} & \textit{0.74} & \textit{0.94} & \textit{0.96} & \textit{0.80} & \textit{1.02} \\
    \\\bottomrule
  \end{tabular}
  \smallskip\par\noindent\footnotesize{$^\dagger$Adjusted mean: computed over tasks with at least one successful trial only; {--} tasks excluded.}
\end{table}

\begin{figure}[!t]
    \centering
    \includegraphics[width=1.0\textwidth]{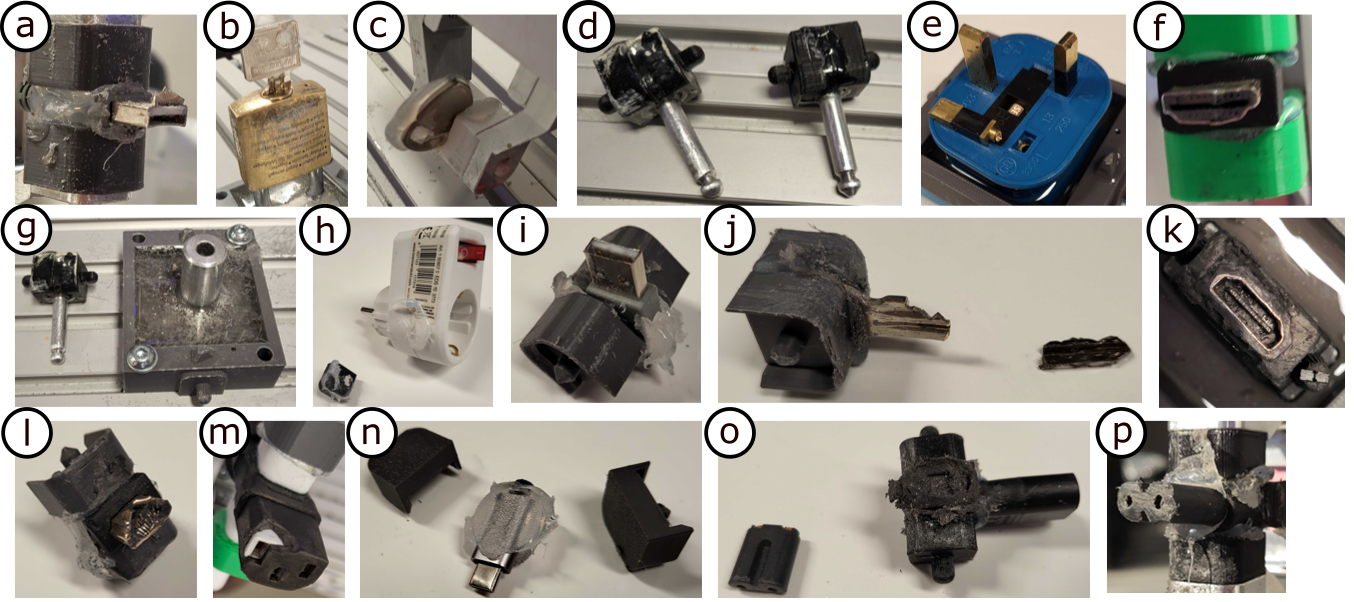}
\caption{%
  \textbf{Representative objects 
  damaged or destroyed during 
  end-to-end deep\,RL training, 
  illustrating the physical 
  consequences of unconstrained 
  policies on real hardware.}
  Observed damage types:
 \textbf{(a)}~extensive destruction 
  of USB-A connector tip;
  \textbf{(b)}~key detached from 
  fixture and jammed during 
  learning;
  \textbf{(c)}~key displaced 
  laterally relative to gripper, 
  original goal transformation 
  invalidated;
  \textbf{(d)}~phone jack connector 
  severely scratched, physical 
  interaction effects altered;
  \textbf{(e)}~connector pin 
  broken;
  \textbf{(f)}~destruction of 
  HDMI insertion tip;
  \textbf{(g)}~severe damage to 
  socket fixture and surrounding 
  environment;
  \textbf{(h)}~gripper mount 
  fractured;
  \textbf{(i)}~USB-A connector 
  displaced axially relative to 
  gripper, original goal 
  transformation invalidated;
  \textbf{(j)}~key bit fractured 
  due to substantial wear;
  \textbf{(k)}~marked damage to 
  internal contacts of HDMI 
  socket;
  \textbf{(l)}~complete destruction 
  of HDMI plug;
  \textbf{(m)}~outer housing of 
  IEC\,C13 connector destroyed;
  \textbf{(n)}~fixture mount 
  fractured;
  \textbf{(o)}~connector insertion 
  tip broken off;
  \textbf{(p)}~plastic connector 
  tip abraded through wear, 
  physical interaction effects 
  significantly altered.
  Several damage types directly 
  alter the physical interaction 
  effects of the task — worn or 
  abraded insertion 
  tips~(a, d, f, p) and displaced 
  objects~(c, i) modify the 
  effective clearance, friction, 
  and contact geometry, 
  systematically biasing subsequent 
  learning trials.
  Destruction rates peak in the 
  force domain at 
  Level~\textsc{IV}, where scaled 
  force commands generate 
  high-energy contact events during 
  failed insertions 
  (Fig.~\ref{fig:destroyed_statistics}).
  The Tactile Skill framework 
  produces no comparable damage, 
  as the impedance-based control 
  structure limits contact forces 
  regardless of the learned 
  parameters.%
}
    \label{fig:destroyed_objects}
\end{figure}

\begin{table}[h]
\centering
\caption{Hyperparameter configuration for PPO and SAC.}
\label{tab:net_arch}
\small
\begin{tabular}{lll}
\toprule
\textbf{Component} & \textbf{PPO} & \textbf{SAC} \\
\midrule
Optimizer                        & Adam          & Adam~\cite{kingma2015adam}  \\
Number of hidden layers          & 3             & 3 \\
Hidden dimension                 & 256           & 256 \\
Activation function              & \texttt{tanh} & \texttt{ReLU} \\
Output activation function       & None          & None \\
Learning rate (Actor and Critic) & $3\times10^{-4}$ & $3\times10^{-4}$ \\
Entropy coefficient / $\alpha$   & $10^{-3}$ (fixed) & $\alpha_\text{init}=0.2$ (trainable) \\
Discount factor $\gamma$         & 0.99          & 0.99 \\
GAE $\lambda$                    & 0.95          & --- \\
Replay buffer size               & 512 (on-policy) & $10^6$ (off-policy) \\
Batch size                       & 256           & 256 \\
Initial exploration w/o training & 1{,}000       & 1{,}000 \\
\bottomrule
\end{tabular}
\end{table}

\setlength{\tabcolsep}{3pt}
\begin{table}[h]
\centering
\caption{Energy consumption per 100 trials per robot, broken down
by system component.}
\label{tab:energy}
\small
\begin{tabular}{lcccc}
\toprule
\textbf{Source} & \textbf{Robot+Master Cont.} & \textbf{Local PC} 
& \textbf{Training PC} & \textbf{Inference} \\
\midrule
Energy \textcolor{cor}{[kJ]} & 42.0 & 38.5 & 6.1 & 0.5 \\
\bottomrule
\end{tabular}
\end{table}

\begin{table}[h]
\centering
\caption{Breakdown of experimental time expenditure by campaign.
Preparation tasks are largely non-recurring per campaign; core experiment 
tasks persisted throughout each campaign's runtime.}
\label{tab:timetable}
\small
\setlength{\tabcolsep}{2pt}
\begin{tabular}{lr|lr}
\hline
\textit{Preparation tasks} & \hspace{-12pt}\textit{Time} \textcolor{cor}{[h]} & 
\textit{Core experiment tasks} & \hspace{-12pt}\textit{Time} \textcolor{cor}{[h]} \\
\hline
\multicolumn{4}{l}{\textbf{Exp. 1: Structured Skill Learning} (22 robots, 22 objects, ${\sim}35{,}200$ scheduled trials, 3 days)}
\\
Experiment design                       & 10  & Daily robot restarts ($n=22$)      & ${\sim}2$ \\
Software \& network comm.\ design       & 600 & Hardware safety \& upkeep          & ${\sim}50$ \\
Algorithm implementation                & 2   & Learning trials                    & ${\sim}80$ \\
Object \& fixture preparation           & 150 & Anomaly logging \& trial invalidation & 30 \\
Start \& goal pose teaching             & 5   & Data logging \& storage mgmt.      & 10 \\
Exploratory runs                        & --  & & \\
\textbf{Subtotal}                       & ${\sim}767$ & \textbf{Subtotal}         & ${\sim}172$ \\
\hline
\multicolumn{4}{l}{\textbf{Exp. 2: Deep Learning} (22 robots, 22 objects, 693{,}000 trials, 68 days)}
\\
Experiment design                       & 50  & Daily robot restarts ($n=22$)      & ${\sim}50$ \\
Software \& network comm.\ design       & 300 & Hardware safety \& upkeep          & ${\sim}200$ \\
Algorithm implementation                & 50  & Learning trials                    & ${\sim}15{,}000$ \\
Object \& fixture preparation           & 150 & Anomaly logging \& trial invalidation & ${\sim}200$ \\
Start \& goal pose teaching             & 5   & Data logging \& storage mgmt.      & ${\sim}20$ \\
Model knowledge tests \& expl.\ runs  & 200 & & \\
\textbf{Subtotal}                       & ${\sim}755$ & \textbf{Subtotal}         & ${\sim}15{,}470$ \\
\hline
\multicolumn{4}{l}{\textbf{Exp. 3: Clearance and Geometry} (2 robots, 12 objects, 18{,}000 trials, 10 days)}
\\
Experiment design                       & 30  & Daily robot restarts ($n=2$)       & 1 \\
Software \& network comm.\ design       & 2   & Hardware safety \& upkeep          & 5 \\
Algorithm implementation                & 2   & Learning trials                    & ${\sim}80$ \\
Object \& fixture preparation           & 12  & Anomaly logging \& trial invalidation & 1 \\
Start \& goal pose teaching             & 1   & Data logging \& storage mgmt.      & 1 \\
Exploratory runs                        & 3   & & \\
\textbf{Subtotal}                       & ${\sim}50$ & \textbf{Subtotal}          & ${\sim}88$ \\
\hline
\textbf{Total}                          & ${\sim}1{,}572$ & \textbf{Total}        & ${\sim}15{,}730$ \\
\hline
\end{tabular}
\end{table}

\begin{table}[h]
\centering
\caption{One-time infrastructure setup 
time for the robot collective.}
\label{tab:infrastructure}
\small
\setlength{\tabcolsep}{4pt}
\begin{tabular}{lr|lr}
\hline
\textit{Software} & 
\textit{Time} \textcolor{cor}{[h]} & 
\textit{Hardware} & 
\textit{Time} \textcolor{cor}{[h]}  \\
\hline
Kubernetes orchestration  & 30  & 
  Robot tables \& levelling   & 310 \\
ML service                & 830 & 
  Cabling                     & 30  \\
MiOS                      & 350 & 
  Object production           & 150 \\
Linux (network, scripts,  &     & 
  Object mounting \&          &     \\
\quad setup)              & 150 & 
  \quad pose teaching         & 10  \\
Robot software setup      & 30  &   & \\
\hline
\textbf{Subtotal}         & ${\sim}1{,}390$ & 
\textbf{Subtotal}         & ${\sim}500$ \\
\hline
\multicolumn{2}{l}{\textbf{Total}} & 
\multicolumn{2}{r}{${\sim}1{,}890$\,h} \\
\hline
\end{tabular}
\end{table}

\begin{figure}[!t]
    \centering
    \includegraphics[width=1.0\textwidth]{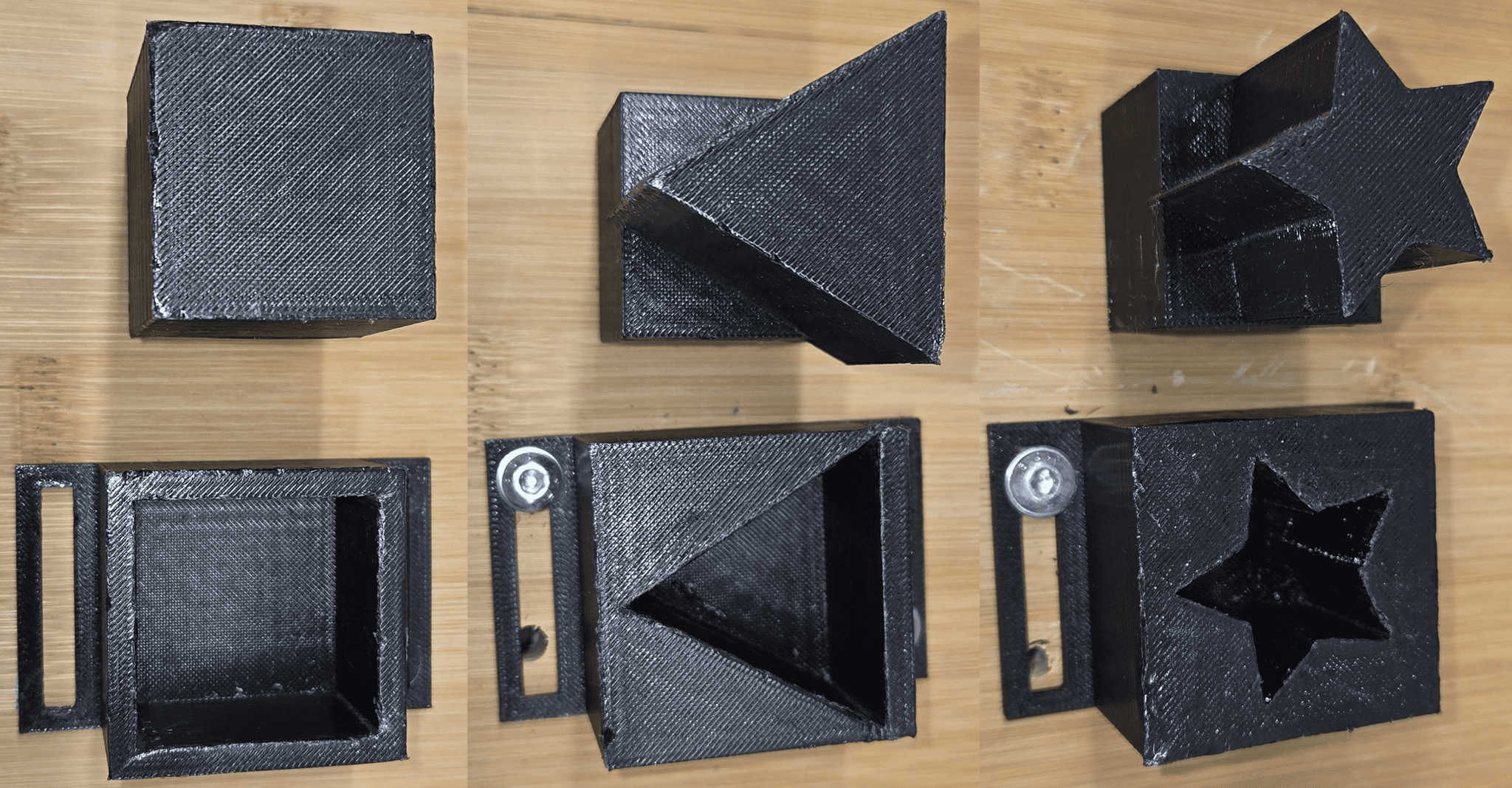}
    \caption{%
  \textbf{Purpose-manufactured peg-and-hole test parts used in the
  clearance and geometry experiment.}
  Three cross-sectional shapes — triangular, square, and star — are shown
  at four clearance levels (0.0, 0.5, 1.0, and 2.0\,mm), together with
  their corresponding holes and mounting fixtures.
  Parts were 3D-printed from PLA to provide controlled, repeatable material conditions.%
}
    \label{fig:clearance_shapes}
\end{figure}

\begin{table}[ht]
\centering
\caption{Industrial assembly tasks validated with German industry partners.
MTM cycle times are normal times (100\% performance, no allowances);
see Supplementary Data file for full
motion-level breakdowns.}
\label{tab:industrial_assemblies}
\begin{tabular}{lllll}
\toprule
 & \textcolor{cor}{\textbf{Case 1}} & \textcolor{cor}{\textbf{Case 2}} & \textcolor{cor}{\textbf{Case 3}} & \textcolor{cor}{\textbf{Case 4}} \\
\midrule
Unique subtasks         & 13                            & 9               & 5                     & 2 \\
Total subtasks          & 17                            & 9              & 23                     & 2 \\
Tolerances (mm)         & ${<}0.05$                     & ${<}0.25$        & ${<}0.2$             & ${<}0.5$ \\
Programmed poses        & 172 (${\sim}8\,\text{h}$)     & 29 (${\sim}1.5\,\text{h}$) & 18 (${\sim}1\,\text{h}$) & 6 (${\sim}0.25\,\text{h}$) \\
Learning time           & ${\sim}3\,\text{h}$           & ${\sim}1.5\,\text{h}$  & ${\sim}1\,\text{h}$   & ${\sim}0.33\,\text{h}$ \\
Robot cycle time        & ${\sim}4\,\text{min}\,20\,\text{s}$ & ${\sim}1\,\text{min}\,15\,\text{s}$ & ${\sim}5\,\text{min}\,50\,\text{s}$ & ${\sim}15\,\text{s}$ \\
Methods-time measurement& ${\sim}1\,\text{min}\,3\,\text{s}$  & ${\sim}21.5\,\text{s}$ & ${\sim}1\,\text{min}\,25\,\text{s}$  & ${\sim}4.8\,\text{s}$ \\
Human cycle time (measured)       & ${\sim}1\,\text{min}\,5\,\text{s}$  & ${\sim}20\,\text{s}$ & ${\sim}1\,\text{min}\,20\,\text{s}$    & ${\sim}5\,\text{s}$ \\
Reliability             & 100\%                         & ${\geq}98\%$           & 100\%          & 100\% \\
Repetitions             & 50                            & 50              & 50                    & 100$^\dagger$ \\
\bottomrule
\multicolumn{5}{l}{\footnotesize $^\dagger$ \textcolor{cor}{Case 4}: 100 repetitions conducted to assess reliability of the tactile} \\
\multicolumn{5}{l}{\footnotesize quality inspection procedure.} \\
\end{tabular}%
\end{table}

\clearpage
\section*{Task Profile Datasheets}\label{sec:task_profiles_supple}

The following pages present the 
automatically generated task profile 
for each of the 22 benchmark tasks, 
based on the best parameter vector 
$\boldsymbol{\theta}^*$ found during 
the Tactile Skill learning run. 
Each profile comprises three 
components.

\textit{Physical effect profile} (left): horizontal bars show the
normalised magnitudes of P1--P5, assigned by expert consensus,
providing a compact fingerprint of the task's physical interaction
regime.

\textit{Trajectory visualisation} 
(centre): end-effector trajectories 
projected onto the $(x,y)$, $(x,z)$, 
and $(y,z)$ planes for each of the 
three skill phases — approach, 
contact, and insertion. The full 
trajectory is 
shown transparently; the part 
of the active phase is shown in 
full colour. Gripper orientation 
is indicated in the upper right 
corner of each panel. Contact force 
vectors are overlaid as red arrows. 
Parameters governing the trajectory 
are colour-coded by functional role: 
red parameters are most important 
for insertion success; orange 
parameters are most important for 
execution speed, as determined by 
the permutation-based importance 
analysis.

\textit{Parameter summary and 
observations} (right): the 
normalised values of 
$\boldsymbol{\theta}^*$ organised 
by phase, together with the 
achieved success rate and mean 
execution time of the best 
solution. Natural-language 
observations summarising the 
most important parameters for 
success and execution speed, 
derived from the permutation-based 
importance analysis, are provided 
at the bottom of the panel.

\begin{figure}[!t]
    \centering
    \includegraphics[
        width=1.0\textheight,
        angle=90
    ]{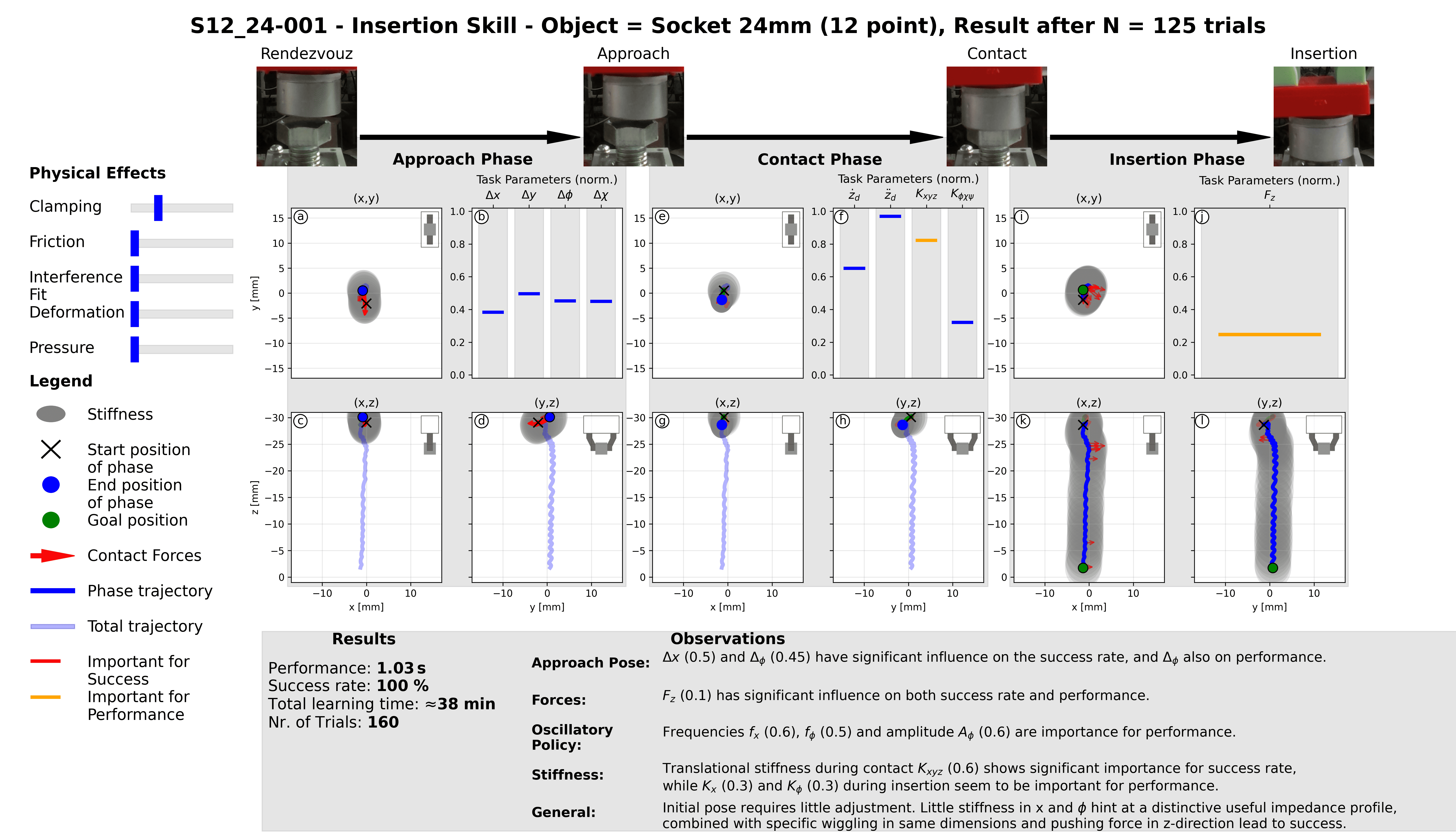}
\end{figure}

\begin{figure}[!t]
    \centering
    \includegraphics[
        width=1.0\textheight,
        angle=90
    ]{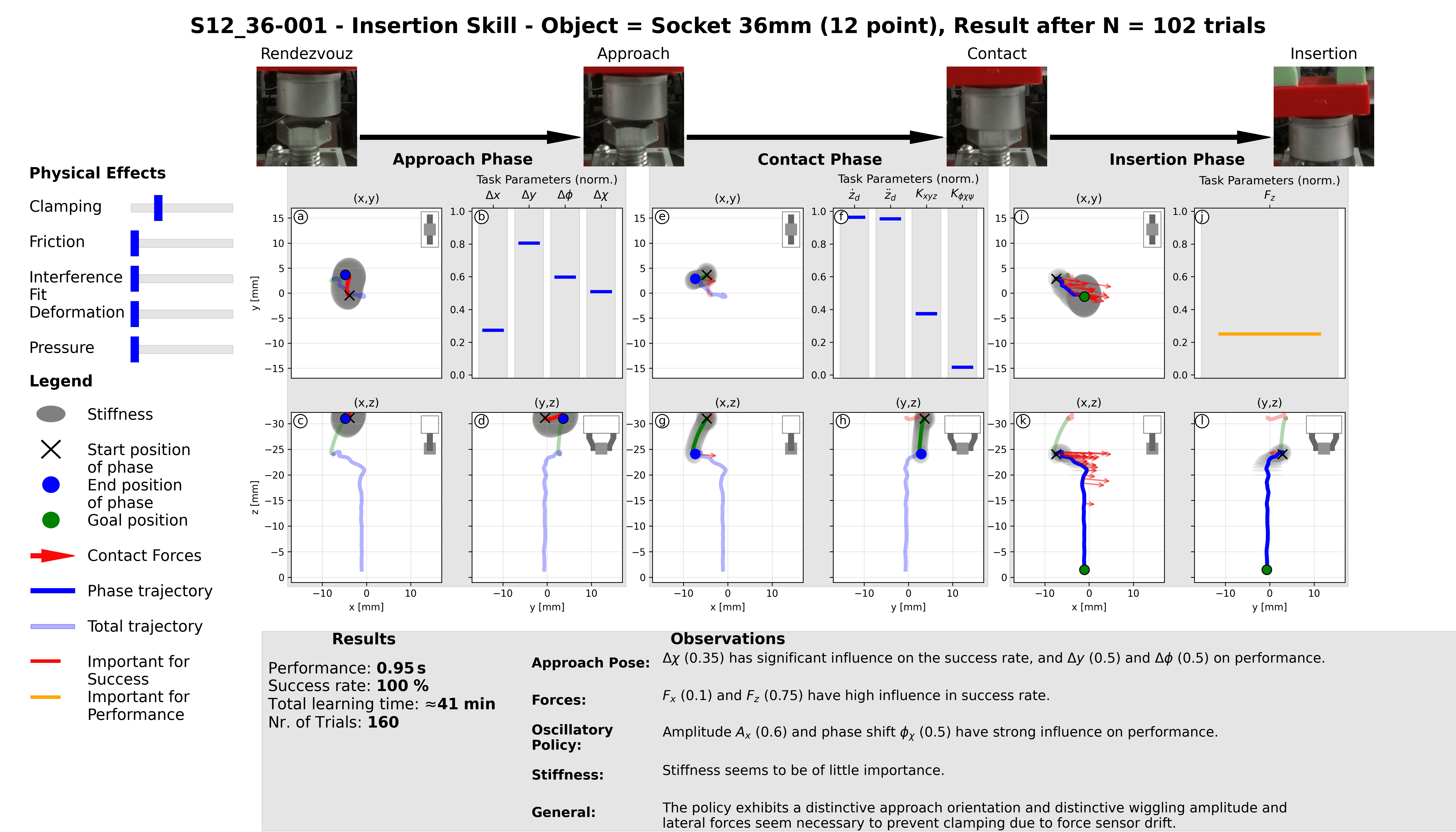}
\end{figure}

\begin{figure}[!t]
    \centering
    \includegraphics[
        width=1.0\textheight,
        angle=90
    ]{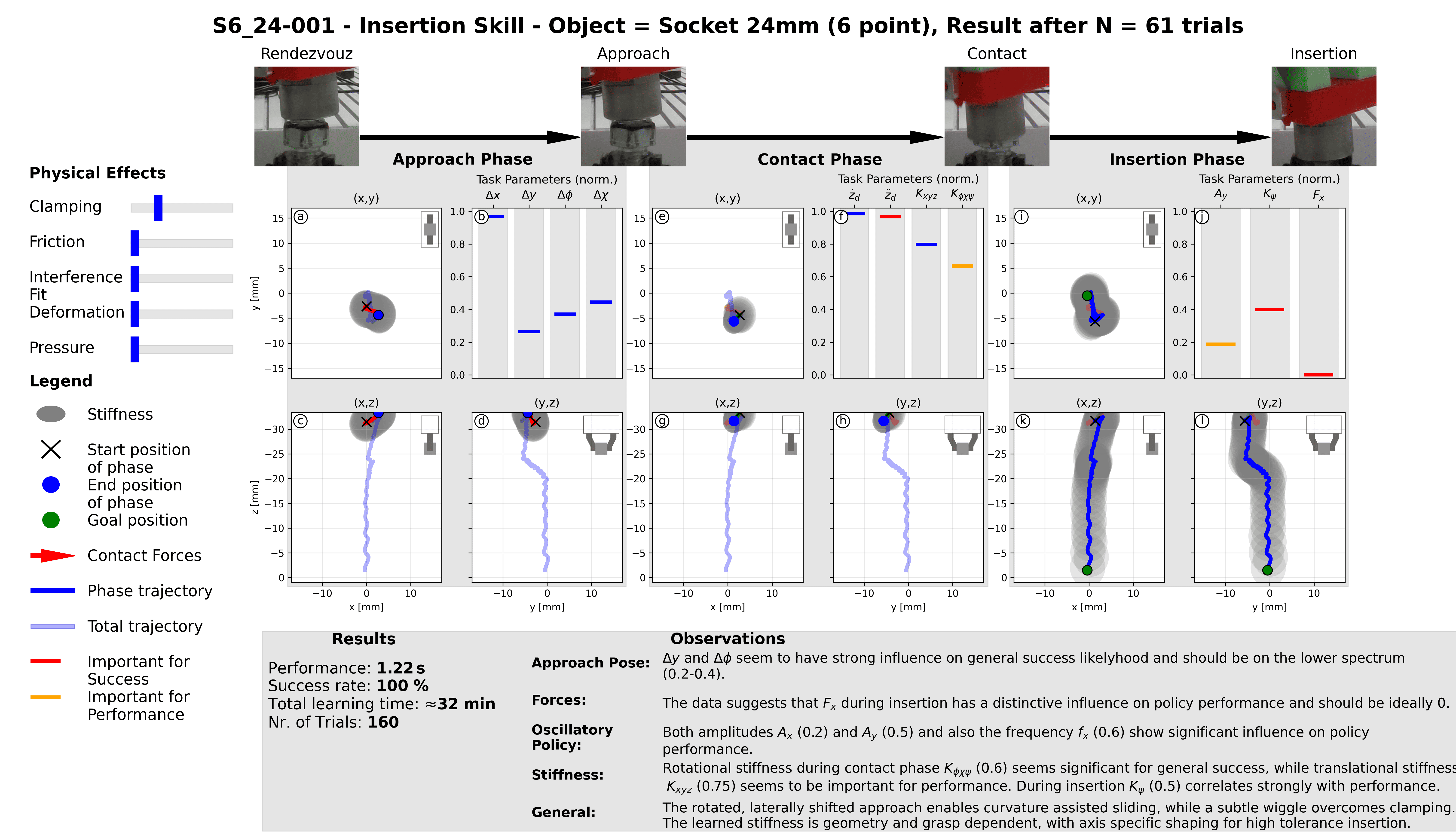}
\end{figure}

\begin{figure}[!t]
    \centering
    \includegraphics[
        width=1.0\textheight,
        angle=90
    ]{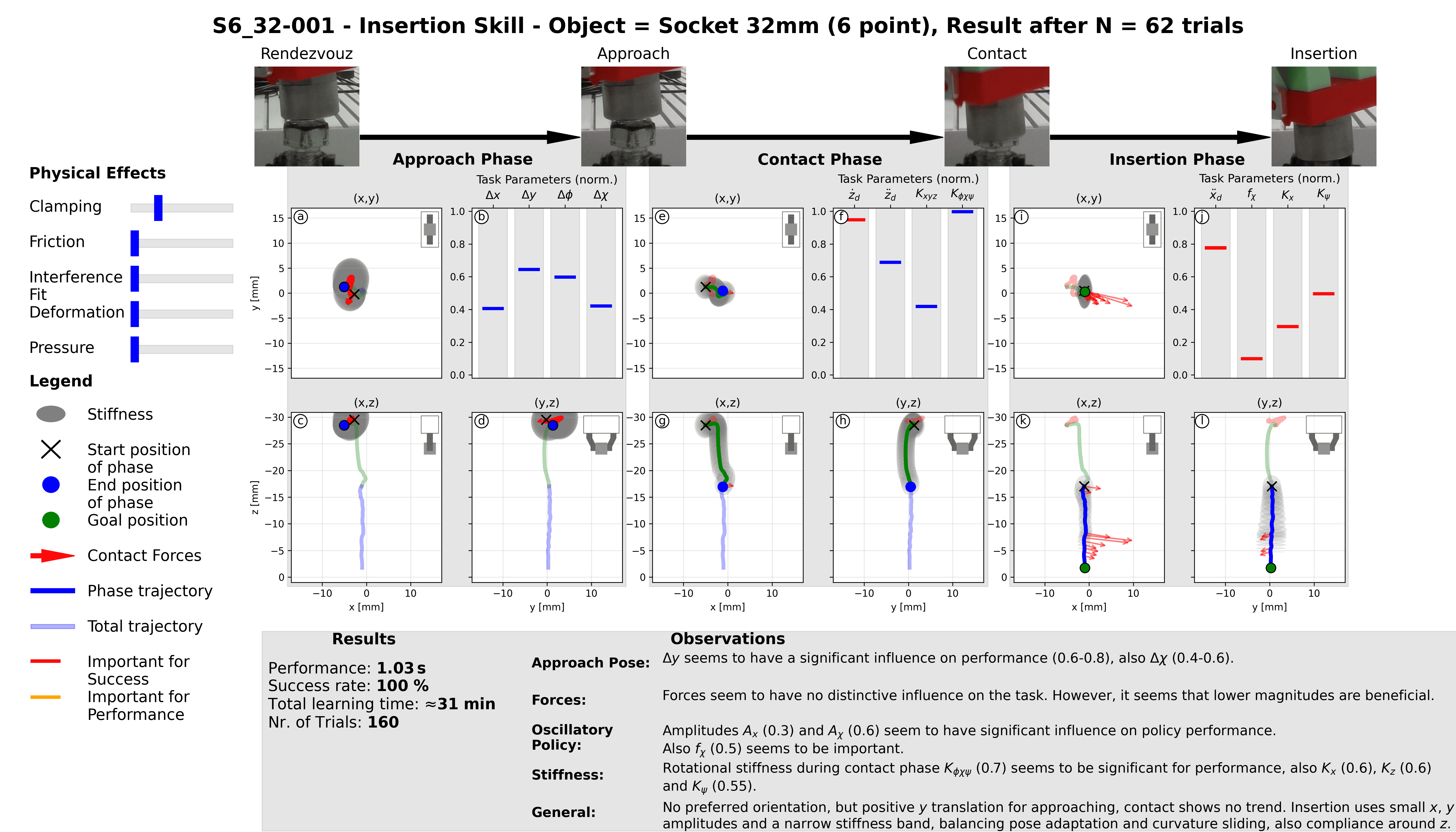}
\end{figure}

\begin{figure}[!t]
    \centering
    \includegraphics[
        width=1.0\textheight,
        angle=90
    ]{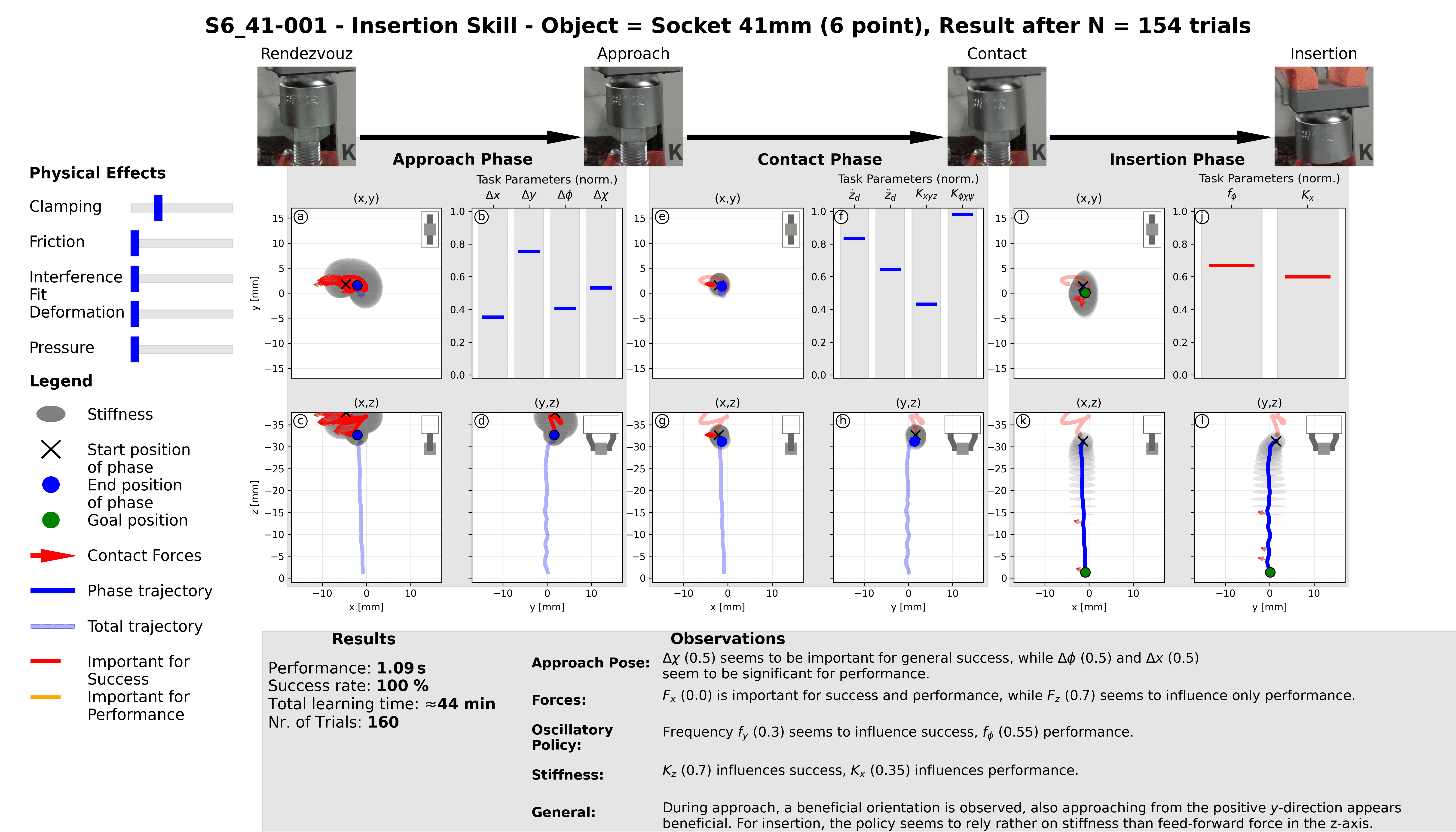}
\end{figure}

\begin{figure}[!t]
    \centering
    \includegraphics[
        width=1.0\textheight,
        angle=90
    ]{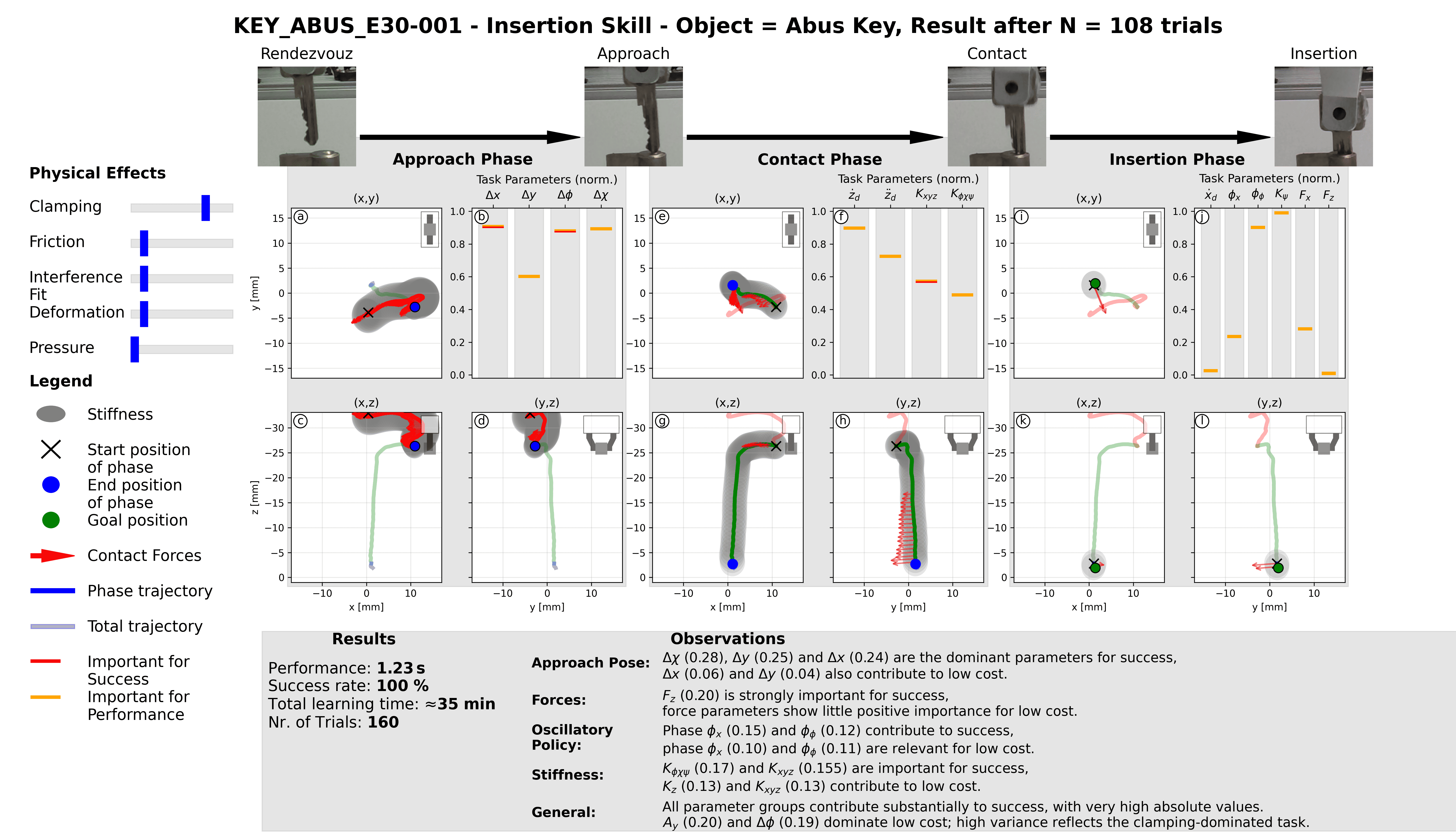}
\end{figure}

\begin{figure}[!t]
    \centering
    \includegraphics[
        width=1.0\textheight,
        angle=90
    ]{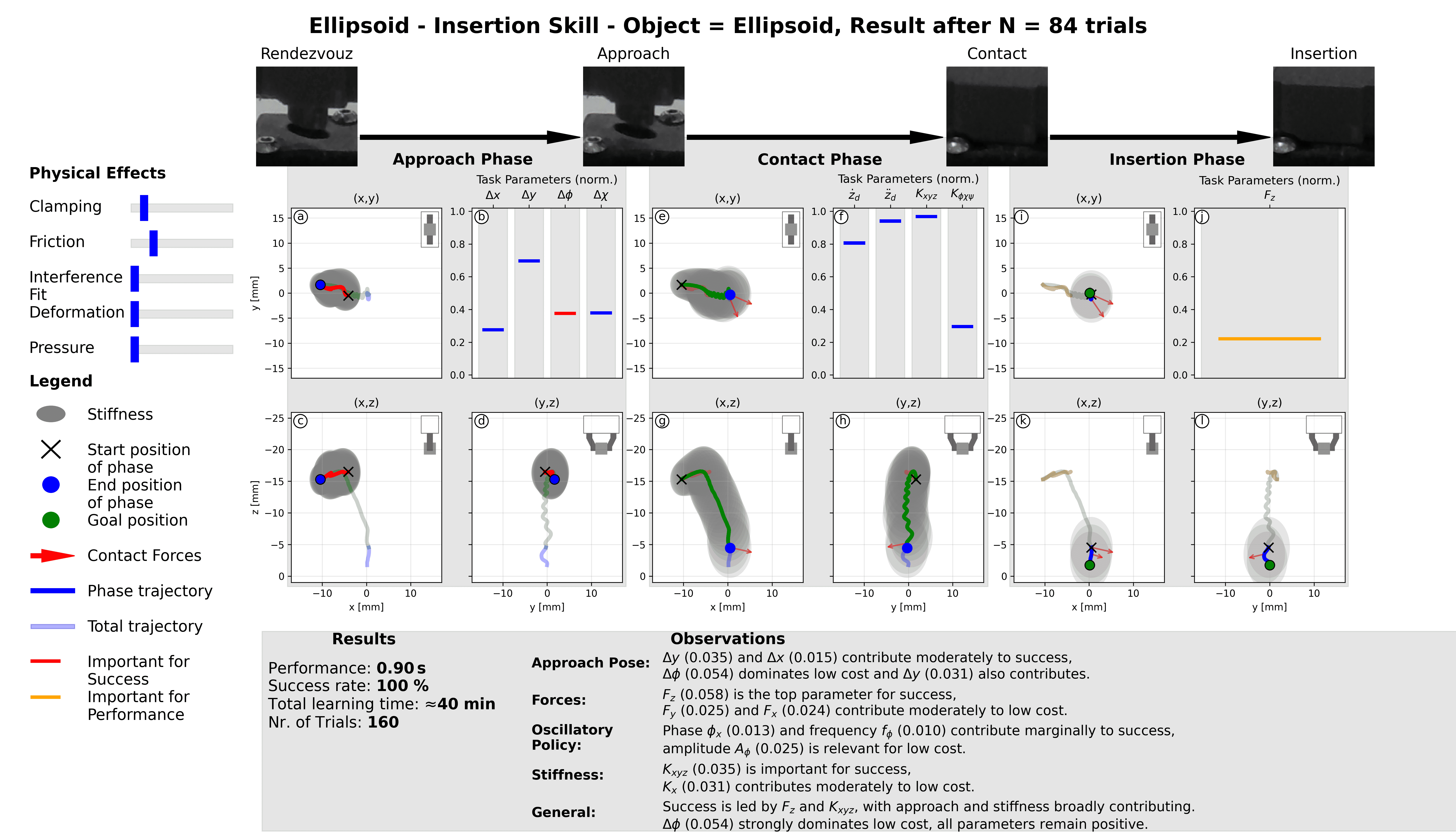}
\end{figure}

\begin{figure}[!t]
    \centering
    \includegraphics[
        width=1.0\textheight,
        angle=90
    ]{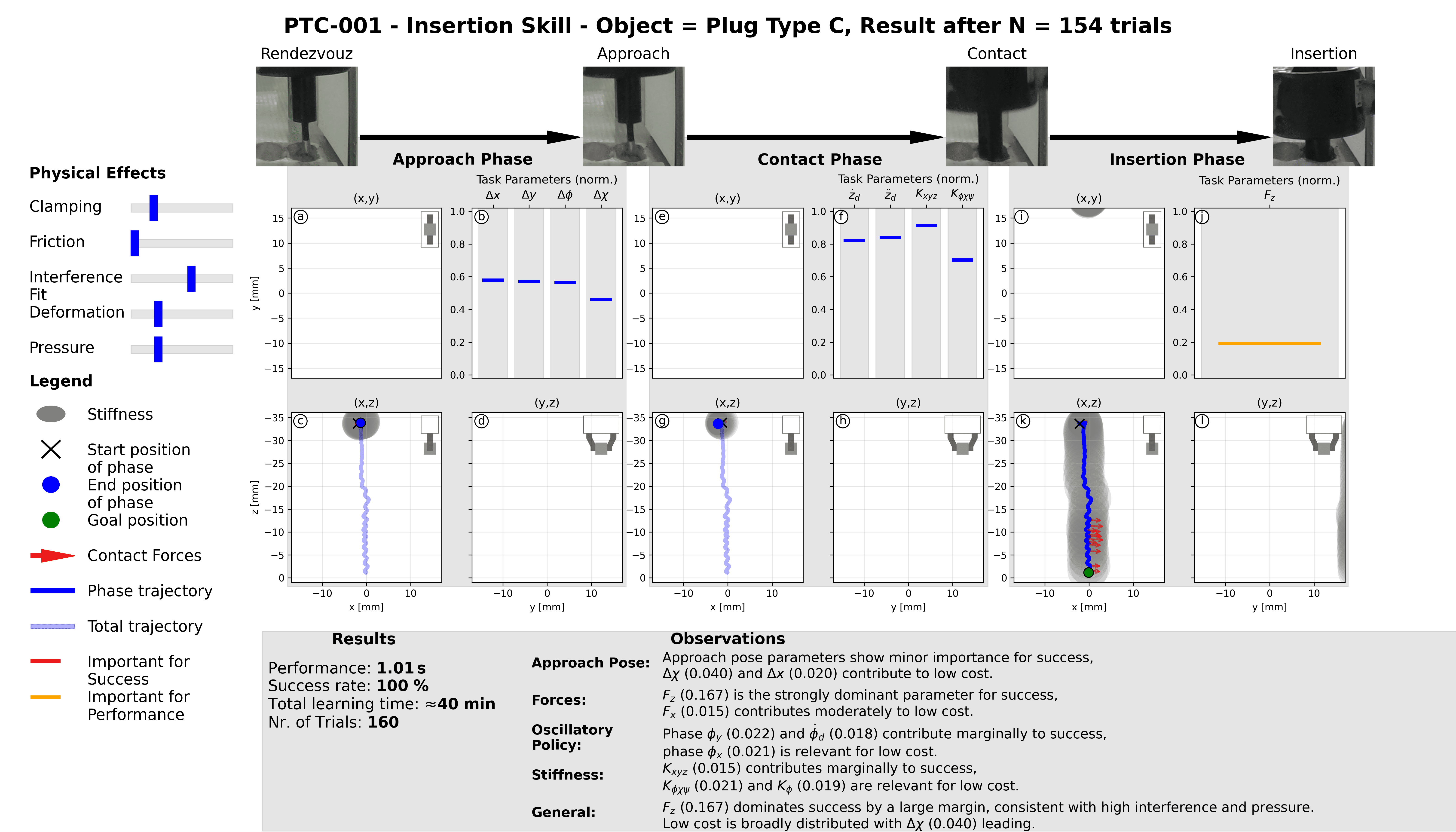}
\end{figure}

\begin{figure}[!t]
    \centering
    \includegraphics[
        width=1.0\textheight,
        angle=90
    ]{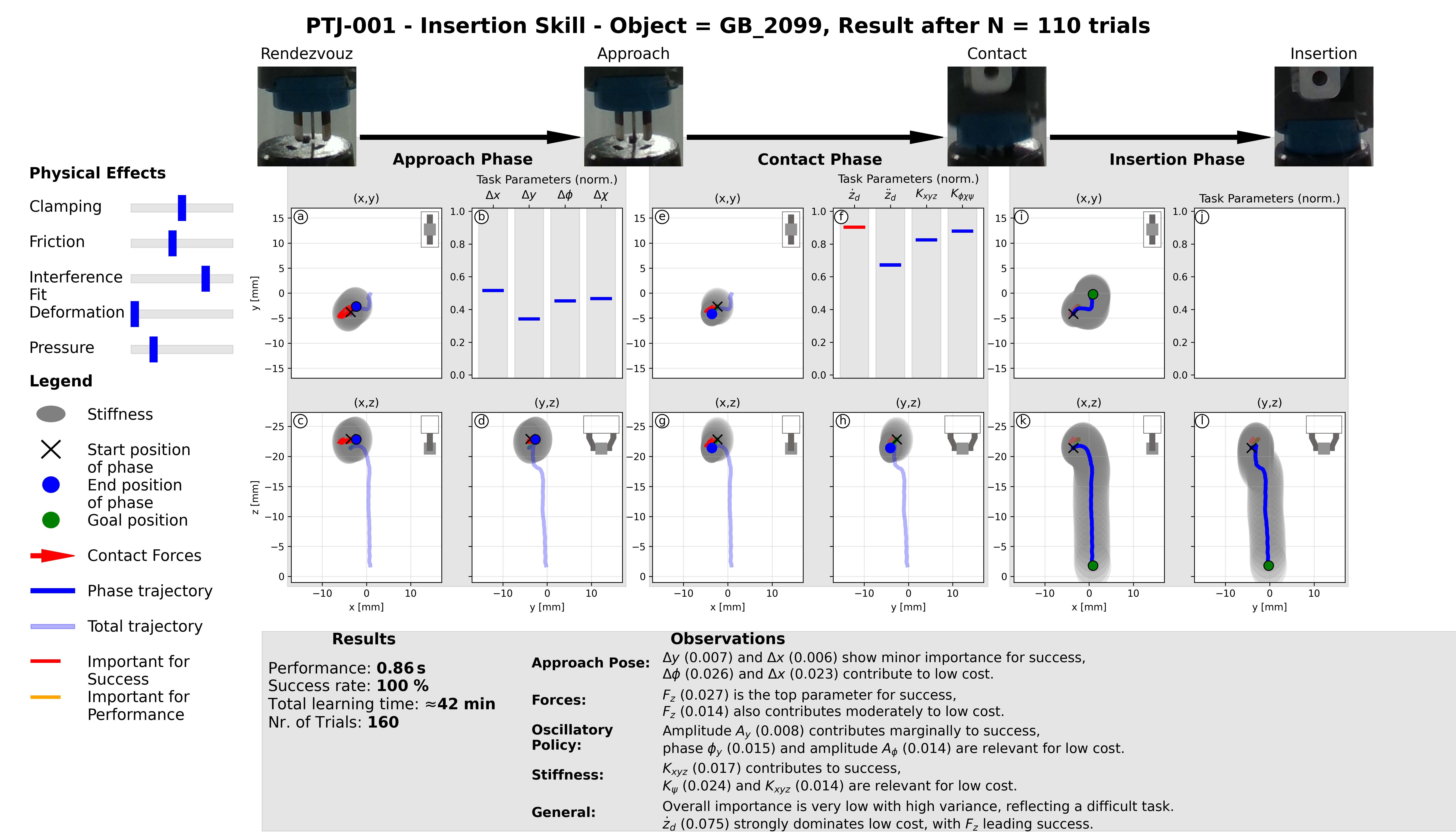}
\end{figure}

\begin{figure}[!t]
    \centering
    \includegraphics[
        width=1.0\textheight,
        angle=90
    ]{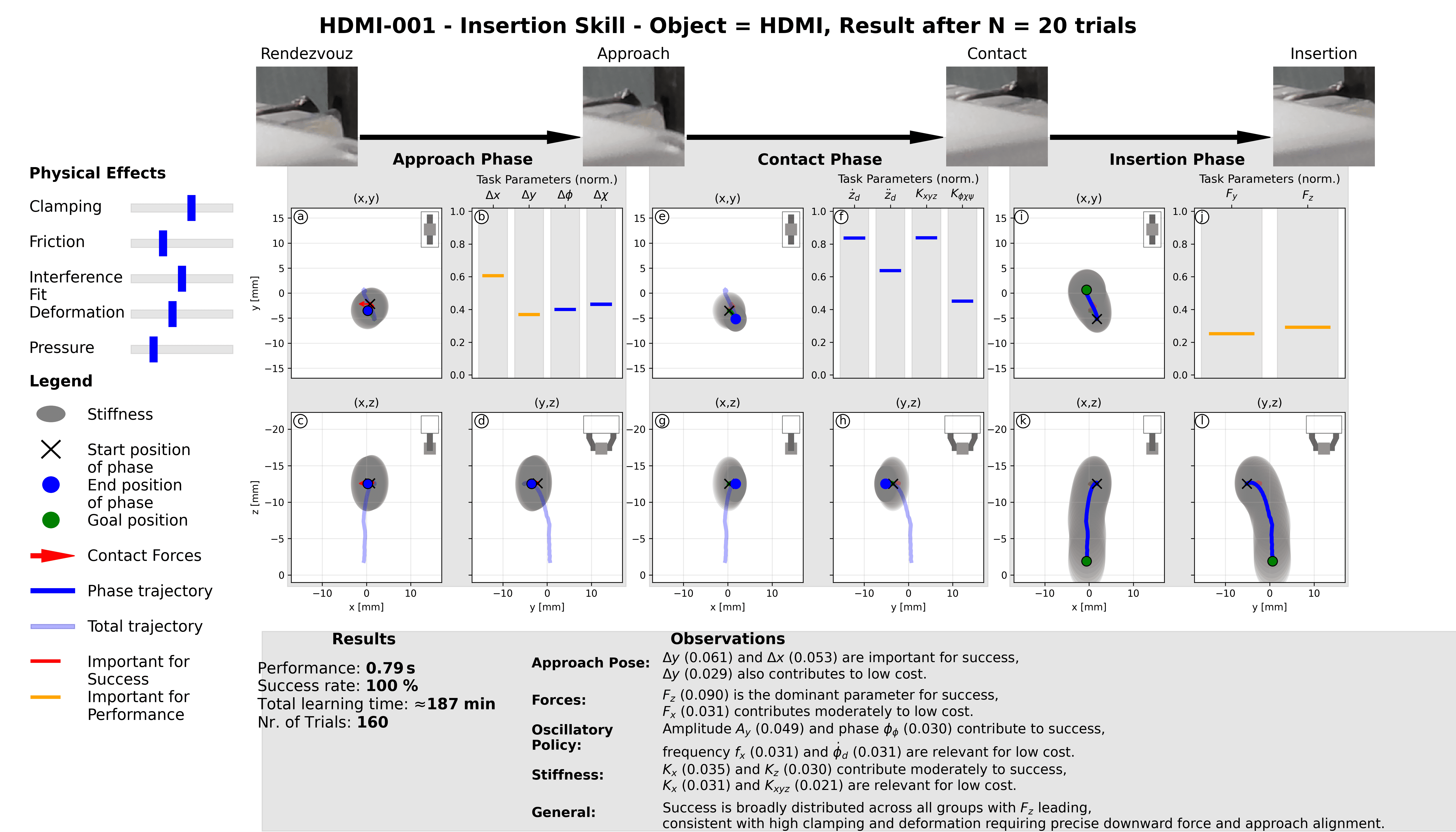}
\end{figure}

\begin{figure}[!t]
    \centering
    \includegraphics[
        width=1.0\textheight,
        angle=90
    ]{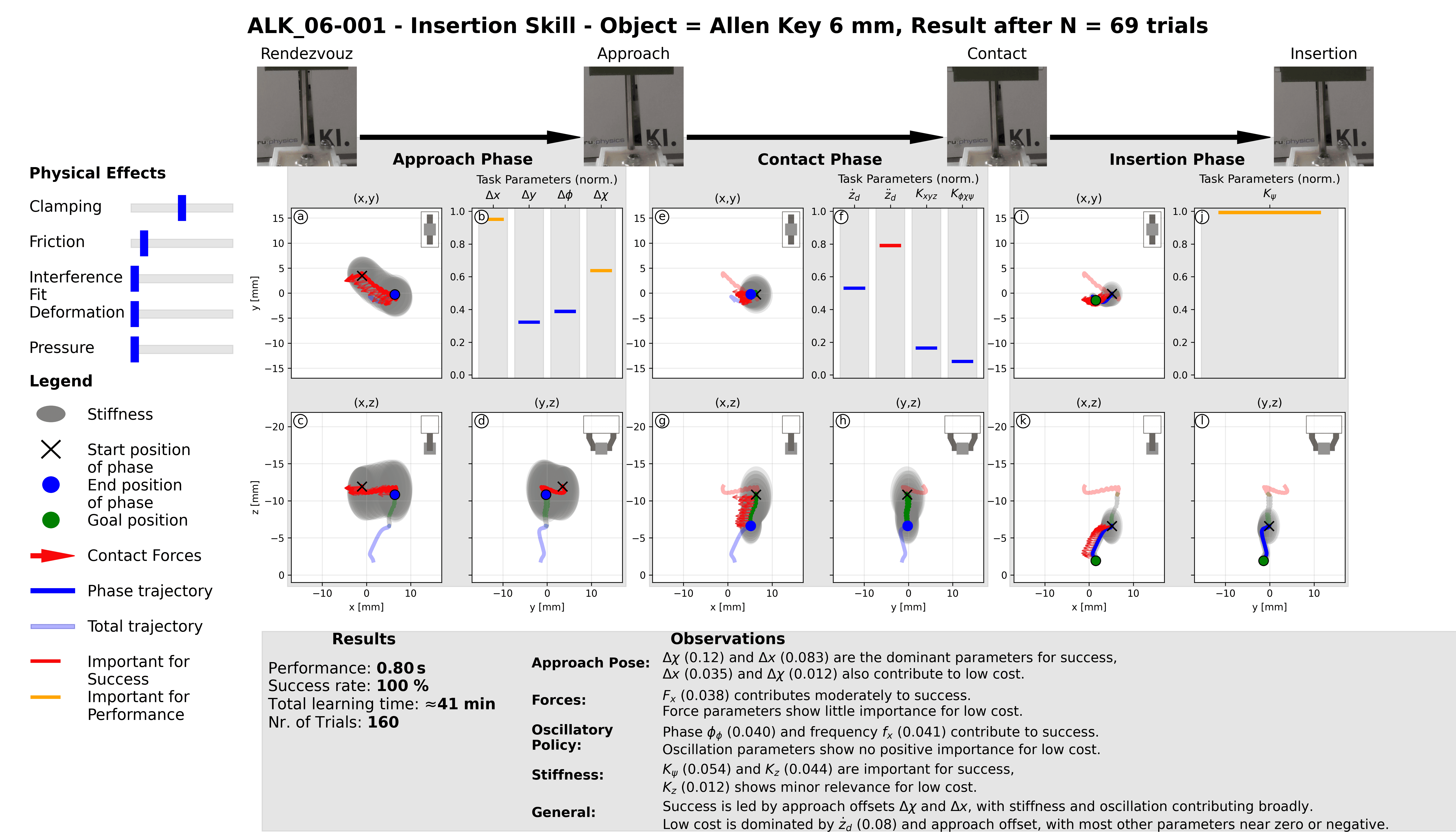}
\end{figure}

\begin{figure}[!t]
    \centering
    \includegraphics[
        width=1.0\textheight,
        angle=90
    ]{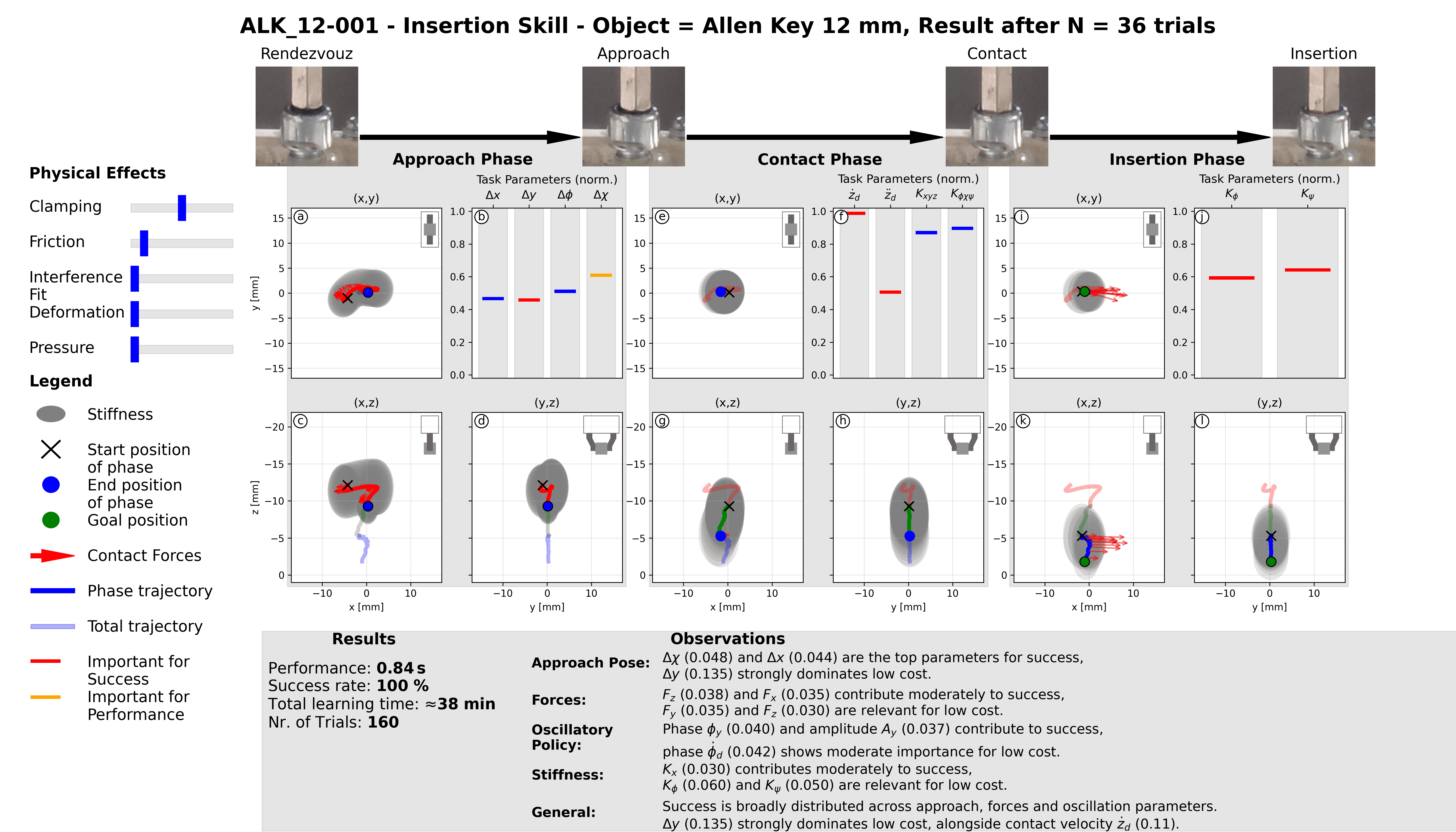}
\end{figure}

\begin{figure}[!t]
    \centering
    \includegraphics[
        width=1.0\textheight,
        angle=90
    ]{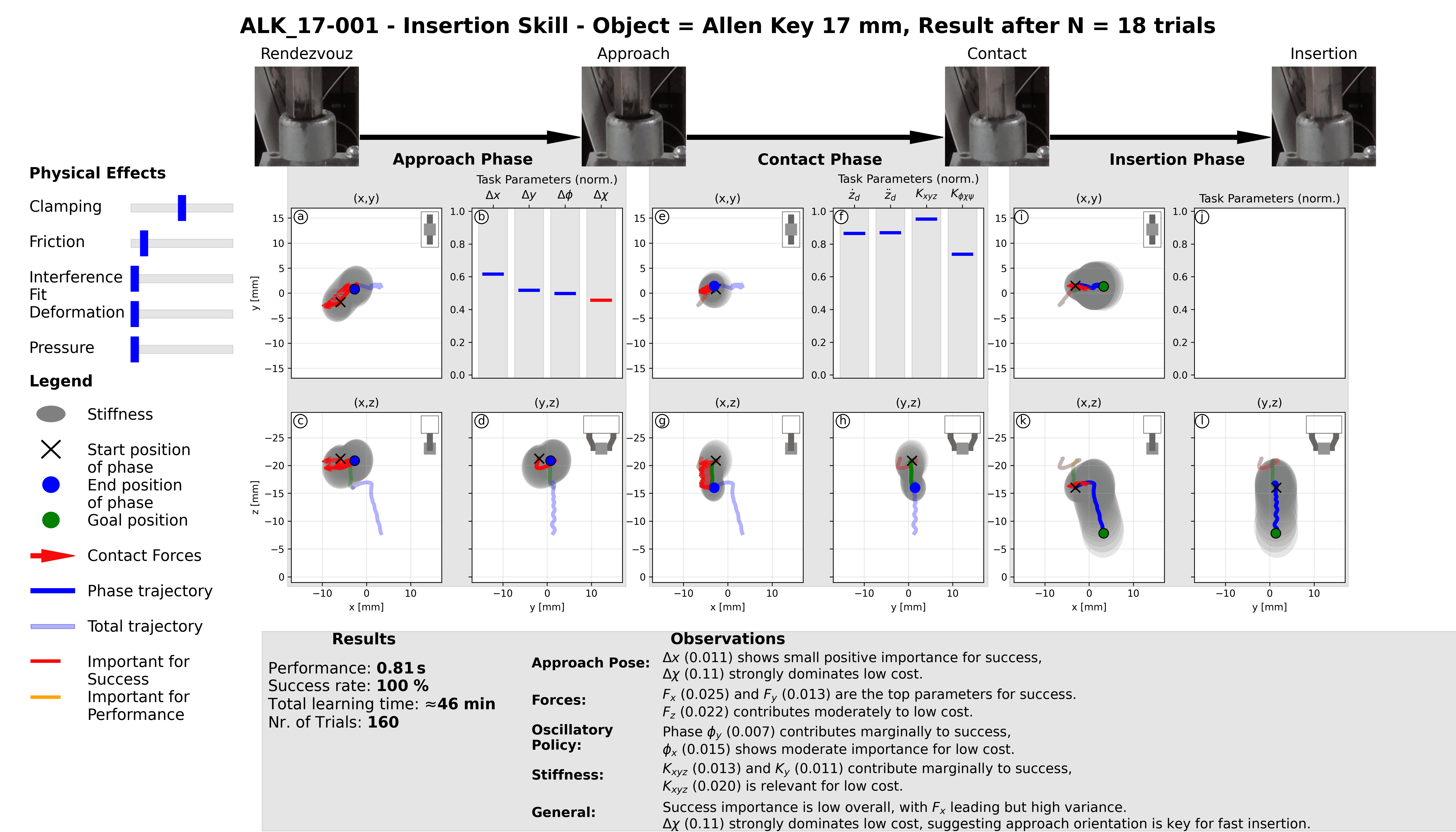}
\end{figure}

\begin{figure}[!t]
    \centering
    \includegraphics[
        width=1.0\textheight,
        angle=90
    ]{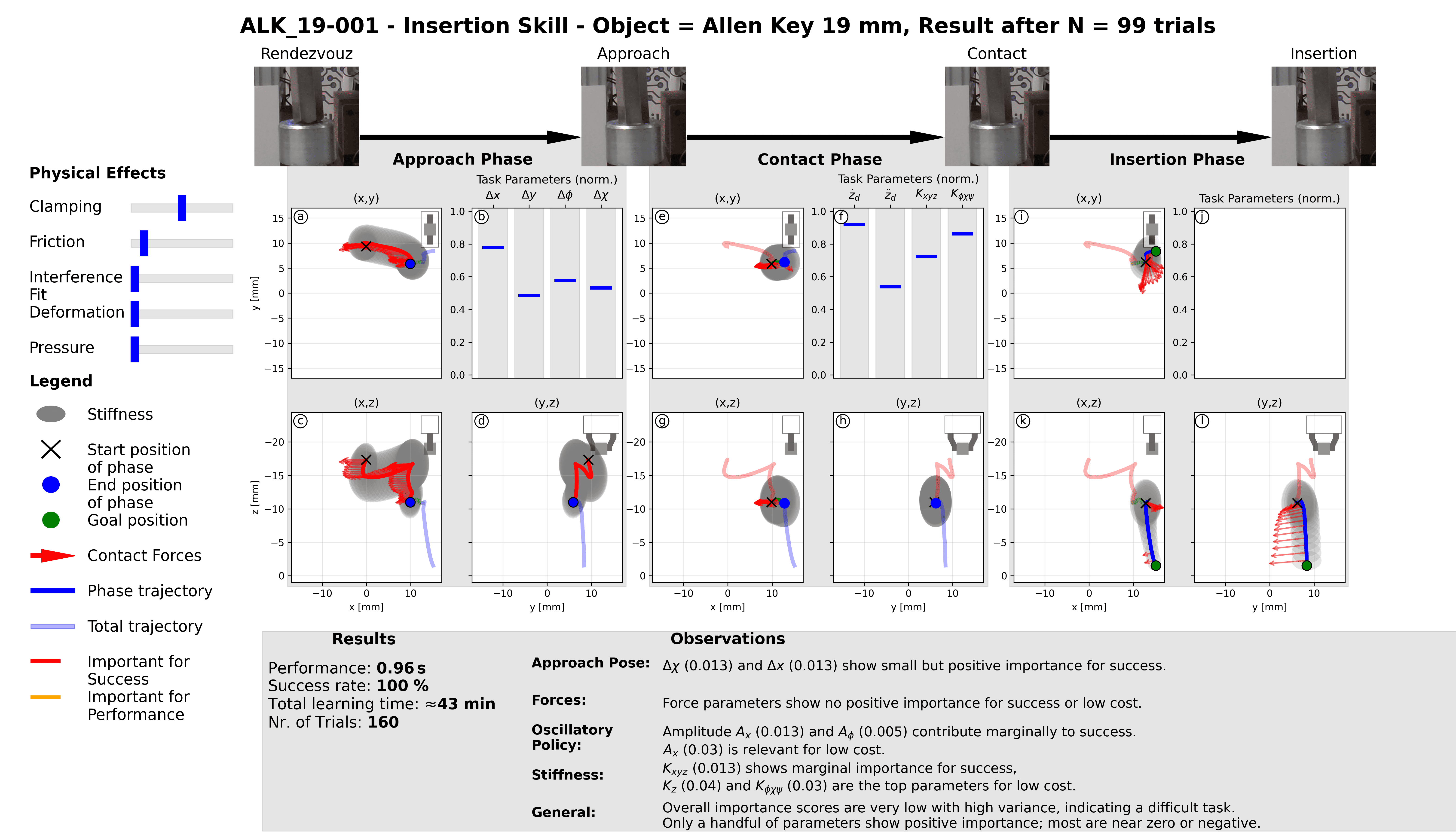}
\end{figure}

\begin{figure}[!t]
    \centering
    \includegraphics[
        width=1.0\textheight,
        angle=90
    ]{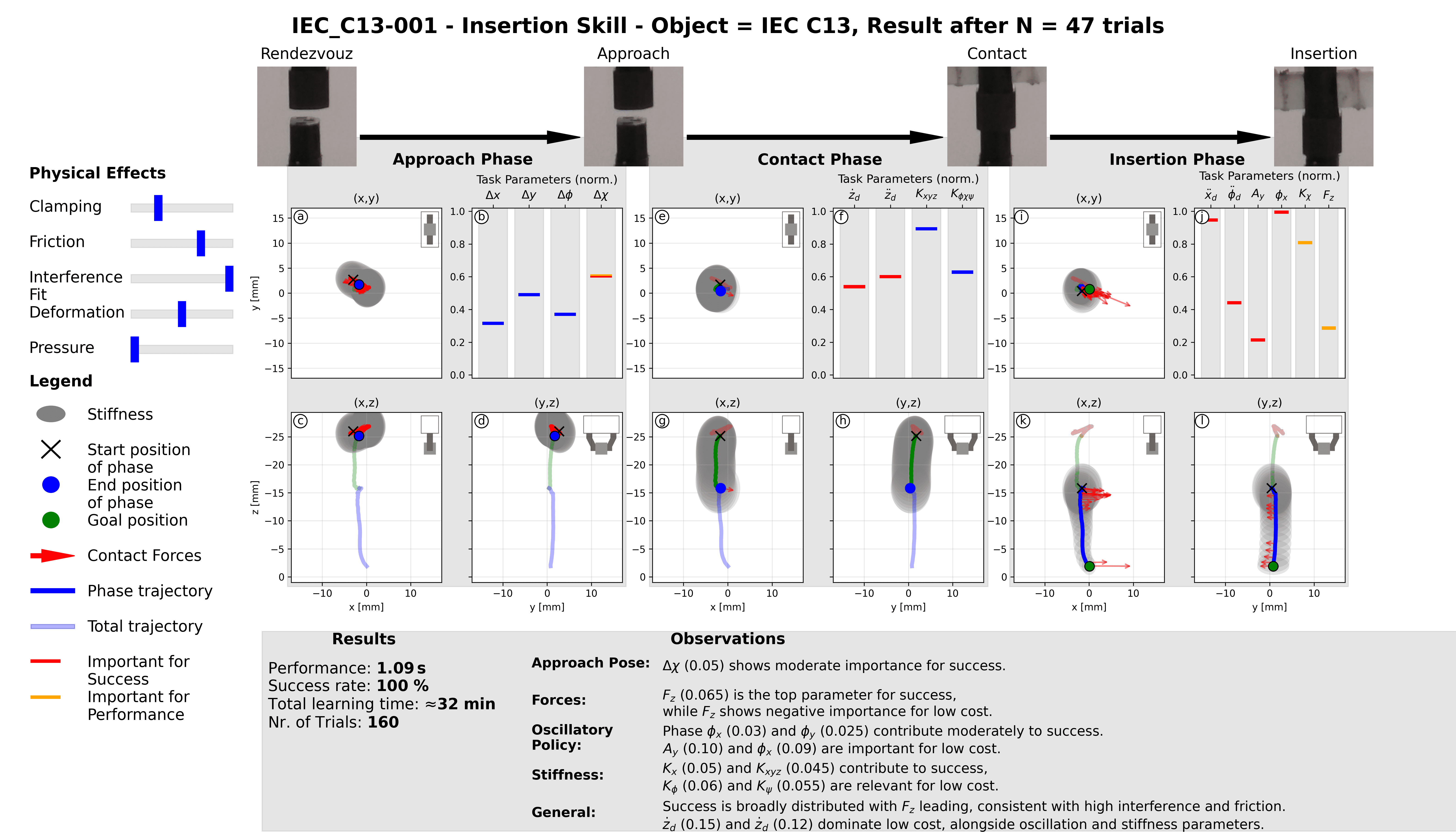}
\end{figure}

\begin{figure}[!t]
    \centering
    \includegraphics[
        width=1.0\textheight,
        angle=90
    ]{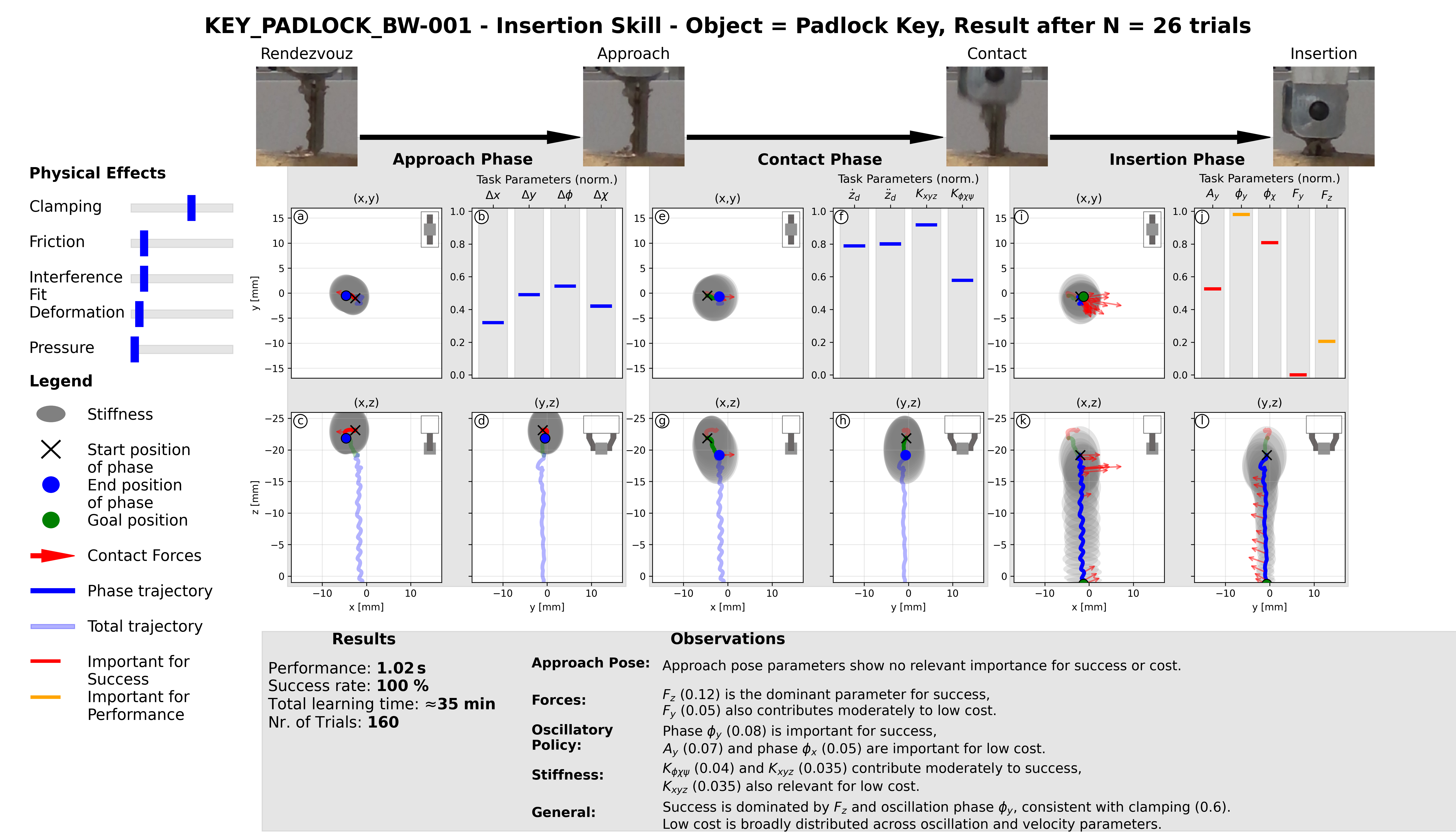}
\end{figure}

\begin{figure}[!t]
    \centering
    \includegraphics[
        width=1.0\textheight,
        angle=90
    ]{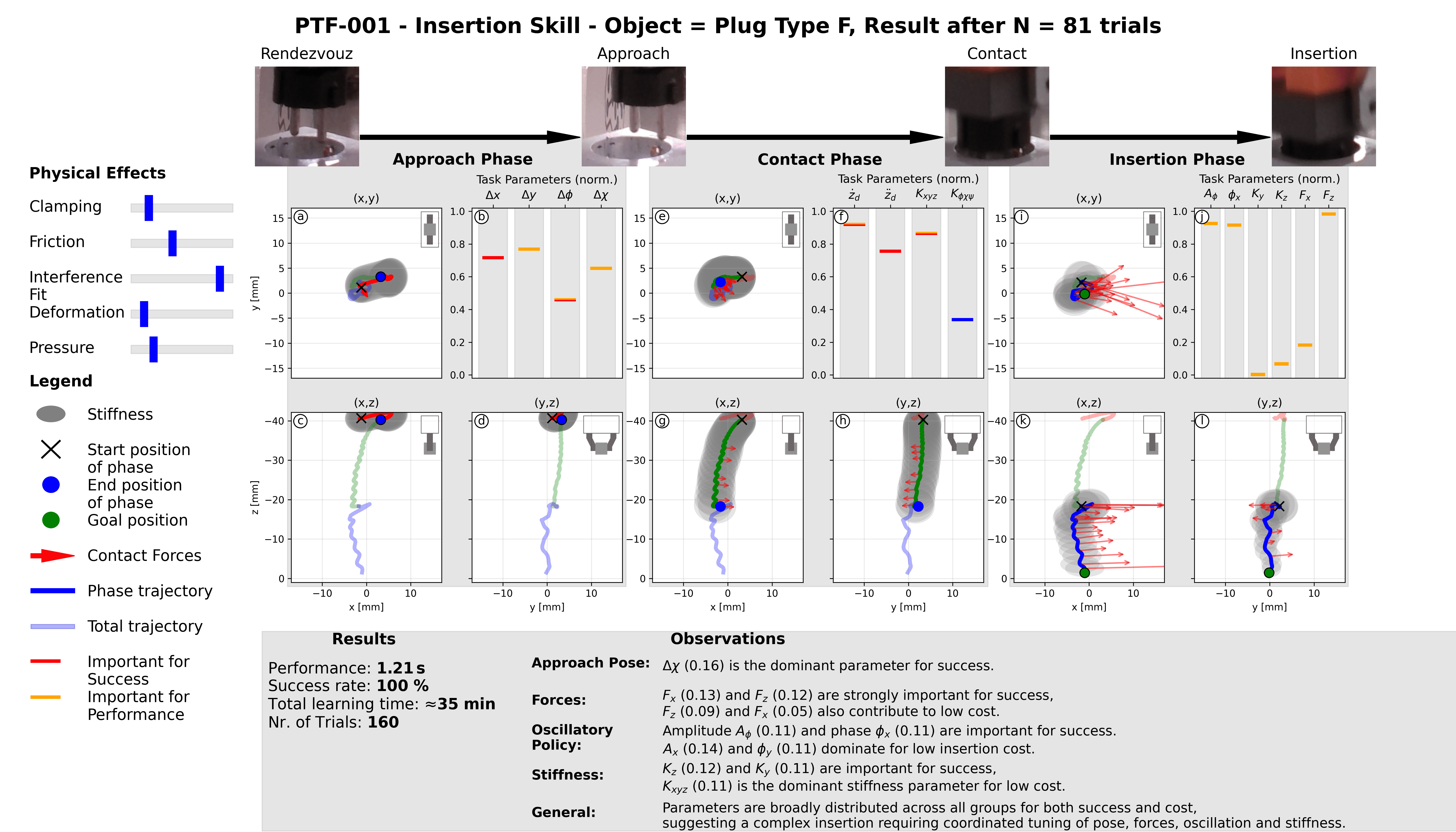}
\end{figure}

\begin{figure}[!t]
    \centering
    \includegraphics[
        width=1.0\textheight,
        angle=90
    ]{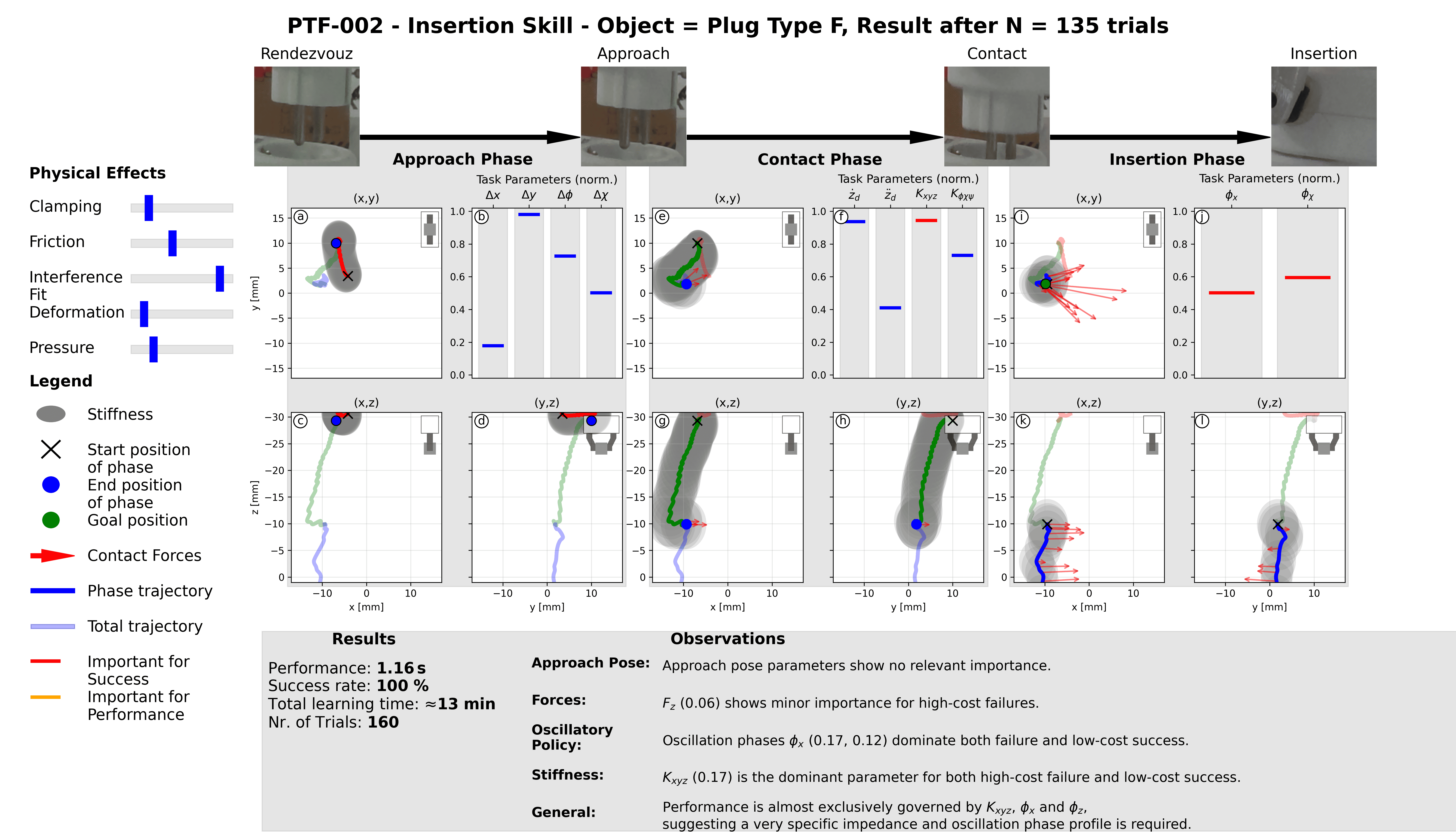}
\end{figure}

\begin{figure}[!t]
    \centering
    \includegraphics[
        width=1.0\textheight,
        angle=90
    ]{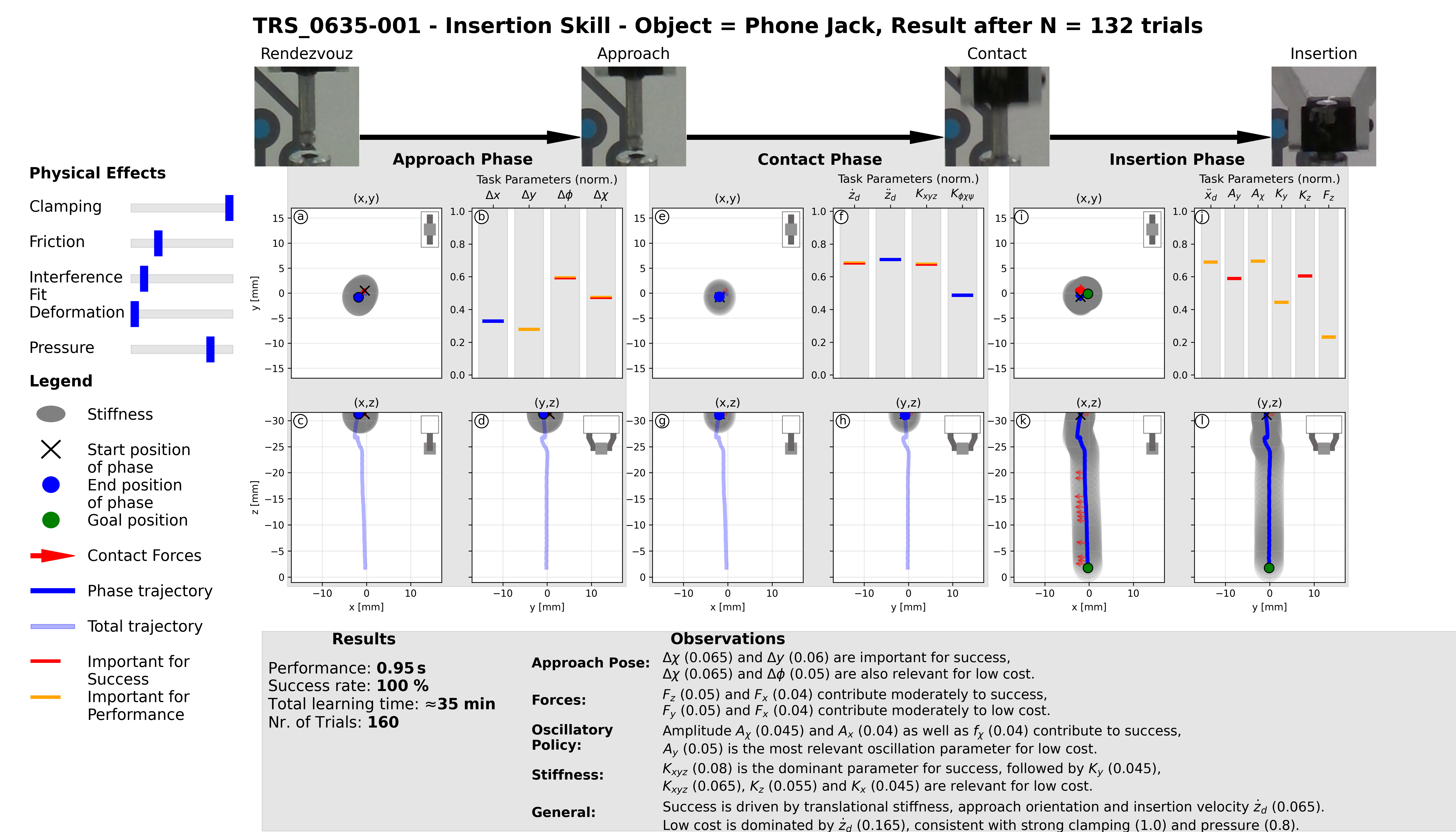}
\end{figure}

\begin{figure}[!t]
    \centering
    \includegraphics[
        width=1.0\textheight,
        angle=90
    ]{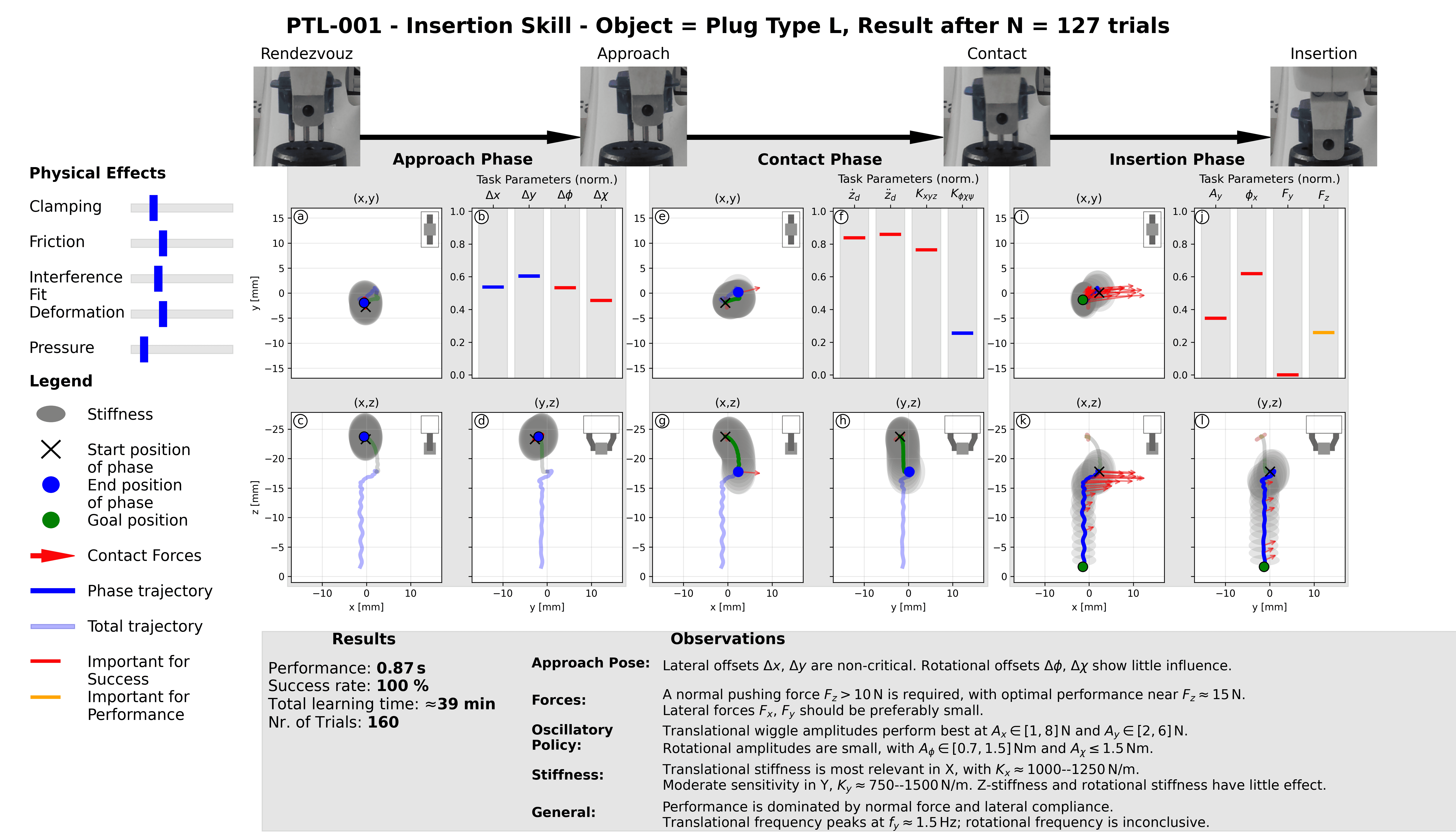}
\end{figure}

\begin{figure}[!t]
    \centering
    \includegraphics[
        width=1.0\textheight,
        angle=90
    ]{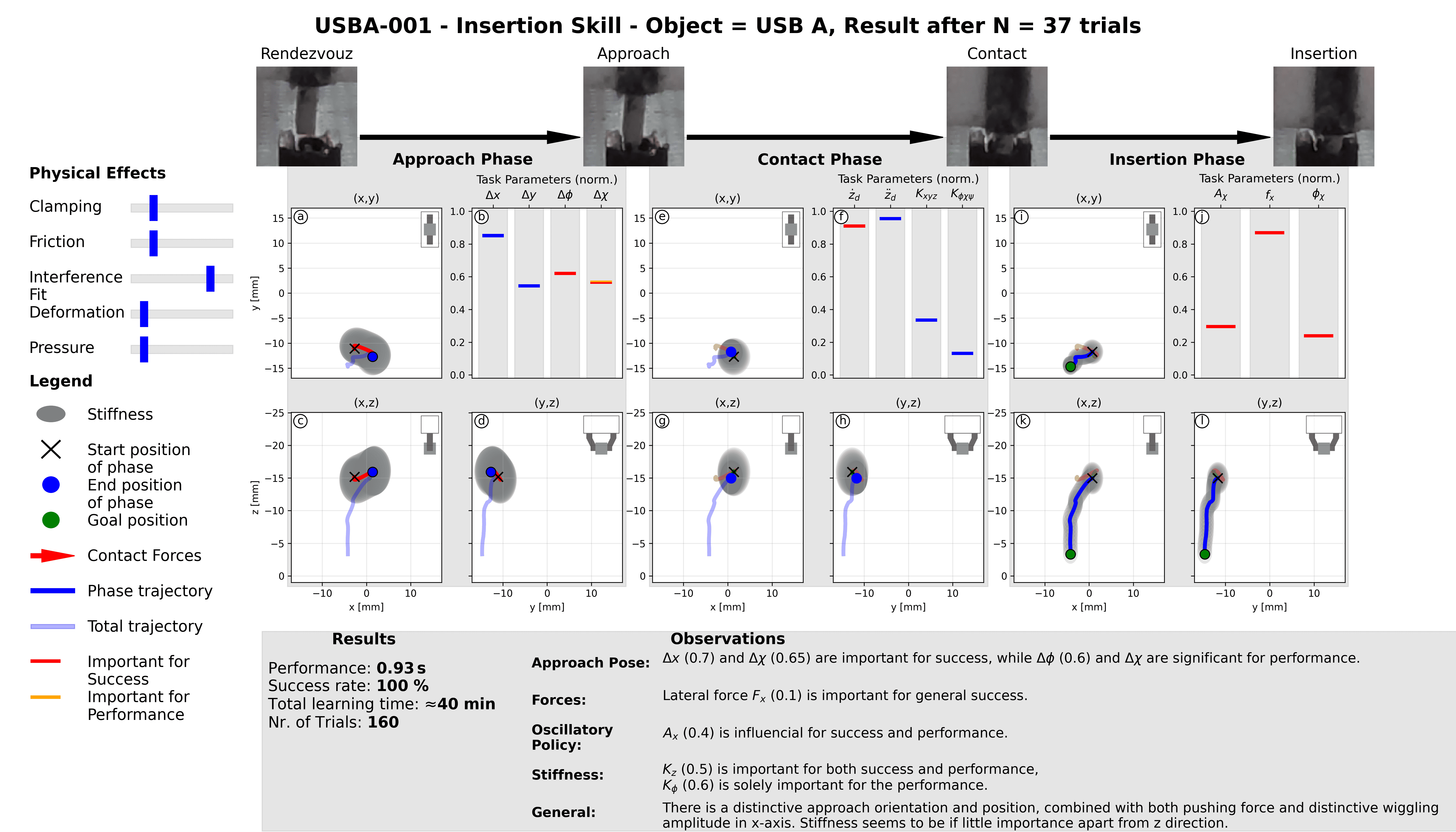}
\end{figure}

\begin{figure}[!t]
    \centering
    \includegraphics[
        width=1.0\textheight,
        angle=90
    ]{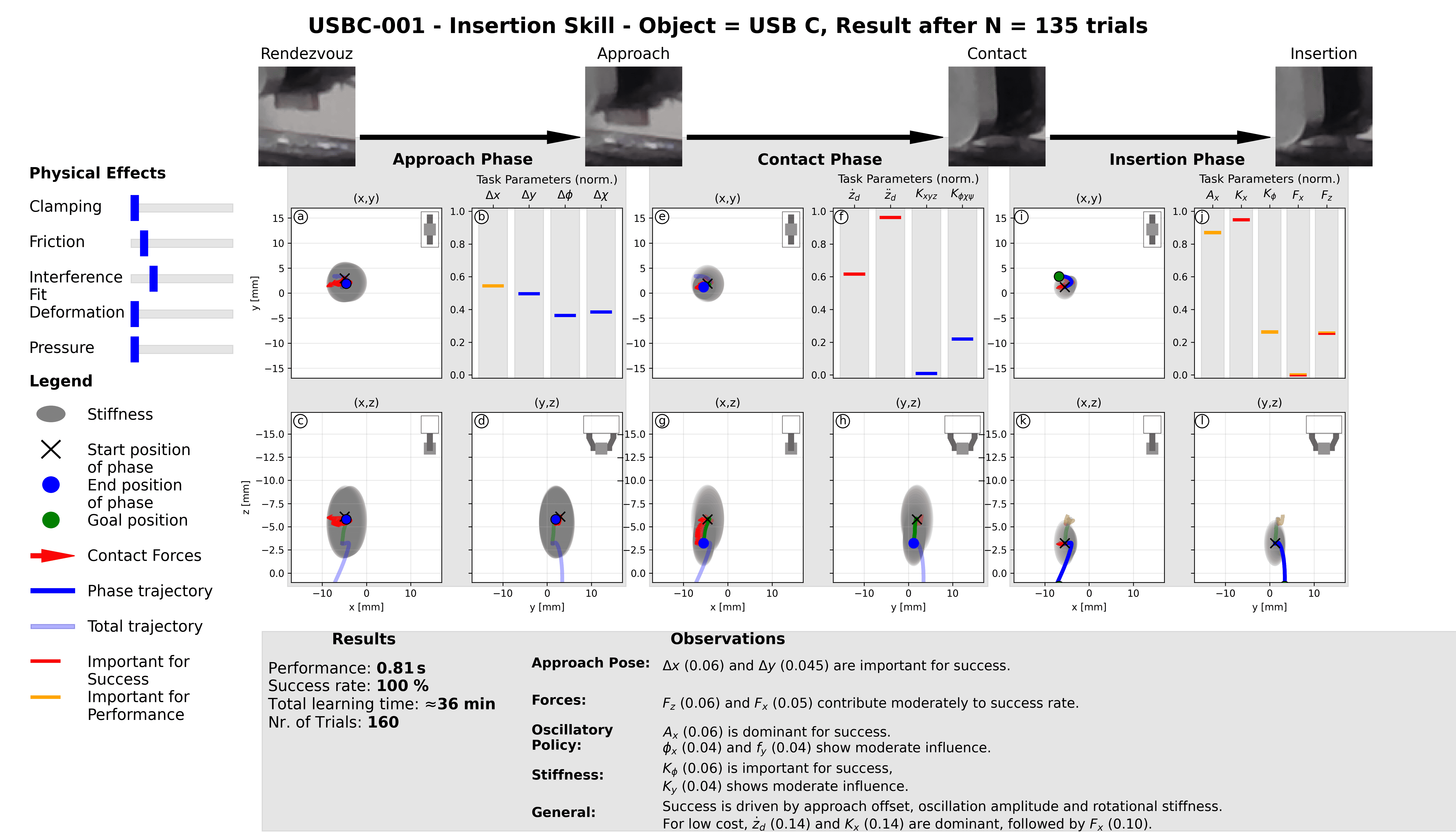}
\end{figure}

\clearpage
\clearpage

\section*{Overview of Supplementary 
Online Data}
The \texttt{Benchmark Statistics/} 
folder contains deep\,RL experiment 
results for all 22 tasks, two 
algorithms, five model knowledge 
levels, and two control domains across 
three independent runs per 
configuration, as well as Tactile Skill 
learning data and the performance 
summary tables reported in this work 
(JSON format).
The \texttt{Clearance Experiment/} 
folder contains per-trial success 
flags, cycle times, and 
distance-to-goal values for the 
controlled clearance and geometry 
experiment across three cross-sectional 
shapes and four clearance levels 
(PKL format).
The \texttt{Industrial Statistics/} 
folder contains per-batch execution 
times and success flags for all four 
industrial assembly workflows 
(PKL format).
The \texttt{Physical Interpretation 
Data/} folder contains precomputed SVM 
permutation importance scores, F1 
scores per task and classification 
target, and normalised physical effect 
magnitude scores P1--P5 per task as 
assigned by expert consensus 
(PKL and JSON format).
The \texttt{Tactile Skill Solution 
Data/} folder contains the best 
learned parameter vector 
$\boldsymbol{\theta}$ and the 
corresponding recorded force--motion 
trajectory and cycle time per task 
(JSON format).

\paragraph{Scripts.}
All analysis and plotting scripts are 
provided in the \texttt{Scripts/} 
folder and are run from that directory 
using relative paths. Required 
dependencies are \texttt{numpy}, 
\texttt{matplotlib}, \texttt{scipy}, 
\texttt{scikit-learn}, and 
\texttt{pillow}.

\paragraph{Videos.}
The \texttt{Videos/Benchmark Tactile 
Skill Solutions/} folder contains 
videos of the best learned policy 
executing each of the 22 benchmark 
insertion tasks.
The \texttt{Videos/Industrial 
Automation/} folder contains full 
assembly run videos for all four 
industrial partners.
The \texttt{Videos/Industrial Learning 
Examples/} folder contains selected 
learning example videos illustrating 
the progression from early exploration 
to reliable execution.

\paragraph{Script--figure mapping.}

\begin{center}
\small
\begin{tabular}{ll}
\hline
\textbf{Script} & \textbf{Figure} \\
\hline
\texttt{Model\_success\_performance\_comparison.py} & 
  Main Fig.~2 \\
\texttt{Clearance\_experiment.py} & 
  Main Fig.~3; Supp.\ Fig.~S2 \\
\texttt{Destroyed\_objects\_statistics.py} & 
  Main Fig.~4 \\
\texttt{Physical\_regression\_per\_model.py} & 
  Main Fig.~5 \\
\texttt{Hilbert\_structure\_analysis.py} & 
  Main Fig.~6 \\
\texttt{Parameter\_interpretation\_cost\_vs\_success.py} & 
  Main Fig.~7; Supp.\ Fig.~S4, S5 \\
\texttt{Industrial\_learning\_results.py} & 
  Main Fig.~8 \\
\texttt{Automation\_performance\_comparison.py} & 
  Main Fig.~8 \\
\texttt{Countries\_wage\_comparison.py} & 
  Main Fig.~8 \\
\texttt{Hilbert\_sensitivity.py} & 
  Supp.\ Fig.~S3 \\
\texttt{Parameter\_interpretation\_cost\_vs\_failure.py} & 
  Supp.\ Fig.~S6 \\
\texttt{F1\_summary\_and\_ablation\_plots.py} & 
  Supp.\ Fig.~S7 \\
\hline
\end{tabular}
\end{center}